\documentclass[10pt]{article}
\usepackage{iftex}
\ifXeTeX
  \usepackage[T1]{fontenc}
\fi
\usepackage{hmi-preprint}
\usepackage{amsmath}
\usepackage{amsfonts,amsthm}
\usepackage{url}
\usepackage{booktabs}
\usepackage{tabularx,multirow,array}
\usepackage{longtable}
\usepackage{placeins}
\usepackage{nicefrac}
\usepackage{microtype}
\usepackage{marvosym}
\usepackage{xcolor}         
\definecolor{cvprblue}{rgb}{0.21,0.49,0.74}
\hypersetup{
    colorlinks=true, 
    linkcolor=cvprblue, 
    citecolor=cvprblue, 
    urlcolor=cvprblue,  
}

\preprintlogos{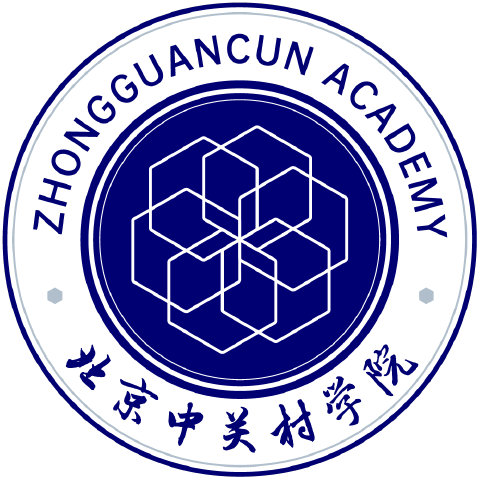}{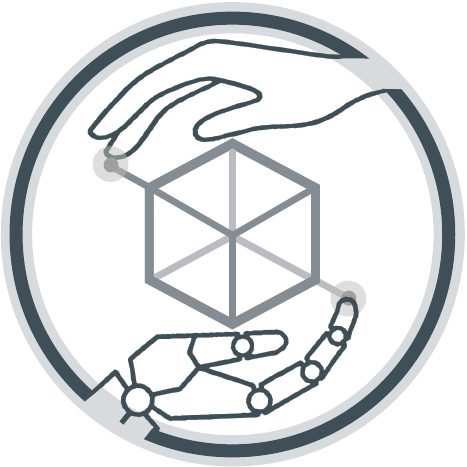}

\preprintdate{}

\runningtitle{GRC-Pose: Generation-Reconstruction Correspondence for Prior-Free 6D Object Pose Tracking}

\preprintfooter{}

\preprintlinkset{}{}{}

\preprintteaser{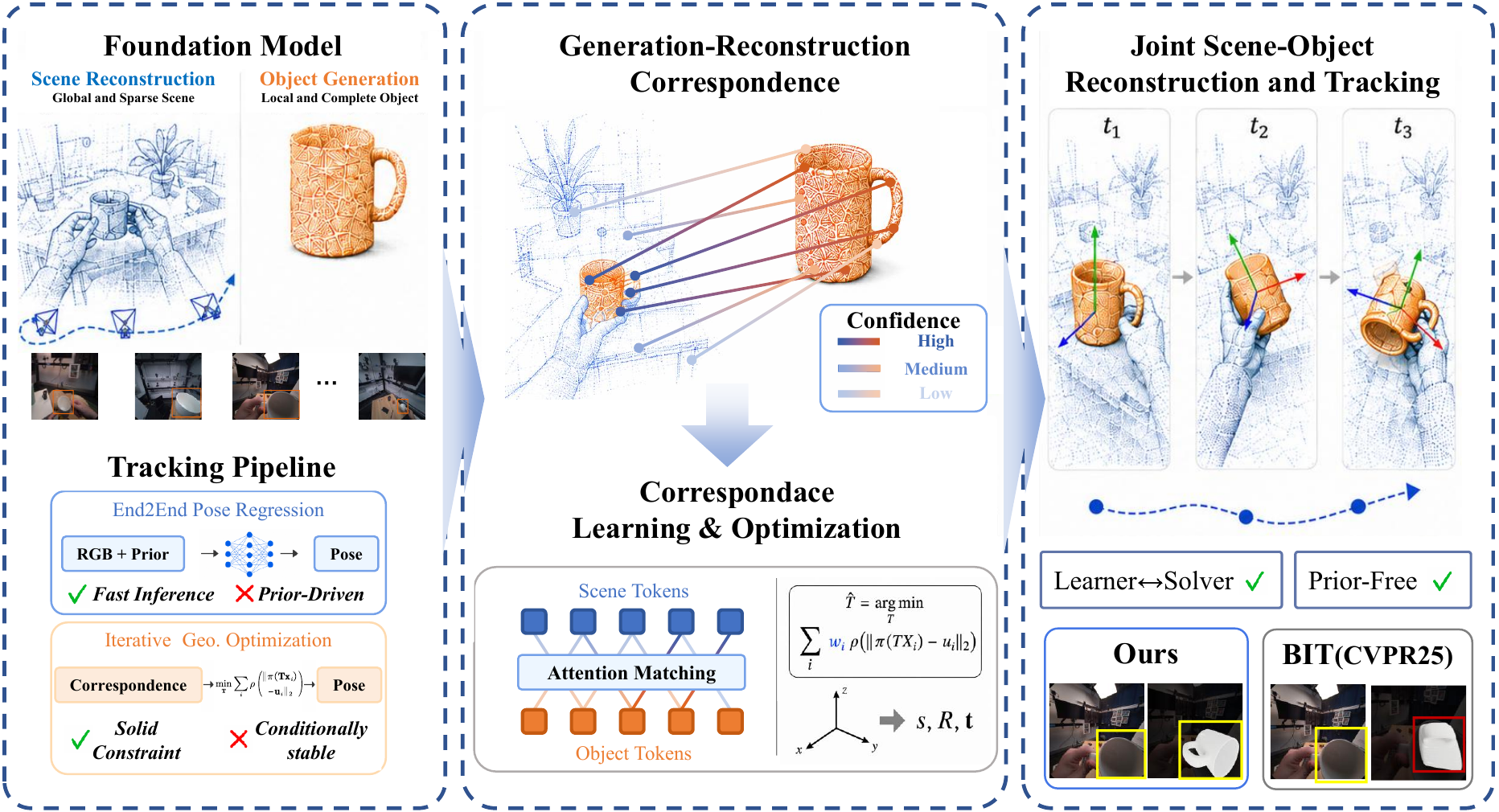}{\textbf{GRC-Pose overview.} Generation-reconstruction
    correspondence provides the geometric evidence for stable 6D
    object tracking across an RGB sequence.}

\newcommand{\SO}{\mathrm{SO}(3)}

\newcommand{\Sim}{\mathrm{Sim}(3)}
\newcommand{\R}{\mathbb{R}}

\newcommand{\G}{\mathbf{G}}
\newcommand{\x}{\mathbf{x}}

\newcommand{\calH}{\mathcal{H}}

\newcommand{\ours}{GRC-Pose}

\newcommand{\papertablestyle}{%
  \small
  \setlength{\tabcolsep}{6pt}%
  \renewcommand{\arraystretch}{1.16}%
}

\title{GRC-Pose: Generation-Reconstruction Correspondence \\
for Prior-Free 6D Object Pose Tracking}

\author[1,2]{Shiyang Liu}
\author[3,2]{Weiquan Lin}
\author[4,2]{Luping Xiao}
\author[1]{Jiadong Tang}
\author[1]{\authorcr Yi Yang}
\author[1]{Yu Gao}
\author[2\,\Letter]{Xingyu~Chen}
\affil[1]{School of Automation, Beijing Institute of Technology}
\affil[2]{Zhongguancun Academy}
\affil[3]{Artificial Intelligence Academy, Xidian University}
\affil[4]{School of Information and Communication Engineering,
Beijing University of Posts and Telecommunications}

\begin{document}

\maketitle

\begin{abstract}
Prior-free 6D object pose tracking seeks to recover the trajectory of an unseen object
from a single RGB video without object-specific CAD models, posed reference
images, or pose annotations.  Geometric foundation models provide complementary
object-centric and scene-centric cues, yet SAM3D CAD is indexed by an
arbitrary object-local surface parameterization, whereas reconstructed evidence is
expressed in a sequence-specific world frame with partial surface coverage.  To
exploit this complementarity,
we formulate tracking as generation-reconstruction correspondence and
introduce GRC-Pose, a correspondence-based framework that combines learned
correspondence prediction with robust pose estimation.  Concretely,
GeoCorr-Matcher estimates weighted object-scene correspondences and per-match
uncertainty for each pose candidate.  FGH-Solver integrates these matches
through multiple robust geometric estimators and sequence-level posterior
inference, while a posterior-gated memory retains only inlier-supported
observations through occlusion and viewpoint change. Extensive evaluation shows that
with SAM3D CAD, GRC-Pose achieves state-of-the-art Average Recall
and motion retention on HOT3D, improving the latter by 58\% over prior art.
On classical benchmarks including YCBInEOAT and LINEMOD, it remains highly competitive.
\end{abstract}

\section{Introduction}

Reliable 6D object pose tracking is important for manipulation, AR/VR, and
robotic interaction~\cite{deng2019poserbpf,wen2023bundlesdf,
wen2024foundationpose,guo2025rgbtrack,ponimatkin2025freepose}.  Egocentric
hand-object interaction is particularly challenging because large inter-frame motion,
severe hand occlusion, motion blur, and intermittent visibility easily lead to
severe pose drift and tracking loss over time
\cite{banerjee2024hot3d,fan2024egochallenge}.  Robust tracking therefore
requires a spatiotemporally consistent object state throughout the sequence rather
than independent framewise estimations.

Existing unseen object pose estimation methods typically rely on either an object-specific
CAD model or posed reference views, and learning-based methods heavily rely on large-scale synthetic
pose supervision~\cite{sahin2020review,wen2024foundationpose,
labbe2022megapose,liu2022gen6d,sun2022onepose,he2022oneposepp,
lee2025any6d}.  These requirements restrict real-world applicability on
unconstrained in-the-wild objects. Furthermore, single-view estimation is inherently ill-posed
due to geometric symmetry and severe occlusion, which frequently induce pose ambiguity or
yield multiple pose candidates for a single observation~\cite{hodan2020epos,haugaard2022surfemb}.
Sequence-level tracking inherently addresses both challenges by replacing static single-view
inference with joint 3D geometry reconstruction and pose optimization across frames.
BundleTrack~\cite{wen2021bundletrack} and BundleSDF~\cite{wen2023bundlesdf} demonstrate
robust RGB-D tracking without CAD models, while BIT~\cite{song2025bit} further advances to
a completely prior-free framework on monocular RGB.

Recent geometric foundation models offer orthogonal geometric strengths.
Feed-forward reconstruction models such as VGGT~\cite{wang2025vggt,wang2026vggtomega} infer
cameras and scene geometry from multiple views, whereas single-image
generation models such as SAM~3D~\cite{sam3dteam2025sam3d} and
TRELLIS~\cite{xiang2024trellis} produce dense object-centric geometry. However, these two
representations resist naive feature or spatial concatenation due to fundamental
misalignments in coordinate frame, scale, and completeness.
To leverage their synergistic potential for tracking, our key insight is to formulate pose estimation as
cross-domain matching and pose solving, using the global scene context and camera poses from multi-view
reconstructions to metrically ground the dense, complete canonical geometry of generative models.

In this paper, we propose \emph{Generation-Reconstruction Correspondence Pose Tracking (GRC-Pose)},
a correspondence-driven framework that aligns SAM3D CAD with feed-forward
multi-view scene reconstructions via deep feature association and robust pose optimization. Specifically,
the \emph{Geometric Correspondence Matcher (GeoCorr-Matcher)} first estimates pose-conditioned weighted
correspondences alongside match-wise uncertainty. Next, the \emph{Factorized Geometric Hybrid Solver (FGH-Solver)}
leverages these predictions to derive complementary 2D--3D and 3D--3D pose candidates,
which are subsequently propagated to temporal posterior filtering. To prevent error accumulation, a
posterior-gated memory bank selectively retains only geometrically consistent keyframe observations.
Finally, sequence-specific adaptation enables online self-supervised calibration of matching confidence
and geometric uncertainty. Our contributions are summarized as follows:
\begin{itemize}
    \item We propose GRC-Pose, a novel correspondence-driven paradigm that formulates prior-free
    6D pose tracking as a cross-domain geometric consensus problem, explicitly maintaining
    multimodal pose distributions to resolve the inherent ambiguities induced by severe occlusion,
    agile ego-motion, and object symmetries.
    \item We design GeoCorr-Matcher and FGH-Solver to bridge the generative and reconstructive domains,
    where pose-conditioned uncertainty-aware matching rejects correspondence noise while factorized
    2D--3D and 3D--3D estimation resolves scale discrepancies, together enabling reliable propagation of
    pose candidates for temporal filtering and posterior-gated memory updates.
    \item Experiments across public benchmarks show that GRC-Pose achieves state-of-the-art
    performance on the egocentric HOT3D dataset featuring rapid motion and occlusion,
    while achieving competitive results on classical
    object-centric datasets such as YCBInEOAT and LINEMOD.
\end{itemize}

\section{Related Work}

\paragraph{Novel-object pose estimation and tracking.}

Novel-object pose methods differ mainly in the object-specific information
available at test time.
FoundationPose~\cite{wen2024foundationpose} unifies CAD- and reference-based
render-and-compare inference; GigaPose~\cite{nguyen2024gigapose} combines
template retrieval with patch correspondences; and GoTrack~\cite{nguyen2025gotrack}
couples object-model refinement with optical-flow registration.  Any6D~\cite{lee2025any6d}
estimates pose and metric size from one RGB-D anchor,
whereas SAM-6D~\cite{lin2024sam6d} uses segmentation and partial-to-partial
RGB-D matching.  FoundPose~\cite{ornek2024foundpose} establishes
CAD-to-image correspondences with foundation features.  FreeZe~\cite{caraffa2024freeze} and its follow-up
FreeZeV2~\cite{caraffa2025freezev2} perform training-free registration with
frozen visual-geometric descriptors, while MatchU~\cite{huang2024matchu}
matches a CAD point cloud to a single RGB-D image and SinRef-6D~\cite{liu2026sinref6d}
uses a pose-labeled RGB-D reference.  SurfEmb~\cite{haugaard2022surfemb} learns dense surface
embeddings; NeuSurfEmb~\cite{milano2024neusurfemb} trains correspondences on a
reconstructed neural surface; and Zero6D~\cite{caraffa2023zero6d} transfers
pretrained visual-geometric descriptors.  These methods collectively target
per-image registration given a CAD model, posed reference, metric depth, or
large-scale pose-supervision assumption.  Probabilistic PnP methods such as
EPro-PnP~\cite{chen2024epropnp} propagate correspondence uncertainty to a
single-frame pose distribution.

\paragraph{Joint reconstruction and pose optimization.}

Joint reconstruction methods estimate object geometry and pose from the same
sequence.  BundleTrack~\cite{wen2021bundletrack} and
BundleSDF~\cite{wen2023bundlesdf} maintain persistent object representations
while optimizing pose from RGB-D video.  HOLD~\cite{fan2024hold} reconstructs
monocular hand-object interactions, and Dyn-HOR~\cite{jiang2025hand}
introduces generated geometry into dynamic reconstruction.  BIT~\cite{song2025bit}
jointly generates geometry and optimizes pose from monocular RGB;
UA-Pose~\cite{li2025uapose} combines uncertainty-aware pose estimation with
online completion; and 6DOPE-GS~\cite{jin2025sixdopegs} couples Gaussian-splat
reconstruction with RGB-D tracking.  GSGTrack~\cite{chen2024gsgtrack} jointly
optimizes a Gaussian object representation and pose from RGB videos, while
KV-Tracker~\cite{taher2025kvtracker} caches a multi-view reconstruction model
for online camera and object mapping.  SGPose~\cite{luo2024sgpose} estimates
object pose from a sparse set of approximately ten onboarding views.  These
systems address adjacent reconstruction-pose objectives under method-specific
mapping or sparse-view protocols.

\paragraph{Geometry foundation models and sequential data association.}

Geometry foundation models provide complementary evidence: DUSt3R~\cite{wang2024dustr},
MASt3R~\cite{leroy2024mast3r}, and VGGT~\cite{wang2025vggt} infer point maps,
camera geometry, and 3D-aware matches from images, whereas
SAM~3D~\cite{sam3dteam2025sam3d} and TRELLIS~\cite{xiang2024trellis} generate
object-centric geometry from a single image.
SuperGlue~\cite{sarlin2020superglue}, LoFTR~\cite{sun2021loftr}, GeoTransformer~\cite{qin2022geotransformer}, and
RegTR~\cite{yew2022regtr} learn attention-based correspondences for image or
point-cloud registration.

\FloatBarrier
\section{Method}
\label{sec:method}

\begin{figure}[tbp]
    \centering
    \includegraphics[width=0.99\textwidth]{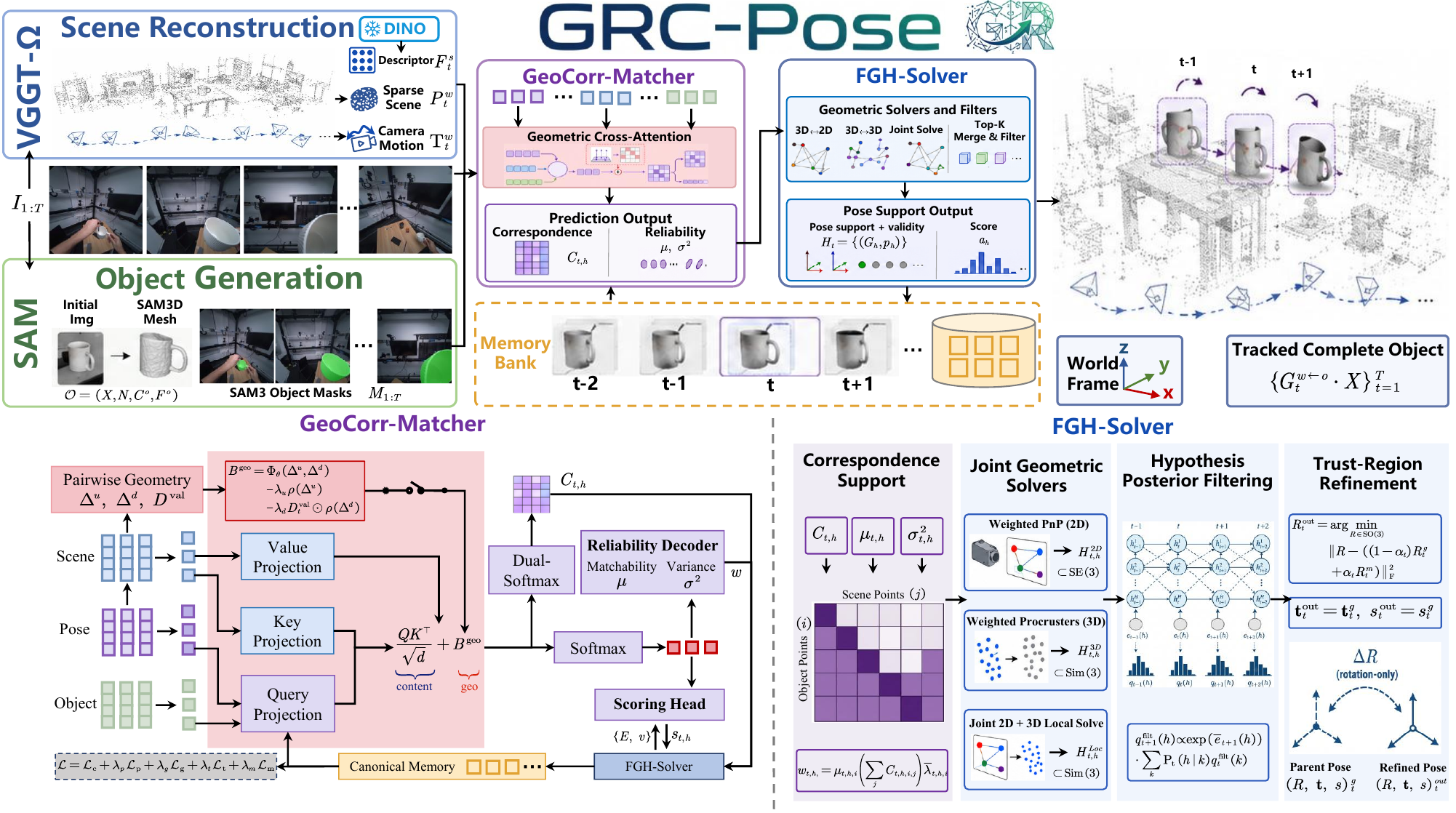}
    \caption{\textbf{Overview of GRC-Pose.}
    The top panel maps input video $I_{1:T}$, SAM3D CAD geometry, and scene-level reconstructions
    to 6D object trajectories in the world frame. The bottom panel details pose-conditioned correspondences
    with match-level uncertainty, while \textbf{FGH-Solver} (right) resolves factorized pose candidates, performs temporal filtering with
    rotation refinement, and guides surface-memory updates.}
    \label{fig:method_overview}
\end{figure}

Given a monocular RGB video $I_{1:T}$, GRC-Pose recovers a continuous 6D object trajectory in
a global world frame by unifying SAM3D CAD geometry with scene-level
reconstructions (Fig.~\ref{fig:method_overview}). The framework infers pose-conditioned correspondences with confidence
that evaluate pose candidates against scene context. The correspondences are then solved into factorized pose candidates,
propagated through a temporal filter with posterior-gated surface memory, and refined online via
sequence-specific self-supervision.

\subsection{Task Formulation}
\label{sec:task}

We represent the 6D object trajectory and reconstructed scene geometry in a global world frame.
Canonical 3D surface points $\mathbf x_i$ supplied by SAM3D CAD are mapped to the world frame at frame $t$
via a similarity transformation $\mathbf G_t\cdot\mathbf x_i = s_tR_t\mathbf x_i + \mathbf t_t$,
where $\mathbf G_t=(s_t,R_t,\mathbf t_t)\in\Sim$, $R_t\in\SO$, $\mathbf t_t\in\R^3$, and $s_t>0$.
Metric CAD fixes scale ($s_t=1$), whereas SAM3D CAD requires joint scale and pose estimation.
Given camera intrinsics $\mathbf K_t$ and VGGT-$\Omega$ reconstructed world-to-camera extrinsics
$\mathbf T_t^{c\leftarrow w}$, a world point $\mathbf y$ projects as
$\pi_t(\mathbf y) = \pi(\mathbf K_t\mathbf T_t^{c\leftarrow w}\mathbf y)$.
Augmenting these object and scene samples with DINOv3~\cite{simeoni2025dinov3} visual descriptors, the system
maintains $H$ pose candidates $\{\mathbf G_t^h\}_{h=1}^{H}$ at frame $t$ associated with a temporal posterior
distribution $q_t(h)$.

\subsection{GeoCorr-Matcher: Pose-Conditioned Correspondence}
\label{sec:matcher}

We construct the initial pose candidate pool $\mathcal H_t^{(0)}$ by aggregating multi-source proposals
followed by Non-Maximum Suppression (NMS) to eliminate redundancy:
\begin{equation}
\label{eq:candidate_pool}
\mathcal H_t^{(0)} = \operatorname{NMS}\!\left(\mathcal H_t^{\mathrm{retr}} \cup \mathcal H_t^{\mathrm{2D\text{-}3D}} \cup \mathcal H_t^{\mathrm{3D\text{-}3D}} \cup \mathcal H_{t-1}^{\mathrm{prop}}\right).
\end{equation}
where $\mathcal H_t^{\mathrm{retr}}$ provides global retrieval candidates,
$\mathcal H_t^{\mathrm{2D\text{-}3D}}$ and $\mathcal H_t^{\mathrm{3D\text{-}3D}}$ derive geometric candidates
via PnP and Procrustes solvers~\cite{lepetit2009epnp,umeyama1991least},
and $\mathcal H_{t-1}^{\mathrm{prop}}$ propagates previous high-posterior candidates using estimated relative
motion $\widehat{\Delta\mathbf G}_{t-1\to t}$~\cite{nguyen2024gigapose,wen2024foundationpose}.

For each candidate $\mathbf G_t^h$, a pose encoder $E_p$ converts its normalized parameters
$\boldsymbol\chi_{t,h}$ consisting of 6D rotation, scale-normalized translation, and log-scale factors,
together with rendered depth and visibility cues, into a candidate embedding $\mathbf P_{t,h}\in\R^{N_o\times d}$.
Dedicated feature encoders simultaneously project canonical object surface context and scene observations
into feature representations $\mathbf Z^o\in\R^{N_o\times d}$ and $\mathbf Z_t^s\in\R^{N_s\times d}$.
GeoCorr-Matcher incorporates $\mathbf P_{t,h}$ to modulate cross-attention, allowing identical appearance
features to infer pose-conditioned soft correspondences.

The matcher constructs a pose-conditioned object-scene affinity matrix. Object queries incorporate canonical surface
features, projected pose embeddings, and historical surface memory, while scene tokens provide attention keys
$\mathbf A_t^s=\mathbf W_k\mathbf Z_t^s$ and values $\mathbf V_t^s=\mathbf W_v\mathbf Z_t^s$.
The resulting attention logits and aggregated scene context are formulated as:
\begin{equation}
\label{eq:attention}
\begin{aligned}
\mathbf Q_{t,h}&=\mathbf W_q\!\left(\mathbf Z^o+\mathbf P_{t,h}+\mathbf W_m\mathbf M_{t-1}^o\right),\\
\mathbf L_{t,h}&=\frac{\mathbf Q_{t,h}(\mathbf A_t^s)^\top}{\sqrt d}+\mathbf B_{t,h}^{\mathrm{geo}},\\
\mathbf O_{t,h}&=\operatorname{softmax}_{j}(\mathbf L_{t,h})\mathbf V_t^s,
\end{aligned}
\end{equation}
The relative geometric bias $\mathbf B_{t,h}^{\mathrm{geo}}$ uses normalized 2D reprojection and 3D depth residuals:
$[\boldsymbol\Delta_{t,h}^{u}]_{ij}=\|\pi_t(\mathbf G_t^h\cdot\mathbf x_i)-\mathbf u_{t,j}\|/\kappa_u$ and $[\boldsymbol\Delta_{t,h}^{d}]_{ij}=|d_{t,h,i}-d_{t,j}|/\kappa_d$,
where $\kappa_u$ and $\kappa_d$ are predefined normalization factors, and $\mathbf D_t^{\mathrm{val}}$ is
a binary mask indicating valid depth observations. We define the relative geometric bias
$\mathbf B_{t,h}^{\mathrm{geo}}\in\R^{N_o\times N_s}$ as:
\begin{equation}
\label{eq:geometric_bias}
\mathbf B_{t,h}^{\mathrm{geo}}
=\Phi_\theta(\boldsymbol\Delta_{t,h}^{u},\boldsymbol\Delta_{t,h}^{d})
-\lambda_u\rho(\boldsymbol\Delta_{t,h}^{u})
-\lambda_d\mathbf D_t^{\mathrm{val}}\odot
\rho(\boldsymbol\Delta_{t,h}^{d}),
\end{equation}
where $\Phi_\theta$ projects relative spatial residual features into a scalar geometric compatibility score,
$\rho$ denotes the Huber penalty function, and $\lambda_u, \lambda_d$ are predefined penalty weights.
Invalid element pairs are masked out prior to attention normalization,
whereas the absence of metric depth deactivates only the 3D penalty term.
Consequently, identical appearance features yield distinct pose-dependent assignments across competing candidates.
Row-wise softmax normalization of $\mathbf L_{t,h}$ aggregates scene-level contextual features into
$\mathbf O_{t,h}$, whereas dual-softmax normalization computes the mutual assignment matrix
$\mathbf C_{t,h}=\operatorname{softmax}_{j}(\mathbf L_{t,h})\odot\operatorname{softmax}_{i}(\mathbf L_{t,h})$.
Row-normalized $\mathbf C_{t,h}$ produces soft expected coordinates $(\bar{\mathbf u}_{t,h,i},\bar{\mathbf p}_{t,h,i}^{w})$ for
downstream geometric pose optimization.

Dual-softmax mutual matching alone does not guarantee that a canonical surface point is observable or
geometrically reliable. A dedicated uncertainty decoder therefore predicts a point-wise matchability score
$m_{t,h,i}\in[0,1]$ alongside a 3D spatial variance vector $\boldsymbol\sigma_{t,h,i}^{2}\in\R_+^3$,
which together formulate the correspondence weighting factor for downstream geometric solving:
\begin{equation}
\label{eq:correspondence_weight}
\begin{aligned}
\bar\lambda_{t,h,i}&=\frac{1}{3}\sum_{c=1}^{3}
(\sigma_{t,h,i,c}^{2}+\epsilon)^{-1},\\
w_{t,h,i}&=m_{t,h,i}\!\left(\sum_j[\mathbf C_{t,h}]_{ij}\right)
\bar\lambda_{t,h,i}.
\end{aligned}
\end{equation}
where $\epsilon$ denotes a small positive constant ensuring numerical stability.
The composite weight $w_{t,h,i}$ explicitly factorizes into point-wise matchability,
mutual assignment confidence, and spatial precision. A candidate-level Transformer encoder
subsequently contextualizes these correspondence-derived features across candidates before
emission scoring.

To enforce strict pose conditioning, a counterfactual contrastive loss
permutes the embeddings of pose candidates $\boldsymbol\chi_{t,h}$ across mismatched candidates, penalizing
correspondences that fail to discriminate the correct geometric pose conditioning vector.

\paragraph{Posterior-gated surface memory.}
To maintain long-term feature consistency across video sequences, the framework anchors temporal evidence
directly to canonical surface coordinates rather than transient image planes. For each canonical
surface point $a$, frame-level Transformer tokens $\mathbf r_{t,h,n}$ are aggregated using their correspondence
weights $w_{t,h,n}$, producing a surface-indexed observation $\widetilde{\mathbf m}_{t,h,a}$.
Historical surface memories from the previous timestep are first transported across pose candidates using
filtering posteriors $q_{t-1}$, producing a pose-conditioned memory state $\mathbf M^-_{t,h,a}$.
The updated surface memory $\mathbf M^+_{t,h,a}$ is then derived via a learned reliability gate $g_{t,h}\in[0,1]$:
\begin{equation}
\label{eq:memory_update}
\mathbf M^+_{t,h,a}=(1-g_{t,h})\mathbf M^-_{t,h,a}+g_{t,h}\widetilde{\mathbf m}_{t,h,a}.
\end{equation}
where the gating factor $g_{t,h}$ explicitly incorporates frame observability, tracking reset flags,
and memory-evidence consistency.

\subsection{Factorized Geometric Hybrid Solver}
\label{sec:solver}

Taking the soft expected 2D projections $\bar{\mathbf u}_{t,h,i}$, 3D scene points $\bar{\mathbf p}_{t,h,i}^{w}$, and correspondence weights $w_{t,h,i}$
inferred by GeoCorr-Matcher, FGH-Solver derives factorized pose candidates via dual 2D--3D and 3D--3D geometric optimizations:
\begin{equation}
\label{eq:solvers}
\begin{aligned}
\widehat{\G}_{t,h}^{\mathrm{proj}}
&=\arg\min_{\substack{\G\in\Sim\\s(\G)=s_t^h}}
\sum_i w_{t,h,i}
\rho\!\left(\left\|\pi_t(\G\!\cdot\!\mathbf x_i)
-\bar{\mathbf u}_{t,h,i}\right\|_2\right),\\
\widehat{\G}_{t,h}^{\mathrm{sim}}
&=\arg\min_{\G\in\Sim}
\sum_i w_{t,h,i}
\left\|\G\!\cdot\!\mathbf x_i-\bar{\mathbf p}_{t,h,i}^{w}\right\|_2^2.
\end{aligned}
\end{equation}
The 2D projective branch minimizes Huber-penalized reprojection errors while constraining the scale factor
$s(\mathbf G)=s_t^h$ when 3D scene geometry is unreliable. Conversely, the 3D similarity branch performs
weighted Procrustes alignment over $\Sim$ when valid 3D scene point clouds are present. Rather than averaging
pose estimates prior to scoring, the solver retains all candidates from both branches that pass geometric support validation with the propagated candidate.

Each retained pose candidate is subsequently evaluated using dense silhouette and RGB residuals.
Following robust standardization across candidates, a weighted sum of these geometric residuals
combined with a learned confidence score $a_{t,h}$ defines the frame-level emission score $e_t(h)$.
Its stop-gradient geometric counterpart supervises temporal posterior calibration.
These per-frame pose candidates and calibrated emission scores constitute the input to
sequence-level temporal filtering.

\paragraph{Temporal Bayesian filtering.}
To ensure sequence-level consistency, the system filters the complete discrete posterior distribution over
pose candidates rather than relying on greedy single-frame selections. Inter-frame relative motion constraints
$\widehat{\Delta\mathbf G}_t$ derived from scene alignments govern transitions between consecutive candidates,
evaluated via Mahalanobis distances in $\Sim$. Given unary emissions $e_t(h)$, frame observability $o_t$, and
temperature scaling factor $\tau_t$, the Bayesian filtering recursion updates the temporal posterior $q_t$ via:
\begin{equation}
\label{eq:filter}
q_{t+1}(h)\propto \exp\!\left(\frac{o_{t+1}e_{t+1}(h)}{\tau_{t+1}}\right) \sum_k\mathsf P_t(h\mid k)q_t(k).
\end{equation}

Evaluated in log-space for numerical stability, this recursion supports Forward-Backward or Viterbi decoding.

\paragraph{Trust-region rotation refinement.}
The Maximum A Posteriori (MAP) estimate derived from the filtered posterior undergoes local rotation refinement.
Keeping geometric translation and scale strictly preserved, a learning-based rotation head predicts $R_t^m$,
which is blended with the geometric
reference $R_t^g$ through:
\begin{equation}
\label{eq:refinement}
\begin{aligned}
R_t^{\mathrm{out}}
&= \arg\min_{\substack{R\in\mathrm{SO}(3)\\
    d_{\mathrm{SO}(3)}(R,R_t^g)\leq\eta_t}}
\Bigl[
    (1-\omega_t)\|R-R_t^g\|_F^2 \\[-1pt]
&\hspace{4.5em} + \omega_t\|R-R_t^m\|_F^2
\Bigr].
\end{aligned}
\end{equation}
where $\eta_t$ defines the $\mathrm{SO}(3)$ trust-region radius and $\omega_t\in[0,1]$ represents confidence.
Solved via chordal interpolation projected onto $\mathrm{SO}(3)$, this formulation prevents
learning-based predictions from diverging outside the geometrically grounded rotation basin.

\subsection{Sequence-Specific Self-Supervised Adaptation}
\label{sec:adaptation}

To adapt tracker parameters online without ground-truth pose annotations, GRC-Pose performs test-time
sequence-specific self-adaptation.  Correspondence, geometric, and dense-measurement evidence is first
materialized in a fixed recording-local candidate bank.  A temperature-scaled target posterior
$\pi_t^\star(h)$ is then formed from the distance between each retained candidate and the fixed structured
geometric parent.

Only the recording-local posterior adapter is optimized via
\begin{equation}
\label{eq:training}
\begin{aligned}
\mathcal L_{\mathrm{seq}}
={}& \mathcal L_{\mathrm{post}}
    + \lambda_g \mathcal L_{\mathrm{geom}}
    + \lambda_H \mathcal H(q_t)
    + \lambda_\delta\|\boldsymbol\delta\|_2^2.
\end{aligned}
\end{equation}
where $\mathcal L_{\mathrm{post}} = D_{\mathrm{KL}}(\pi_t^\star \| q_t)$ calibrates the posterior against the
geometric target, $\mathcal L_{\mathrm{geom}}$ keeps the bounded candidate update close to its parent, and
$\mathcal H(q_t)$ is an entropy penalty.  Correspondence and memory evidence remain fixed in the candidate
bank, while observed motion enters the posterior transport.  Full construction, loss terms, and adaptation
protocols are detailed in the supplementary material.

\FloatBarrier
\section{Experiments}
\label{sec:experiments}

\begin{figure}[tbp]
\centering
\includegraphics[width=\linewidth]{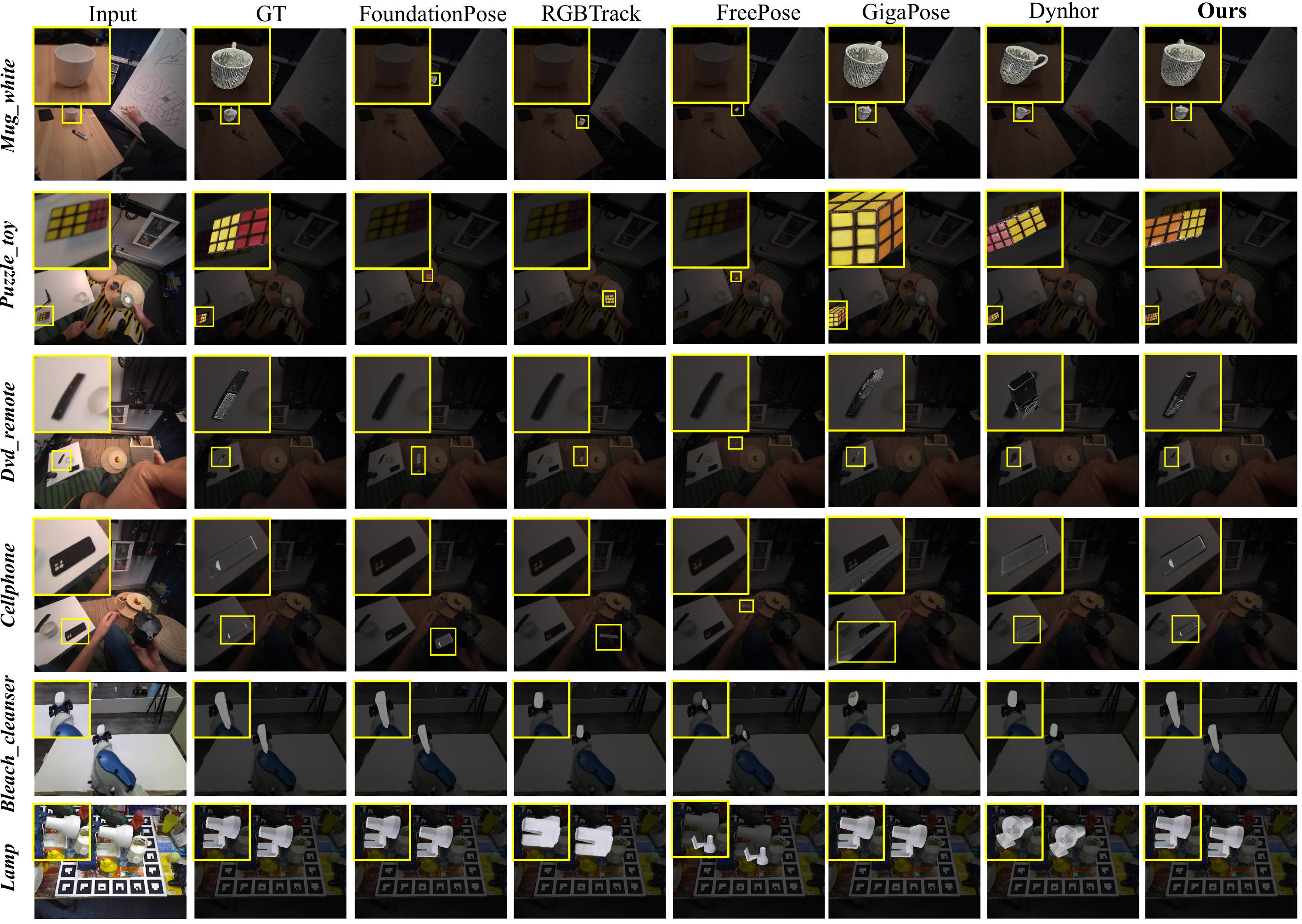}
\caption{\textbf{Qualitative comparison across tracking benchmarks.} Pose tracking results on HOT3D, YCBInEOAT and LINEMOD with SAM3D CAD.}
\label{fig:exp_hot3d_qualitative}
\end{figure}

\subsection{Experimental Setup}
\label{sec:exp_setup}

\paragraph{Datasets and protocols.}
We evaluate GRC-Pose on HOT3D~\cite{banerjee2024hot3d} for egocentric hand-object interactions and on
YCBInEOAT~\cite{wen2020se3tracknet} and LINEMOD~\cite{hinterstoisser2012linemod} for object-centric tracking.
Within each protocol, methods use shared observation sequences, camera calibrations, SAM3 object masks, and the
corresponding Metric CAD or SAM3D CAD condition; initialization-dependent baselines receive a shared
non-ground-truth first-frame pose seed.

\paragraph{Evaluation metrics.}
Following the standard BOP protocol~\cite{hodan2020bop}, we report AR (MSPD, MSSD, and VSD) and ADD/ADD-S AUC
on Metric CAD benchmarks. On HOT3D, $\mathrm{Motion}_{50,.5d}$ requires both adjacent endpoints to meet the
$50$-pixel MSPD and $0.5d$ MSSD thresholds, with camera-invariant relative pose-error drift below $30^\circ$ and
$0.5d$, where $d$ is the object diameter; complete formulations are in the supplementary material.

\subsection{Evaluation on Dynamic Egocentric Benchmark}
\label{sec:exp_hot3d}

\begin{figure}[tbp]
\centering
\includegraphics[width=0.74\linewidth]{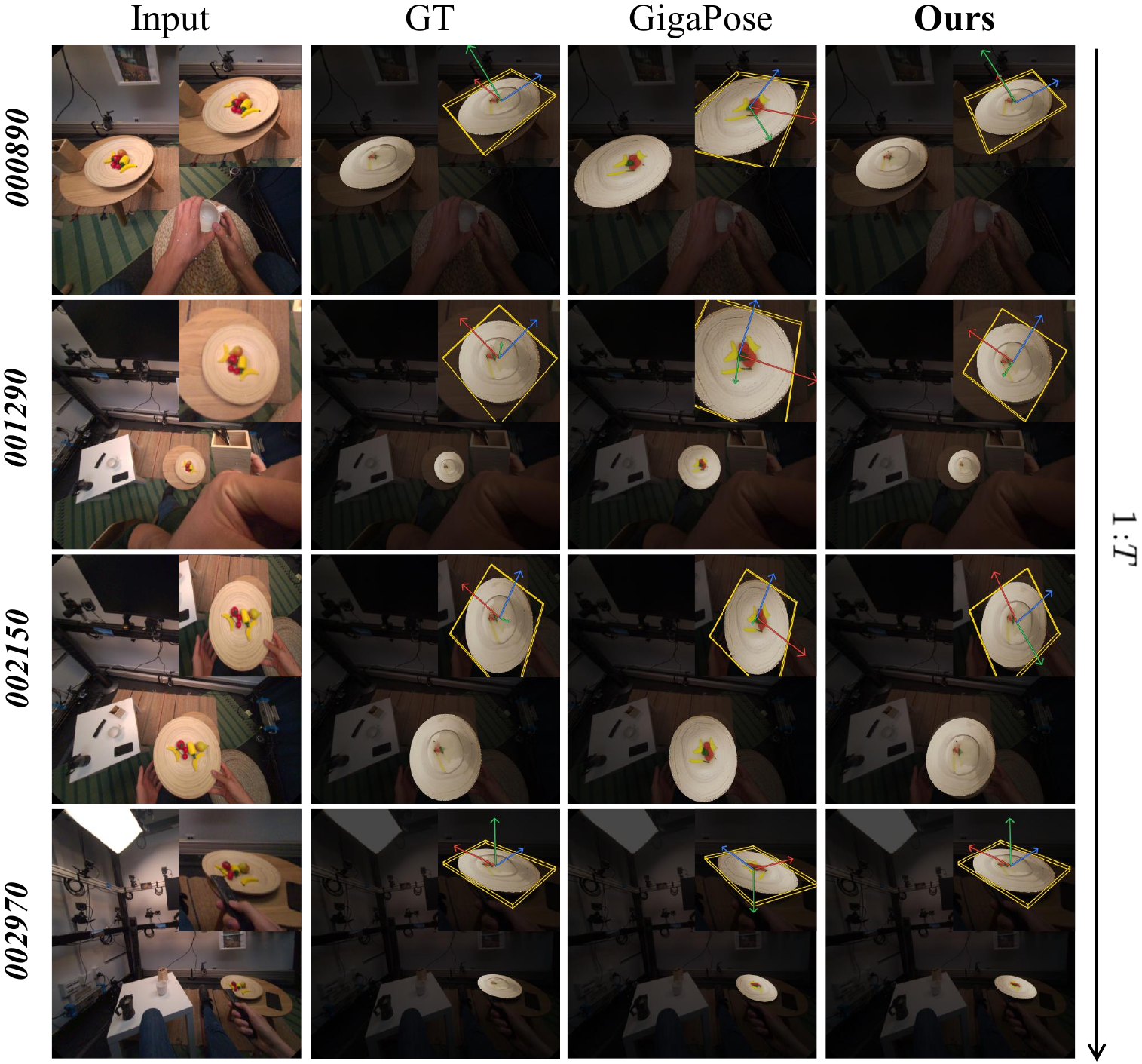}
\caption{\textbf{Qualitative temporal tracking on HOT3D.} Images are cropped around the target object with yellow 3D bounding boxes and RGB object-coordinate axes visualizing the recovered 6D pose. Columns follow chronological order.}
\label{fig:exp_hot3d_temporal}
\end{figure}

Tab.~\ref{tab:exp_main} evaluates 6D pose tracking with SAM3D CAD across the three benchmarks. GRC-Pose achieves the highest Avg.\ AR
(34.0\%), $\mathrm{AR}_{\mathrm{MSPD}}$ (79.0\%), and Motion (26.9\%), exceeding Dynhor and GigaPose
by 9.9 and 15.0 Motion points. Fig.~\ref{fig:exp_hot3d_temporal} shows the plate trajectory across
viewpoint changes and partial hand occlusion, where GRC-Pose maintains a scene-consistent sequence of pose candidates across
chronological frames. BIT reconstructs geometry from the initial observation; its 1.6\% Avg.\ AR indicates
that this reconstruction does not sustain registration through the long egocentric streams.

Per-asset evaluations on the nine assets with at least ten scored keys show GRC-Pose leading or tying the top baseline on seven of these nine, with the
largest margins on keyboard (34.2\% AR, +12.0\% over GigaPose) and eraser (42.5\% AR, +7.2\% over Dynhor).
Supplementary threshold sweeps show the Motion lead over Dynhor widening from 2.9 points at $30\text{ pixels}/0.3d$
to 9.9 points at $50\text{ pixels}/0.5d$, consistent with the temporal alignment in Fig.~\ref{fig:exp_hot3d_qualitative}.

\subsection{Evaluation on Classical Object-centric Benchmarks}
\label{sec:exp_objectcentric}
Tab.~\ref{tab:exp_main} reports the shared SAM3D CAD comparison on YCBInEOAT and LINEMOD. GRC-Pose achieves the highest macro ADD-S average (79.8\%) across all 20 object
identities and the leading LINEMOD ADD-S AUC (86.1\%); it reaches 41.2\% ADD AUC and 67.0\% ADD-S AUC on
YCBInEOAT. Across specific identities, GRC-Pose reaches 93.1\% ADD-S AUC on sugar box and leads on ape,
bowl, and camera; complete per-object rows are reported in the supplementary material.

\begin{table}[tbp]
\caption{\textbf{Main quantitative comparison with SAM3D CAD.} PF denotes prior-free; all values are percentages, and best and second-best values are bold and underlined. HOT3D uses RGB-derived VGGT-$\Omega$ proxy depth where a method consumes scene geometry. Full BOP components, Metric CAD controls, and complete per-asset/per-object results are in the supplementary material. $\ddagger$ FoundationPose uses independent register on every fixed evaluation key.}
\label{tab:exp_main}
\centering
\papertablestyle
\setlength{\tabcolsep}{3.5pt}
\begin{tabular*}{\linewidth}{@{\extracolsep{\fill}}lcccccccc@{}}
\toprule
\multirow{2}{*}{Method} & \multirow{2}{*}{PF} & \multicolumn{3}{c}{HOT3D} & \multicolumn{2}{c}{YCBInEOAT} & \multicolumn{2}{c}{LINEMOD} \\
\cmidrule(lr){3-5}\cmidrule(lr){6-7}\cmidrule(l){8-9}
& & Avg.\ AR $\uparrow$ & AR$_{\mathrm{MSPD}}$ $\uparrow$ & Motion $\uparrow$ & ADD $\uparrow$ & ADD-S $\uparrow$ & ADD $\uparrow$ & ADD-S $\uparrow$ \\
\midrule
FoundationPose$^{\ddagger}$ & $\times$ & 3.4 & 7.0 & 2.4 & \textbf{43.1} & \textbf{67.2} & 38.2 & 84.1 \\
RGBTrack & $\times$ & 3.4 & 9.2 & 1.3 & 5.6 & 21.5 & 0.4 & 2.1 \\
GigaPose & $\times$ & 31.4 & 69.2 & 11.9 & 16.1 & 31.3 & 10.4 & 37.5 \\
FreePose & $\times$ & 1.9 & 5.6 & 0.0 & 0.0 & 5.7 & 1.7 & 18.3 \\
Dynhor & $\times$ & \underline{31.9} & \underline{76.8} & \underline{17.0} & 2.0 & 25.5 & \textbf{59.9} & \underline{85.5} \\
BIT & \checkmark & 1.6 & 4.0 & 0.0 & 2.3 & 7.8 & 0.6 & 1.7 \\
\midrule
\textbf{Ours} & \checkmark & \textbf{34.0} & \textbf{79.0} & \textbf{26.9} & \underline{41.2} & \underline{67.0} & \underline{44.8} & \textbf{86.1} \\
\bottomrule
\end{tabular*}
\end{table}

\subsection{Ablation Study}
\label{sec:exp_ablation}

\begin{figure}[tbp]
\centering
\includegraphics[width=0.70\linewidth]{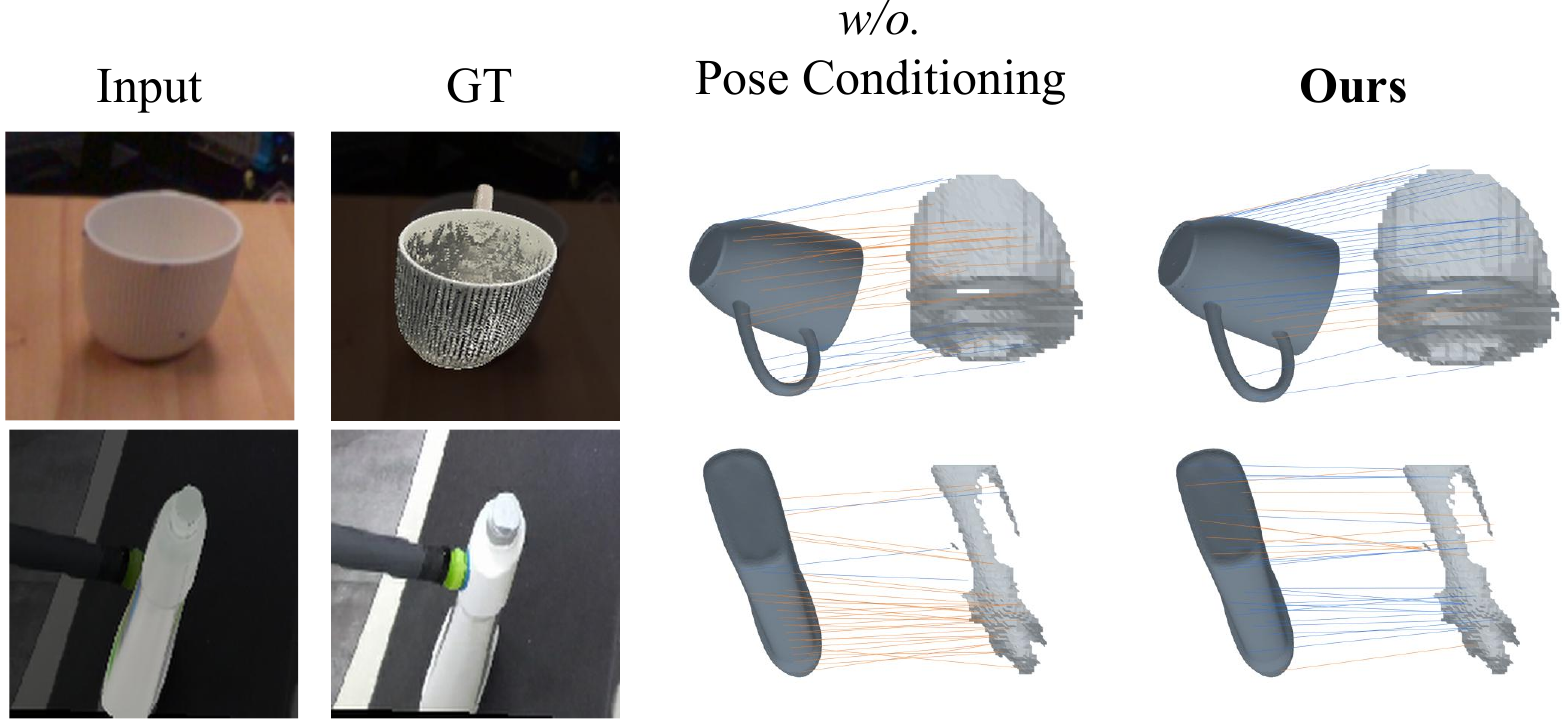}
\caption{\textbf{Qualitative visualization of pose-conditioned correspondences.} Blue and orange links denote $0.1d$ inliers and outliers.}
\label{fig:exp_pose_conditioning}
\end{figure}

Tab.~\ref{tab:exp_mechanisms}(a) isolates pose-conditioned correspondence on HOT3D with SAM3D CAD: all rows retain the
same matcher, candidate hypotheses, image and geometric features, and geometric solver. We measure hard correspondences before solver selection and temporal filtering. Candidate-pose
conditioning raises Inlier@$0.1d$ from 13.94\% to 19.35\%, reduces the 3D residual from 58.42\% to 54.78\%, and lowers
duplicate assignment from 37.04\% to 34.32\%; the contrastive loss further yields 20.34\%, 49.99\%, and 30.98\%.
Bootstrap intervals over 23 recording-asset sequences exclude zero for the full-to-base changes (Supplementary).
Fig.~\ref{fig:exp_pose_conditioning} shows the same concentration on bleach (3/32 to 22/32) and mug (9/32 to 28/32).

\begin{table}[tbp]
\caption{\textbf{Component ablations.} Both panels share the same HOT3D recording--asset streams under SAM3D CAD, adaptation seeds, and model checkpoints. All entries except Supports/frame are percentages; residuals use the shared correspondence field before and after the shared posterior decoder.}
\label{tab:exp_mechanisms}
\centering
\papertablestyle
\setlength{\tabcolsep}{2.5pt}
\begin{minipage}[t]{0.485\linewidth}
\vspace{0pt}
\begin{tabular*}{\linewidth}{@{\extracolsep{\fill}}lccc@{}}
\toprule
\multicolumn{4}{c}{\textbf{(a) Pose-conditioned correspondence}} \\
\addlinespace[3pt]
Variant & \shortstack{Inlier@\\$0.1d$ $\uparrow$} & \shortstack{3D residual\\$/d$ $\downarrow$} & \shortstack{Duplicate\\assignment $\downarrow$} \\
\midrule
w/o pose cond. & 13.94 & 58.42 & 37.04 \\
+ pose cond. & 19.35 & 54.78 & 34.32 \\
\textbf{+ contrastive} & \textbf{20.34} & \textbf{49.99} & \textbf{30.98} \\
\bottomrule
\end{tabular*}
\end{minipage}\hfill
\begin{minipage}[t]{0.485\linewidth}
\vspace{0pt}
\begin{tabular*}{\linewidth}{@{\extracolsep{\fill}}lccc@{}}
\toprule
\multicolumn{4}{c}{\textbf{(b) Solver support and selection}} \\
\addlinespace[3pt]
Variant & \shortstack{Best residual\\$/d$ $\downarrow$} & \shortstack{Selected residual\\$/d$ $\downarrow$} & \shortstack{Supports\\/ frame $\uparrow$} \\
\midrule
2D--3D only & 21.64 & 24.25 & 64.0 \\
3D--3D only & \underline{19.75} & \textbf{22.52} & 64.0 \\
Direct pose avg. & 20.48 & 22.98 & 64.0 \\
\textbf{Delayed union} & \textbf{19.74} & \underline{22.63} & \textbf{86.1} \\
\bottomrule
\end{tabular*}%
\end{minipage}
\end{table}

\begin{figure}[tbp]
\centering
\includegraphics[width=0.90\linewidth]{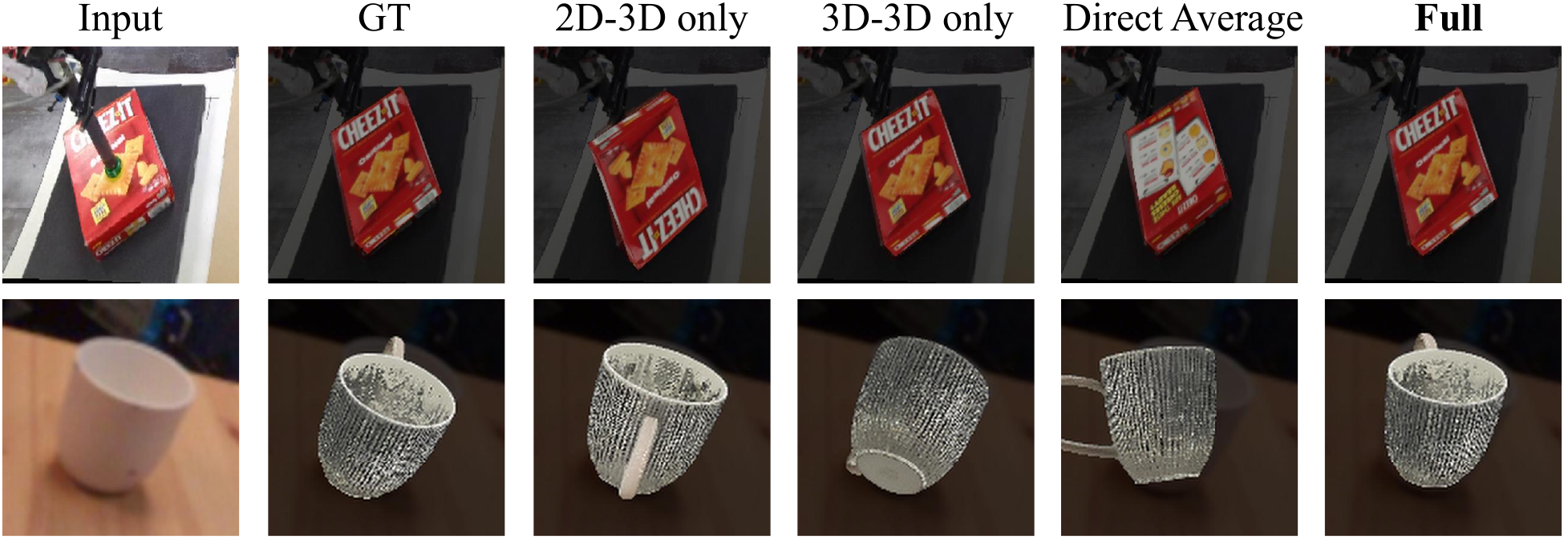}
\caption{\textbf{Qualitative comparison of geometric solving strategies.} Branch-specific supports retain distinct pose modes, whereas direct averaging produces an intermediate pose. This illustration accompanies the solver mechanism diagnostic in Tab.~\ref{tab:exp_mechanisms}.}
\label{fig:exp_solver_union}
\end{figure}

Panel (b) uses the same 23 HOT3D recording--asset streams under SAM3D CAD and three all-frame adaptation seeds. Independent 3D--3D
alignment is available for 34.47\% of parent hypotheses and differs from its paired 2D--3D parent by a median
$25.75^\circ$ and 2.47 cm. Delayed solver union retains 86.1 hypotheses per frame; posterior decoding selects
3D--3D support on 52.4\% of frames and 2D--3D support on the other 47.6\%. Direct averaging combines the 2D--3D
and 3D--3D candidates before posterior selection, whereas delayed solver union retains both. Best residual is measured over retained candidates,
whereas selected residual is measured after posterior decoding. Fig.~\ref{fig:exp_solver_union} provides
YCBInEOAT and HOT3D examples where retaining branch candidates until posterior selection avoids the intermediate pose produced by direct averaging.

Under causal online execution, unsupported updates fall back to their geometric parent candidate; full-video filtering
is in the supplementary material. The 1.27M-parameter posterior runs at 1.202 ms per frame (831.9 FPS) on one A100 GPU;
self-adaptation takes a median 27.23 seconds and reaches 95\% of its final improvement within 10.37 seconds.

\section{Conclusion}
We introduce GRC-Pose, a prior-free framework for 6D tracking through generation-reconstruction correspondence.
GeoCorr-Matcher builds object-scene support, FGH-Solver selects candidates, and posterior-gated memory retains reliable state.
GRC-Pose leads HOT3D tracking with SAM3D CAD in Average Recall and Motion retention, with competitive object-centric results.
It tracks a designated object throughout a video.

\paragraph{Use of Generative AI.} Generative AI tools assisted code development and manuscript preparation.
The authors independently verified all content and take full responsibility for this submission.

\FloatBarrier
\begingroup
\small
\setlength{\bibsep}{3pt plus 1pt}
\bibliographystyle{unsrtnat}
\bibliography{main}

@misc{simeoni2025dinov3,
  title         = {{DINOv3}},
  author        = {Sim{\'e}oni, Oriane and Vo, Huy V. and Seitzer, Maximilian and others},
  year          = {2025},
  eprint        = {2508.10104},
  archivePrefix = {arXiv}
}

@inproceedings{deng2019poserbpf,
  title     = {{PoseRBPF}: A Rao-Blackwellized Particle Filter for 6D Object Pose Tracking},
  author    = {Deng, Xinke and Mousavian, Arsalan and Xiang, Yu and Xia, Fei and Bretl, Timothy and Fox, Dieter},
  booktitle = {Robotics: Science and Systems},
  year      = {2019}
}

@inproceedings{he2022oneposepp,
  title     = {{OnePose++}: Keypoint-Free One-Shot Object Pose Estimation without {CAD} Models},
  author    = {He, Xingyi and Sun, Jiaming and Wang, Yuang and Huang, Di and Bao, Hujun and Zhou, Xiaowei},
  booktitle = {Advances in Neural Information Processing Systems},
  volume    = {35},
  year      = {2022}
}

@inproceedings{labbe2022megapose,
  title     = {{MegaPose}: 6D Pose Estimation of Novel Objects via Render \& Compare},
  author    = {Labb{\'e}, Yann and Manuelli, Lucas and Mousavian, Arsalan and Tyree, Stephen and Birchfield, Stan and Tremblay, Jonathan and Carpentier, Justin and Aubry, Mathieu and Fox, Dieter and Sivic, Josef},
  booktitle = {Conference on Robot Learning},
  year      = {2022}
}

@inproceedings{liu2022gen6d,
  title     = {{Gen6D}: Generalizable Model-Free 6-{DoF} Object Pose Estimation from {RGB} Images},
  author    = {Liu, Yuan and Wen, Yilin and Peng, Sida and Lin, Cheng and Long, Xiaoxiao and Komura, Taku and Wang, Wenping},
  booktitle = {European Conference on Computer Vision},
  pages     = {20--37},
  year      = {2022}
}

@inproceedings{ornek2024foundpose,
  title     = {{FoundPose}: Unseen Object Pose Estimation with Foundation Features},
  author    = {{\"O}rnek, Evin P{\i}nar and Labb{\'e}, Yann and Tekin, Bugra and Ma, Lingni and Keskin, Cem and Forster, Christian and Hodan, Tomas},
  booktitle = {European Conference on Computer Vision},
  year      = {2024}
}

@inproceedings{sun2022onepose,
  title     = {{OnePose}: One-Shot Object Pose Estimation without {CAD} Models},
  author    = {Sun, Jiaming and Wang, Zihao and Zhang, Siyu and He, Xingyi and Zhao, Hongcheng and Zhang, Guofeng and Zhou, Xiaowei},
  booktitle = {Proceedings of the IEEE/CVF Conference on Computer Vision and Pattern Recognition},
  pages     = {6825--6834},
  year      = {2022}
}

@inproceedings{sam3dteam2025sam3d,
  title     = {{SAM 3D}: 3Dfy Anything in Images},
  author    = {Chen, Xingyu and Chu, Fu-Jen and Gleize, Pierre and Liang, Kevin J. and others},
  booktitle = {Proceedings of the IEEE/CVF Conference on Computer Vision and Pattern Recognition},
  pages     = {7220--7232},
  year      = {2026}
}

@inproceedings{wang2026vggtomega,
  title     = {{VGGT}-$\Omega$},
  author    = {Wang, Jianyuan and Chen, Minghao and Zhang, Shangzhan and Karaev, Nikita and Sch{\"o}nberger, Johannes and Labatut, Patrick and Bojanowski, Piotr and Novotny, David and Vedaldi, Andrea and Rupprecht, Christian},
  booktitle = {Proceedings of the IEEE/CVF Conference on Computer Vision and Pattern Recognition (CVPR)},
  year      = {2026}
}

@inproceedings{jiang2025hand,
  title     = {Hand-held Object Reconstruction from RGB Video with Dynamic Interaction},
  author    = {Jiang, Shijian and Ye, Qi and Xie, Rengan and Huo, Yuchi and Chen, Jiming},
  booktitle = {Proceedings of the IEEE/CVF Conference on Computer Vision and Pattern Recognition},
  pages     = {12220--12230},
  year      = {2025}
}

@article{lepetit2009epnp,
  title   = {{EPnP}: An Accurate {O}(n) Solution to the {PnP} Problem},
  author  = {Lepetit, Vincent and Moreno-Noguer, Francesc and Fua, Pascal},
  journal = {International Journal of Computer Vision},
  volume  = {81},
  number  = {2},
  pages   = {155--166},
  year    = {2009}
}

@article{umeyama1991least,
  title   = {Least-Squares Estimation of Transformation Parameters Between Two Point Patterns},
  author  = {Umeyama, Shinji},
  journal = {IEEE Transactions on Pattern Analysis and Machine Intelligence},
  volume  = {13},
  number  = {4},
  pages   = {376--380},
  year    = {1991}
}

@inproceedings{qin2022geotransformer,
  title     = {Geometric Transformer for Fast and Robust Point Cloud Registration},
  author    = {Qin, Zheng and Yu, Hao and Wang, Changjian and Guo, Yulan and Peng, Yuxing and Xu, Kai},
  booktitle = {Proceedings of the IEEE/CVF Conference on Computer Vision and Pattern Recognition},
  pages     = {11143--11152},
  year      = {2022}
}

@inproceedings{yew2022regtr,
  title     = {{REGTR}: End-to-End Point Cloud Correspondences With Transformers},
  author    = {Yew, Zi Jian and Lee, Gim Hee},
  booktitle = {Proceedings of the IEEE/CVF Conference on Computer Vision and Pattern Recognition},
  pages     = {6677--6686},
  year      = {2022}
}

@inproceedings{wen2024foundationpose,
  title     = {FoundationPose: Unified 6D Pose Estimation and Tracking of Novel Objects},
  author    = {Wen, Bowen and Yang, Wei and Kautz, Jan and Birchfield, Stan},
  booktitle = {Proceedings of the IEEE/CVF Conference on Computer Vision and Pattern Recognition},
  pages     = {17868--17879},
  year      = {2024}
}

@inproceedings{hodan2020bop,
  title     = {{BOP} Challenge 2020 on 6D Object Localization},
  author    = {Hodan, Tomas and Sundermeyer, Martin and Drost, Bertram and Labbe, Yann and Brachmann, Eric and Michel, Frank and Rother, Carsten and Matas, Jiri},
  booktitle = {Computer Vision -- ECCV 2020 Workshops},
  pages     = {577--594},
  year      = {2020},
  doi       = {10.1007/978-3-030-66096-3_39}
}

@misc{banerjee2024hot3d,
  title         = {Introducing {HOT3D}: An Egocentric Dataset for 3D Hand and Object Tracking},
  author        = {Banerjee, Prithviraj and Shkodrani, Sindi and Moulon, Pierre and Hampali, Shreyas and Zhang, Fan and Fountain, Jade and Miller, Edward and Basol, Selen and Newcombe, Richard and Wang, Robert and Engel, Jakob Julian and Hodan, Tomas},
  year          = {2024},
  eprint        = {2406.09598},
  archivePrefix = {arXiv}
}

@inproceedings{fan2024egochallenge,
  title     = {Benchmarks and Challenges in Pose Estimation for Egocentric Hand Interactions with Objects},
  author    = {Fan, Zicong and Ohkawa, Takehiko and Yang, Linlin and Lin, Nie and Zhou, Zhishan and others},
  booktitle = {European Conference on Computer Vision},
  pages     = {428--448},
  year      = {2024}
}

@inproceedings{hinterstoisser2012linemod,
  title     = {Model Based Training, Detection and Pose Estimation of Texture-Less {3D} Objects in Heavily Cluttered Scenes},
  author    = {Hinterstoisser, Stefan and Lepetit, Vincent and Ilic, Slobodan and Holzer, Stefan and Bradski, Gary and Konolige, Kurt and Navab, Nassir},
  booktitle = {Asian Conference on Computer Vision},
  pages     = {548--562},
  year      = {2012}
}

@inproceedings{wen2020se3tracknet,
  title     = {{se(3)-TrackNet}: Data-driven 6D Pose Tracking by Calibrating Image Residuals in Synthetic Domains},
  author    = {Wen, Bowen and Bekris, Kostas},
  booktitle = {IEEE/RSJ International Conference on Intelligent Robots and Systems},
  year      = {2020}
}

@inproceedings{nguyen2024gigapose,
  title     = {GigaPose: Fast and Robust Novel Object Pose Estimation via One Correspondence},
  author    = {Nguyen, Van Nguyen and Groueix, Thibault and Salzmann, Mathieu and Lepetit, Vincent},
  booktitle = {Proceedings of the IEEE/CVF Conference on Computer Vision and Pattern Recognition},
  pages     = {9903--9913},
  year      = {2024}
}

@inproceedings{guo2025rgbtrack,
  title     = {{RGBTrack}: Fast, Robust Depth-Free 6D Pose Estimation and Tracking},
  author    = {Guo, Teng and Yu, Jingjin},
  booktitle = {2025 IEEE/RSJ International Conference on Intelligent Robots and Systems (IROS)},
  pages     = {18872--18879},
  year      = {2025},
  doi       = {10.1109/IROS60139.2025.11247381}
}

@inproceedings{ponimatkin2025freepose,
  title     = {6D Object Pose Tracking in Internet Videos for Robotic Manipulation},
  author    = {Ponimatkin, Georgy and Cifka, Martin and Soucek, Tomas and Fourmy, Mederic and Labbe, Yann and Petrik, Vladimir and Sivic, Josef},
  booktitle = {International Conference on Learning Representations},
  year      = {2025}
}

@article{sahin2020review,
  title   = {A Review on Object Pose Recovery: From 3D Bounding Box Detectors to Full 6D Pose Estimators},
  author  = {Sahin, Caner and Garcia-Hernando, Guillermo and Sock, Juil and Kim, Tae-Kyun},
  journal = {Image and Vision Computing},
  volume  = {96},
  pages   = {103898},
  year    = {2020},
  doi     = {10.1016/j.imavis.2020.103898}
}

@inproceedings{haugaard2022surfemb,
  title     = {{SurfEmb}: Dense and Continuous Correspondence Distributions for Object Pose Estimation with Learnt Surface Embeddings},
  author    = {Haugaard, Rasmus Laurvig and Buch, Anders Glent},
  booktitle = {Proceedings of the IEEE/CVF Conference on Computer Vision and Pattern Recognition},
  pages     = {6749--6758},
  year      = {2022}
}

@article{chen2024epropnp,
  title   = {{EPro-PnP}: Generalized End-to-End Probabilistic Perspective-n-Points for Monocular Object Pose Estimation},
  author  = {Chen, Hansheng and Tian, Wei and Wang, Pichao and Wang, Fan and Xiong, Lu and Li, Hao},
  journal = {IEEE Transactions on Pattern Analysis and Machine Intelligence},
  year    = {2024},
  doi     = {10.1109/TPAMI.2024.3354997}
}

@inproceedings{sun2021loftr,
  title     = {{LoFTR}: Detector-Free Local Feature Matching with Transformers},
  author    = {Sun, Jiaming and Shen, Zehong and Wang, Yuang and Bao, Hujun and Zhou, Xiaowei},
  booktitle = {Proceedings of the IEEE/CVF Conference on Computer Vision and Pattern Recognition},
  pages     = {8922--8931},
  year      = {2021}
}

@inproceedings{sarlin2020superglue,
  title     = {{SuperGlue}: Learning Feature Matching with Graph Neural Networks},
  author    = {Sarlin, Paul-Edouard and DeTone, Daniel and Malisiewicz, Tomasz and Rabinovich, Andrew},
  booktitle = {Proceedings of the IEEE/CVF Conference on Computer Vision and Pattern Recognition},
  pages     = {4938--4947},
  year      = {2020}
}

@inproceedings{wen2021bundletrack,
  title     = {{BundleTrack}: 6D Pose Tracking for Novel Objects without Instance or Category-Level 3D Models},
  author    = {Wen, Bowen and Bekris, Kostas},
  booktitle = {IEEE/RSJ International Conference on Intelligent Robots and Systems},
  pages     = {8067--8074},
  year      = {2021},
  doi       = {10.1109/IROS51168.2021.9635991}
}

@inproceedings{wen2023bundlesdf,
  title     = {{BundleSDF}: Neural 6-{DoF} Tracking and 3D Reconstruction of Unknown Objects},
  author    = {Wen, Bowen and Tremblay, Jonathan and Blukis, Valts and Tyree, Stephen and M{\"u}ller, Thomas and Evans, Alex and Fox, Dieter and Kautz, Jan and Birchfield, Stan},
  booktitle = {Proceedings of the IEEE/CVF Conference on Computer Vision and Pattern Recognition},
  pages     = {606--617},
  year      = {2023}
}

@inproceedings{fan2024hold,
  title     = {{HOLD}: Category-agnostic 3D Reconstruction of Interacting Hands and Objects from Video},
  author    = {Fan, Zicong and Parelli, Maria and Kadoglou, Maria Eleni and Chen, Xu and Kocabas, Muhammed and Black, Michael J. and Hilliges, Otmar},
  booktitle = {Proceedings of the IEEE/CVF Conference on Computer Vision and Pattern Recognition},
  pages     = {494--504},
  year      = {2024}
}

@inproceedings{hodan2020epos,
  title     = {{EPOS}: Estimating 6D Pose of Objects with Symmetries},
  author    = {Hodan, Tomas and Barath, Daniel and Matas, Jiri},
  booktitle = {Proceedings of the IEEE/CVF Conference on Computer Vision and Pattern Recognition},
  pages     = {11703--11712},
  year      = {2020}
}

@inproceedings{song2025bit,
  title     = {Prior-Free 3D Object Tracking},
  author    = {Song, Xiuqiang and Jin, Li and Zhang, Zhengxian and Li, Jiachen and Zhong, Fan and Zhang, Guofeng and Qin, Xueying},
  booktitle = {Proceedings of the IEEE/CVF Conference on Computer Vision and Pattern Recognition},
  pages     = {1200--1209},
  year      = {2025}
}

@inproceedings{li2025uapose,
  title     = {{UA-Pose}: Uncertainty-Aware 6D Object Pose Estimation and Online Object Completion with Partial References},
  author    = {Li, Ming-Feng and Yang, Xin and Wang, Fu-En and Basak, Hritam and Sun, Yuyin and Gayaka, Shreekant and Sun, Min and Kuo, Cheng-Hao},
  booktitle = {Proceedings of the IEEE/CVF Conference on Computer Vision and Pattern Recognition},
  pages     = {1180--1189},
  year      = {2025}
}

@misc{taher2025kvtracker,
  title         = {{KV-Tracker}: Real-Time Pose Tracking with Transformers},
  author        = {Taher, Marwan and Alzugaray, Ignacio and Mazur, Kirill and Kong, Xin and Davison, Andrew J.},
  year          = {2025},
  eprint        = {2512.22581},
  archivePrefix = {arXiv}
}

@misc{chen2024gsgtrack,
  title         = {{GSGTrack}: Gaussian Splatting-Guided Object Pose Tracking from {RGB} Videos},
  author        = {Chen, Zhiyuan and Lu, Fan and Yu, Guo and Li, Bin and Qu, Sanqing and Huang, Yuan and Fu, Changhong and Chen, Guang},
  year          = {2024},
  eprint        = {2412.02267},
  archivePrefix = {arXiv}
}

@misc{luo2024sgpose,
  title         = {Object Gaussian for Monocular 6D Pose Estimation from Sparse Views},
  author        = {Luo, Luqing and Sun, Shichu and Yang, Jiangang and Zheng, Linfang and Du, Jinwei and Liu, Jian},
  year          = {2024},
  eprint        = {2409.02581},
  archivePrefix = {arXiv}
}

@inproceedings{wang2024dustr,
  title     = {{DUSt3R}: Geometric 3D Vision Made Easy},
  author    = {Wang, Shuzhe and Leroy, Vincent and Cabon, Yohann and Chidlovskii, Boris and Revaud, J{\'e}r{\^o}me},
  booktitle = {Proceedings of the IEEE/CVF Conference on Computer Vision and Pattern Recognition},
  year      = {2024}
}

@inproceedings{leroy2024mast3r,
  title     = {Grounding Image Matching in 3D with {MASt3R}},
  author    = {Leroy, Vincent and Cabon, Yohann and Revaud, J{\'e}r{\^o}me},
  booktitle = {European Conference on Computer Vision},
  year      = {2024}
}

@inproceedings{wang2025vggt,
  title     = {{VGGT}: Visual Geometry Grounded Transformer},
  author    = {Wang, Jianyuan and Chen, Minghao and Karaev, Nikita and Vedaldi, Andrea and Rupprecht, Christian and Novotny, David},
  booktitle = {Proceedings of the IEEE/CVF Conference on Computer Vision and Pattern Recognition},
  pages     = {5294--5306},
  year      = {2025}
}

@inproceedings{xiang2024trellis,
  title     = {Structured 3D Latents for Scalable and Versatile 3D Generation},
  author    = {Xiang, Jianfeng and Lv, Zelong and Xu, Sicheng and Deng, Yu and Wang, Ruicheng and Zhang, Bowen and Chen, Dong and Tong, Xin and Yang, Jiaolong},
  booktitle = {Proceedings of the IEEE/CVF Conference on Computer Vision and Pattern Recognition},
  pages     = {21469--21480},
  year      = {2025}
}

@misc{nguyen2025gotrack,
  title         = {{GoTrack}: Generic 6DoF Object Pose Refinement and Tracking},
  author        = {Nguyen, Van Nguyen and Forster, Christian and Shkodrani, Sindi and Lepetit, Vincent and Tekin, Bugra and Keskin, Cem and Hodan, Tomas},
  year          = {2025},
  eprint        = {2506.07155},
  archivePrefix = {arXiv}
}

@inproceedings{jin2025sixdopegs,
  title     = {{6DOPE-GS}: Online 6D Object Pose Estimation using Gaussian Splatting},
  author    = {Jin, Yufeng and Prasad, Vignesh and Jauhri, Snehal and Franzius, Mathias and Chalvatzaki, Georgia},
  booktitle = {Proceedings of the IEEE/CVF International Conference on Computer Vision},
  year      = {2025}
}

@inproceedings{milano2024neusurfemb,
  title     = {{NeuSurfEmb}: A Complete Pipeline for Dense Correspondence-based 6D Object Pose Estimation without {CAD} Models},
  author    = {Milano, Francesco and Chung, Jen Jen and Blum, Hermann and Siegwart, Roland and Ott, Lionel},
  booktitle = {2024 IEEE/RSJ International Conference on Intelligent Robots and Systems (IROS)},
  pages     = {8882--8889},
  year      = {2024},
  doi       = {10.1109/IROS58592.2024.10801399}
}

@inproceedings{lee2025any6d,
  title     = {{Any6D}: Model-free 6D Pose Estimation of Novel Objects},
  author    = {Lee, Taeyeop and Wen, Bowen and Kang, Minjun and Kang, Gyuree and Kweon, In So and Yoon, Kuk-Jin},
  booktitle = {Proceedings of the IEEE/CVF Conference on Computer Vision and Pattern Recognition},
  pages     = {11633--11643},
  year      = {2025}
}

@inproceedings{lin2024sam6d,
  title     = {{SAM-6D}: Segment Anything Model Meets Zero-Shot 6D Object Pose Estimation},
  author    = {Lin, Jiehong and Liu, Lihua and Lu, Dekun and Jia, Kui},
  booktitle = {Proceedings of the IEEE/CVF Conference on Computer Vision and Pattern Recognition},
  pages     = {27906--27916},
  year      = {2024}
}

@misc{caraffa2023zero6d,
  title         = {Object 6D Pose Estimation Meets Zero-Shot Learning},
  author        = {Caraffa, Andrea and Boscaini, Davide and Hamza, Amir and Poiesi, Fabio},
  year          = {2023},
  eprint        = {2312.00947},
  archivePrefix = {arXiv}
}

@inproceedings{caraffa2024freeze,
  title     = {{FreeZe}: Training-Free Zero-Shot 6D Pose Estimation with Geometric and Vision Foundation Models},
  author    = {Caraffa, Andrea and Boscaini, Davide and Hamza, Amir and Poiesi, Fabio},
  booktitle = {European Conference on Computer Vision},
  year      = {2024}
}

@misc{caraffa2025freezev2,
  title         = {Accurate and Efficient Zero-Shot 6D Pose Estimation with Frozen Foundation Models},
  author        = {Caraffa, Andrea and Boscaini, Davide and Poiesi, Fabio},
  year          = {2025},
  eprint        = {2506.09784},
  archivePrefix = {arXiv}
}

@inproceedings{huang2024matchu,
  title     = {{MatchU}: Matching Unseen Objects for 6D Pose Estimation from {RGB-D} Images},
  author    = {Huang, Junwen and Yu, Hao and Yu, Kuan-Ting and Navab, Nassir and Ilic, Slobodan and Busam, Benjamin},
  booktitle = {Proceedings of the IEEE/CVF Conference on Computer Vision and Pattern Recognition (CVPR)},
  month     = {June},
  pages     = {10095--10105},
  year      = {2024}
}

@article{liu2026sinref6d,
  title   = {Scalable Unseen Objects 6-DoF Absolute Pose Estimation with Robotic Integration},
  author  = {Liu, Jian and Sun, Wei and Zeng, Kai and Zheng, Jin and Yang, Hui and Rahmani, Hossein and Mian, Ajmal and Wang, Lin},
  journal = {IEEE Transactions on Robotics},
  year    = {2026},
  eprint  = {2503.05578},
  archivePrefix = {arXiv}
}
\endgroup

\clearpage
\appendix
\setcounter{table}{0}
\setcounter{figure}{0}
\setcounter{equation}{0}
\renewcommand{\thetable}{A\arabic{table}}
\renewcommand{\thefigure}{A\arabic{figure}}
\renewcommand{\theequation}{A\arabic{equation}}
\renewcommand{\theHtable}{appendix.\arabic{table}}
\renewcommand{\theHfigure}{appendix.\arabic{figure}}
\renewcommand{\theHequation}{appendix.\arabic{equation}}

\newcommand{\suppTableFont}{\small}
\newsavebox{\suppTableBox}
\newcommand{\suppfit}[2]{%
  \sbox{\suppTableBox}{#2}%
  \ifdim\wd\suppTableBox>#1
    \resizebox{#1}{!}{\usebox{\suppTableBox}}%
  \else
    \usebox{\suppTableBox}%
  \fi
}
\setlength{\LTcapwidth}{\textwidth}
\makeatletter
\patchcmd{\LT@makecaption}{\reset@font}{\reset@font\normalsize}{}{}
\patchcmd{\LT@makecaption}{#2: }{#2. }{}{}
\patchcmd{\LT@makecaption}{#2: }{#2. }{}{}
\makeatother

\section{Method Details and Implementation}
This document reports locked implementation settings and the protocol,
breakdown, control, and qualitative results omitted from the main paper.
\subsection{Executable Implementation}

This section records the tensor interfaces, numerical settings, and
optimization boundaries used by the released implementation.

\paragraph{Stored inputs and coordinate chart.}
For a SAM3D CAD surface point $\widetilde{\mathbf x}_i$, onboarding writes the
object point used by all later stages as
\begin{equation}
\begin{aligned}
\mathbf x_i&\leftarrow s_{\mathrm{on}}R_{\mathrm{on}}
\widetilde{\mathbf x}_i+\mathbf t_{\mathrm{on}},\\
\boldsymbol\chi_{t,h}&=\left[\operatorname{vec}_6(R_{t,h}),
\frac{\mathbf t_{t,h}}{d_o},\log s_{t,h}\right]\in\R^{10}.
\end{aligned}
\label{eq:supp_implementation_chart}
\end{equation}
The implementation removes non-finite vertices and invalid faces before this
step, retains the SAM3D CAD triangle topology when available, and otherwise builds
a convex-hull surface.  Metric CAD controls read the dataset object chart
directly and set $s_{t,h}=1$.  DINOv3 ViT-B/16 descriptors, camera calibration,
scene points, and all candidate rows are materialized before the
recording-local adapter is initialized.

\paragraph{Frozen support and stored candidate evidence.}
We use DINOv3 ViT-B/16 descriptors: four intermediate feature maps are
$\ell_2$-normalized and concatenated from $512\times512$ mask-centered crops
with $1.35\times$ enlargement.  These descriptors and the calibrated object
surface receive no adapter gradient.  We retain $240$ global proposals per
frame before local matching and NMS, then keep $K=64$ rows for HOT3D and
LINEMOD with SAM3D CAD and $K=128$ for the remaining controls.  RGB-D controls require
at least $8$ matches and $6$ inliers.  RANSAC-PnP uses a $5$-pixel threshold,
$512$ iterations, SQPnP, and $30$ LM refinement steps; weighted Kabsch uses
$192$ iterations and a $2$ cm inlier threshold ($1.5$ cm in the Kabsch-only
control). HOT3D uses RGB-derived proxy geometry rather than measured depth;
its matched 2D--3D and 3D--3D support controls are specified in
Table~\ref{tab:supp_solver_matched}.

Candidate construction writes $\mathbf A\in\R^{T\times K\times17}$,
$\mathbf G\in\R^{T\times K\times4\times4}$, and
$\mathbf V\in\{0,1\}^{T\times K}$ for the score/geometry summary, pose
matrix, and validity mask.  For every frame, rows are first grouped by their
stored rank.  If a rank contains both a structured parent and a validated
solver descendant, the latter is retained; otherwise the row with the larger
stored score is retained.  The resulting top-$K$ rows form the immutable
support used by adaptation.  Its 17-dimensional evidence vector is
\begin{equation}
\begin{aligned}
\mathbf a_{t,h}=[&s_{\mathrm{fused}},s_{\mathrm{can}},s_{\mathrm{native}},s_{\mathrm{prop}},\\
&\rho_{\mathrm{inl}},\bar c_{\mathrm{DINO}},F_{1}^{\mathrm{mask}},
\epsilon_{\mathrm{depth}},\\
&\Delta_{\mathrm{box}},\rho_{\mathrm{LoFTR}},\epsilon_{\mathrm{LoFTR}},
E_{\mathrm{solve}},\\
&m_{\mathrm{match}},H_{\mathrm{match}},o_{\mathrm{solve}},
\Delta_{\mathrm{heldout}},r]_{t,h}\in\R^{17}.
\end{aligned}
\label{eq:supp_candidate_evidence}
\end{equation}
where the entries respectively record the fused, canonical-prior, native, and
proposal scores; RANSAC inlier ratio; mean DINO similarity; mask F1; normalized
depth residual; image-center displacement; LoFTR inlier ratio and residual;
and the solver-coupled energy, match mass, match entropy, observability,
held-out improvement, and rank.  Rank is divided by $K$, image-center distance
by $1000$ px, and LoFTR residual by $0.03$; non-finite entries are zeroed.
Thus correspondence, appearance, and dense-measurement evidence are fixed in
the candidate rows before adaptation; the v2.4 adapter optimizes the
recording-local posterior over this fixed bank while retaining the upstream
matcher, geometric solvers, and canonical memory.
GeoCorr-Matcher controls use $64$ canonical surface slots with aggregation momentum
$0.95$.

\paragraph{Posterior adapter.}
The candidate input
$[\mathbf A,\boldsymbol\chi]\in\R^{T\times K\times27}$ is mapped by
$\operatorname{LN}\rightarrow\operatorname{Linear}(27,128)
\rightarrow\operatorname{GELU}\rightarrow\operatorname{Linear}(128,128)$.
Three pre-norm Transformer encoder blocks operate on $[T,K,128]$ with four
attention heads, hidden width $128$, feed-forward width $512$, and dropout
$0.05$.  Softmax frame pooling produces $[T,128]$ tokens.  The temporal input
is $[\tau,\sin(2^b\pi\tau),\cos(2^b\pi\tau)]_{b=0}^{3}\in\R^9$, projected
to $128$ dimensions and processed by three blocks with the same settings.

Frame-pooling and emission heads are
$\operatorname{LN}(128)\rightarrow128\rightarrow\operatorname{GELU}
\rightarrow1$.  The pose-update head outputs six Lie coordinates, while the
reliability head is
$\operatorname{LN}(128)\rightarrow128\rightarrow\operatorname{GELU}
\rightarrow64\rightarrow\operatorname{GELU}\rightarrow3$ for
observability, temperature, and rotation noise.  The final emission and
pose-update layers are zero initialized.  Relative motion
$\widehat R\in\R^{(T-1)\times3\times3}$,
$\widehat{\mathbf t}\in\R^{(T-1)\times3}$, and an edge-validity mask feed the
log-domain recursion; a missing edge is stored as $(I_3,\mathbf0,0)$ and
closes the transport gate.  The recursion uses rotation cost plus $0.02$ times
normalized translation cost, transport gain $1.0$, temperature floor $0.05$,
and rotation-noise floor $0.03$; the Lie update is clipped to $8^{\circ}$ and
$1.5$ cm.

\subsection{Sequence-Specific Self-Supervised Adaptation}
\label{sec:supp_adaptation_implementation}

\paragraph{Recording-local inputs and geometric target.}
A new adapter parameter set $\Theta$ is initialized for each recording--asset.
The runner loads candidate rows from the stored solver-coupled candidate bank,
the parent $\bar{\G}_t$ from the preceding structured geometric stage, and
the observed relative-motion cache from the scene-alignment stage.  These three
inputs are fixed throughout the $600$ adaptation steps.  In particular,
$\bar{\G}_t$ is a preceding geometric output rather than an annotated pose.
For a retained candidate $\mathbf G_{t,h}$, the geometric E-step target is
\begin{equation}
\begin{aligned}
D_{t,h}&=d_{\SO}(R_{t,h},\bar R_t)+
\frac{\|\mathbf t_{t,h}-\bar{\mathbf t}_t\|_2}{d_o},\\
\pi_t^\star(h)&=\operatorname{stopgrad}\!\left[
\operatorname{softmax}_h(-D_{t,h}/0.01)\right].
\end{aligned}
\label{eq:supp_adapter}
\end{equation}
The rotation distance removes the Sim(3) scale of each $3\times3$ block before
computing the $\SO$ geodesic.  Consequently, $\pi_t^\star$ expresses which
stored candidates agree with the fixed geometric parent; it is not rebuilt
from annotations or from a second correspondence pass during adaptation.

\paragraph{Adapted posterior and motion transport.}
The candidate input is the concatenation of the frozen evidence and its pose
chart, $[\mathbf a_{t,h},\boldsymbol\chi_{t,h}]\in\R^{27}$.  The architecture
is specified above; here we expand the filtering notation in
Eq.~\ref{eq:filter}. Let
$\ell_{t,h}=s_{t,h}^{\mathrm{fused}}+E_\Theta(\mathbf z_{t,h})$ be the
frozen fused score plus the learned emission.  With candidate-pair transition
cost
\begin{equation}
\begin{aligned}
\Delta R_t(i,h)&=R_{t,h}R_{t-1,i}^{\top},\\
\Delta\mathbf t_t(i,h)&=\mathbf t_{t,h}-\Delta R_t(i,h)\mathbf t_{t-1,i},\\
C_t(i,h)&=d_{\SO}\!\left(\Delta R_t(i,h),\widehat R_{t-1\rightarrow t}\right)^2\\
&\quad+0.02\min\!\left\{
\frac{\|\Delta\mathbf t_t(i,h)-\widehat{\mathbf t}_{t-1\rightarrow t}\|_2^2}{d_o^2},9\right\},\\
\mathsf P_{t-1}(h\mid i)&=\operatorname{softmax}_{h}
\!\left[-C_t(i,h)/\sigma_{R,t}^{2}\right],\\
r_t(h)&=\operatorname{logsumexp}_{i}\!\left[
\log q_{t-1}(i)+\log\mathsf P_{t-1}(h\mid i)\right],\\
\log q_t(h)&=\operatorname{logsoftmax}_{h}\!\left[
o_t\ell_{t,h}/\tau_t+v_t r_t(h)\right].
\end{aligned}
\label{eq:supp_transport}
\end{equation}
where $o_t$, $\tau_t$, and $\sigma_{R,t}$ are the learned observability,
temperature, and rotation-noise outputs, respectively.  A motion edge is
valid only at cached quality at least $0.1$; otherwise it is represented by
$(I_3,\mathbf0)$ with $v_t=0$, which closes the transport term.  The
implementation uses transport gain $1.0$, $\tau_t\geq0.05$, and
$\sigma_{R,t}\geq0.03$.  The first frame uses only its emission.

\paragraph{Objective and optimization boundary.}
The fixed target and learned posterior define
\begin{equation}
\begin{aligned}
\mathcal L_{\mathrm{seq}}={}&D_{\mathrm{KL}}(\pi_t^\star\|q_t)
+20\mathcal L_{\mathrm{geom}}\\
&+0.002\mathcal H(q_t)+0.001\|\boldsymbol\delta\|_2^2,\\
\mathcal L_{\mathrm{geom}}={}&\frac{1}{T}\sum_t\sum_h\pi_t^\star(h)
\frac{\|\mathbf G^{\mathrm{upd}}_{t,h}-\bar{\G}_t\|_1}{16},\\
\mathcal H(q_t)={}&-\frac{1}{T}\sum_t\sum_hq_t(h)\log q_t(h).
\end{aligned}
\label{eq:supp_adapter_objective}
\end{equation}
where $\mathcal L_{\mathrm{geom}}$ is the $\pi_t^\star$-weighted elementwise
$\ell_1$ distance from the bounded Lie-updated candidate to its geometric
parent.  The positive entropy coefficient is an entropy penalty, and the last
term constrains the six-dimensional Lie update.  AdamW updates only $\Theta$
for $600$ complete-sequence steps with learning rate $2\times10^{-4}$, weight
decay $10^{-4}$, gradient clipping $1.0$, and seed $41$; decoding uses the
lowest-total-loss checkpoint. These settings describe the full-video
adaptation control. The all-frame protocol uses the full recording,
whereas the causal and cross-fit controls mask or withhold the scored temporal
chunk during adaptation. The online control uses prefix-causal adaptation;
its scored frames and initialization are specified in
Table~\ref{tab:supp_causal}. Candidate rows, the parent, DINO descriptors,
object surface, observed motion, and canonical memory remain frozen.

\paragraph{Bounded output fusion.}
For $h_t^*=\arg\max_hq_t(h)$, the raw adapter output applies a Lie residual
bounded to $8^\circ$ and $1.5$ cm to $\mathbf G_{t,h_t^*}$.  The released
output retains the parent translation and scale and fuses only rotation:
\begin{equation}
\begin{aligned}
\iota_t&=\sqrt{o_t\bigl(1-\widetilde{\mathcal H}(q_t)\bigr)},\\
\eta_t&=\sqrt{\sum_h\pi_t^\star(h)
d_{\SO}(R_{t,h},\bar R_t)^2},\\
\omega_t&=\iota_t\min\!\left\{1,
\frac{\eta_t}{d_{\SO}(R^{\mathrm{upd}}_t,\bar R_t)}\right\},\\
R_t^{\mathrm{mix}}&=(1-\omega_t)\bar R_t+\omega_tR_t^{\mathrm{upd}},\\
R_t^{\mathrm{out}}&=\operatorname{Proj}_{\SO}(R_t^{\mathrm{mix}}),\\
(\mathbf t_t^{\mathrm{out}},s_t^{\mathrm{out}})&=
(\bar{\mathbf t}_t,\bar s_t).
\end{aligned}
\label{eq:supp_adapter_fusion}
\end{equation}
where $\widetilde{\mathcal H}$ is entropy normalized by $\log K$.  The fusion
weight is zero when the support radius is below the lane-specific minimum or
when the parent lies outside the retained support.  Thus the adapted branch
can change rotation only inside the candidate-supported neighborhood; the
geometric parent is returned when that condition is not met.

\paragraph{Dense measurement and bounded refinement.}

For the selected pose, the GPU rasterizer minimizes
\begin{equation}
\begin{aligned}
\ell_{\mathrm{meas}}={}&0.05\ell_{\mathrm{sil}}+3\ell_{\mathrm{depth}}
+3\ell_{\mathrm{p2pl}}+0.15\ell_{\mathrm{rgb}},\\
\mathcal L_{\mathrm{refine}}={}&\ell_{\mathrm{meas}}
+0.1\ell_{\Delta}+0.05\ell_{\mathrm{temp}}.
\end{aligned}
\label{eq:supp_raster}
\end{equation}
Depth and point-to-plane residuals use Huber thresholds of $10$ and $5$ mm,
respectively; chromaticity uses a Huber threshold of $0.10$.  The RGB term is
zero when the input mesh has no valid texture.  Without measured depth, the
depth, point-to-plane, and depth-derived occlusion terms are disabled.

\begin{table}[tbp]
\caption{Algorithmic procedure of \ours{}.  Candidate support is fixed before
adaptation; only $\Theta$ and bounded local pose variables are optimized.
Causal and cross-fit controls restrict the frames used in lines~12--14.}
\centering
\suppTableFont
\setlength{\tabcolsep}{4pt}
\renewcommand{\arraystretch}{1.12}
\begin{tabularx}{\columnwidth}{@{}r >{\raggedright\arraybackslash}X@{}}
\toprule
& \textbf{Algorithm 1: Recording-local 6D pose tracking} \\
\midrule
1 & \textbf{Input:}
$\mathcal O,\{I_t,M_t,\mathbf K_t,\mathbf p^w_{t,j},\mathbf u_{t,j},d_{t,j},v_{t,j},
\mathbf f^s_{t,j}\}_{t=1}^{T},\{\widetilde{\mathbf x}_i,\mathbf f_i^o\}_{i=1}^{N_o}$. \\
2 & $\mathbf x_i\leftarrow s_{\mathrm{on}}R_{\mathrm{on}}
\widetilde{\mathbf x}_i+\mathbf t_{\mathrm{on}}$. \\
3 & \textbf{for} $t=1,\ldots,T$ \textbf{do} \\
4 & \quad $\calH_t^{\mathrm{retr}}\leftarrow
\operatorname{Top}_{240}\!\bigl(\operatorname{DINO}(I_t,\{\mathbf f_i^o\})\bigr)$. \\
5 & \quad $\mathbf L_{t,h}\leftarrow
\mathbf Q_{t,h}(\mathbf A_t^s)^\top/\sqrt d+\mathbf B_{t,h}^{\mathrm{geo}}$;
\quad $\mathbf C_{t,h}\leftarrow
\operatorname{softmax}_j(\mathbf L_{t,h})\odot
\operatorname{softmax}_i(\mathbf L_{t,h})$. \\
6 & \quad $(\bar{\mathbf u}_{t,h,i},\bar{\mathbf p}_{t,h,i}^w,w_{t,h,i})
\leftarrow\operatorname{GeoCorr}(\mathbf C_{t,h},\mathbf L_{t,h})$;
\quad $\calH_t^{\mathrm{2D\text{-}3D}}\leftarrow
\operatorname{RANSAC\text{-}PnP}(\mathbf x,\bar{\mathbf u},w)$. \\
7 & \quad $v_t\leftarrow\mathbf 1[\exists j:\,v_{t,j}=1]$; \quad
\textbf{if} $v_t=1$ \textbf{then}\quad $\calH_t^{\mathrm{3D\text{-}3D}}\leftarrow
\operatorname{Kabsch}(\mathbf x,\bar{\mathbf p}^{w},w)$ \\
8 & \quad \textbf{else}\quad $\calH_t^{\mathrm{3D\text{-}3D}}\leftarrow\varnothing$
\quad \textbf{end if} \\
9 & \quad \textbf{if} $t>1$ \textbf{then}\quad
$\calH_t^{\mathrm{prop}}\leftarrow
\operatorname{Propagate}(\calH_{t-1},\widehat{\Delta\G}_{t-1\to t})$
\quad \textbf{else}\quad $\calH_t^{\mathrm{prop}}\leftarrow\varnothing$
\quad \textbf{end if} \\
10 & \quad $\calH_t\leftarrow\operatorname{TopK}\!\left(
\operatorname{NMS}(\calH_t^{\mathrm{retr}}\cup
\calH_t^{\mathrm{2D\text{-}3D}}\cup
\calH_t^{\mathrm{3D\text{-}3D}}\cup
\calH_t^{\mathrm{prop}})\right)$;
\quad $\{\mathbf a_{t,h},\boldsymbol\chi_{t,h}\}_{h\in\calH_t}\leftarrow
\operatorname{Store}(\cdot)$. \\
11 & \textbf{end for} \\
12 & $\Theta\leftarrow\operatorname{Init}_{\mathrm{recording\text{-}asset}}()$; \quad
\textbf{for} $n=1,\ldots,600$ \textbf{do} \\
13 & \quad $\mathbf z_{t,h}\leftarrow
\operatorname{Tr}_{\mathrm{hyp}}\!\left(\operatorname{MLP}
\left(\operatorname{LN}[\mathbf a_{t,h},\boldsymbol\chi_{t,h}]\right)\right)$;
\quad $(e_t,o_t,\tau_t,\sigma_t^R,\boldsymbol\delta_t)
\leftarrow\operatorname{Tr}_{\mathrm{temp}}(\{\mathbf z_{t,h}\}_{t,h},\gamma_4(t/T))$. \\
14 & \quad $\log q_t(h)\leftarrow
\log\operatorname{softmax}_h\!\left(o_te_t(h)/\tau_t+v_t r_t(h)\right)$;
\quad $\Theta\leftarrow\operatorname{AdamW}
\left(\Theta,\nabla_\Theta\mathcal L_{\mathrm{seq}}\right)$. \\
15 & \textbf{end for} \\
16 & \textbf{for} $t=1,\ldots,T$ \textbf{do}\quad
$h_t^*\leftarrow\arg\max_hq_t(h)$;
\quad $\G_t^{\mathrm{par}}\leftarrow\G_{t,h_t^*}$;
\quad $\G_t^{\mathrm{upd}}\leftarrow
\operatorname{Adam}_{100}(\mathcal L_{\mathrm{refine}};\G_t^{\mathrm{par}})$. \\
17 & \quad \textbf{if}
$\ell_{\mathrm{meas}}(\G_t^{\mathrm{upd}})\leq
\ell_{\mathrm{meas}}(\G_t^{\mathrm{par}})$ \textbf{then}\quad
$\G_t^{\mathrm{out}}\leftarrow
\operatorname{Outlet}(\G_t^{\mathrm{par}},\G_t^{\mathrm{upd}},I_t,\eta_t,b_t)$ \\
18 & \quad \textbf{else}\quad $\G_t^{\mathrm{out}}\leftarrow\G_t^{\mathrm{par}}$
\quad \textbf{end if} \\
19 & \textbf{end for}; \quad \textbf{output}\quad
$\{\G_t^{\mathrm{out}}\}_{t=1}^{T}$. \\
\bottomrule
\end{tabularx}

\label{tab:supp_algorithm}
\end{table}

\FloatBarrier
\section{Additional Experimental Results}
\label{sec:supp_experiments}

\subsection{Protocol Details and Temporal Metric}

Within each dataset/CAD panel, all methods share the chronological source
stream, calibration, corresponding Metric CAD or SAM3D CAD, and scored keys;
SAM3D CAD alignment is fixed per benchmark before temporal inference, while Metric CAD
rows use the prescribed dataset chart.  Measured RGB-D is used on YCBInEOAT
and LINEMOD for 3D--3D support and depth/point-to-plane measurement, whereas
HOT3D uses RGB-derived VGGT-$\Omega$ scene evidence. The matched solver
controls in Table~\ref{tab:supp_solver_matched} compare 2D--3D and 3D--3D
support using this evidence; proxy geometry is not treated as measured depth
in the dense depth and point-to-plane measurement terms.

For adjacent scored keys $(t-1,t)\in\mathcal E$, let
$E_t=(T_t^{\mathrm{gt}})^{-1}\hat T_t$ be the relative error between estimated
and ground-truth object-to-camera transforms in the camera frame,
$\Delta E_t=E_{t-1}^{-1}E_t$, $\mathbf t(\cdot)$ extract translation, and
$c_t^{p,q}=\mathbf{1}[\mathrm{MSPD}_t<p,\,\mathrm{MSSD}_t<qd]$.  The HOT3D
Motion diagnostic counts a pair only when both endpoint poses satisfy the BOP
correctness gate and their relative error is stable:
\[
\begin{aligned}
\mathrm{Motion}_{p,q}
&=\frac{1}{|\mathcal E|}\sum_{(t-1,t)\in\mathcal E}c_{t-1}^{p,q}c_t^{p,q}\\[-1pt]
&\quad\cdot\mathbf{1}\!\left[\theta(\Delta E_t)<30^\circ,\,
\|\mathbf t(\Delta E_t)\|_2<qd\right].
\end{aligned}
\]
Here $d$ is the object diameter.  We use recording--asset cluster bootstrap
intervals for HOT3D and scene bootstrap intervals for YCBInEOAT; all reported
intervals are two-sided 95\% percentile intervals.

\subsection{Complete Fine-Grained Results}
\label{sec:supp_complete_results}

The printed tables retain aggregate results and controls that answer distinct
protocol questions.  The complete frozen per-asset and per-object evaluator
rows, including every method and both CAD lanes, are supplied in the
accompanying structured record.

The aggregate comparison is frame weighted within each dataset and CAD lane;
rows are never averaged across datasets or object charts.  BOP AR summarizes
MSPD, MSSD, and VSD recall, while ADD AUC, ADD-S AUC, and ADD-$0.1d$ use the
same fixed object chart and scored keys as the corresponding AR row.  The
fine-grained tables then expose the recording--asset or object identities
behind each aggregate, and the HOT3D threshold sweep evaluates the same
chronological trajectories at progressively broader correctness gates.

\paragraph{Aggregate results.}
{\suppTableFont
\setlength{\tabcolsep}{2pt}
\renewcommand{\arraystretch}{1.08}
\begin{longtable}{@{}c@{}}
\caption{Complete aggregate results across all three benchmarks and both CAD conditions. All values except rotation and translation are percentages; translation is reported in centimeters; best values are bold.}\label{tab:supp_all_protocol_aggregate}\\
\endfirsthead
\endhead
\endfoot
\endlastfoot
\begin{minipage}{\linewidth}
\centering
\textit{HOT3D}\par\smallskip
\suppfit{\linewidth}{\begin{tabular}{llrrrrr}
\toprule
Object CAD & Method & AR$\uparrow$ & AR$_{\mathrm{MSPD}}\uparrow$ & AR$_{\mathrm{MSSD}}\uparrow$ & AR$_{\mathrm{VSD}}\uparrow$ & Motion$\uparrow$ \\
\midrule
\multirow{7}{*}{SAM3D CAD} & FoundationPose & 3.4 & 7.0 & 3.0 & 0.0 & 2.4 \\
 & RGBTrack & 3.4 & 9.2 & 0.9 & 0.1 & 1.3 \\
 & GigaPose & 31.4 & 69.2 & \textbf{22.6} & \textbf{2.4} & 11.9 \\
 & FreePose & 1.9 & 5.6 & 0.0 & 0.0 & 0.0 \\
 & Dynhor & 31.9 & 76.8 & 17.9 & 1.0 & 17.0 \\
 & BIT & 1.6 & 4.0 & 0.7 & 0.1 & 0.0 \\
 & \textbf{Ours} & \textbf{34.0} & \textbf{79.0} & 20.9 & 2.1 & \textbf{26.9} \\
\midrule
\multirow{7}{*}{Metric CAD} & FoundationPose & 2.5 & 5.7 & 1.7 & 0.0 & 0.4 \\
 & RGBTrack & 4.3 & 9.1 & 2.4 & 1.4 & 1.5 \\
 & GigaPose & \textbf{48.5} & 74.6 & \textbf{44.2} & \textbf{26.7} & 22.1 \\
 & FreePose & 1.6 & 4.9 & 0.0 & 0.0 & 0.0 \\
 & Dynhor & 38.8 & 80.2 & 31.8 & 4.3 & 27.8 \\
 & BIT & 2.4 & 6.7 & 0.5 & 0.0 & 0.0 \\
 & \textbf{Ours} & 43.1 & \textbf{81.1} & 34.6 & 13.5 & \textbf{36.9} \\
\bottomrule
\end{tabular}}
\end{minipage}\\[8pt]
\begin{minipage}{\linewidth}
\centering
\textit{YCBInEOAT}\par\smallskip
\suppfit{\linewidth}{\begin{tabular}{llrrrrrrrrr}
\toprule
Object CAD & Method & AR$\uparrow$ & AR$_{\mathrm{MSPD}}\uparrow$ & AR$_{\mathrm{MSSD}}\uparrow$ & AR$_{\mathrm{VSD}}\uparrow$ & \shortstack{ADD\\AUC$\uparrow$} & \shortstack{ADD-S\\AUC$\uparrow$} & ADD-$0.1d\uparrow$ & \shortstack{Rot.\\(deg.)$\downarrow$} & \shortstack{Trans.\\(cm)$\downarrow$} \\
\midrule
\multirow{7}{*}{SAM3D CAD} & FoundationPose & \textbf{46.3} & \textbf{61.4} & \textbf{53.3} & 24.1 & \textbf{43.1} & \textbf{67.2} & \textbf{41.0} & 60.9 & 7.5 \\
 & RGBTrack & 17.2 & 42.6 & 8.7 & 0.2 & 5.6 & 21.5 & 0.3 & 109.9 & 42.4 \\
 & GigaPose & 20.7 & 39.8 & 19.7 & 2.6 & 16.1 & 31.3 & 6.3 & \textbf{53.1} & 49.1 \\
 & FreePose & 13.1 & 37.7 & 1.5 & 0.0 & 0.0 & 5.7 & 0.0 & 111.6 & 67.0 \\
 & Dynhor & 15.3 & 33.6 & 11.2 & 1.0 & 2.0 & 25.5 & 0.0 & 153.0 & 18.5 \\
 & BIT & 7.2 & 18.3 & 3.0 & 0.3 & 2.3 & 7.8 & 0.3 & 98.8 & 36.5 \\
 & \textbf{Ours} & 45.4 & 56.5 & 53.1 & \textbf{26.6} & 41.2 & 67.0 & 40.7 & 71.8 & \textbf{6.7} \\
\midrule
\multirow{7}{*}{Metric CAD} & FoundationPose & \textbf{88.7} & \textbf{98.5} & \textbf{96.8} & \textbf{70.9} & 84.5 & 95.9 & 85.2 & 27.9 & 0.7 \\
 & RGBTrack & 55.1 & 91.1 & 50.1 & 24.2 & 46.4 & 56.2 & 35.7 & 40.0 & 49.4 \\
 & GigaPose & 27.2 & 51.7 & 22.9 & 7.2 & 9.1 & 33.3 & 1.9 & 118.8 & 75.9 \\
 & FreePose & 14.8 & 44.5 & 0.0 & 0.0 & 0.0 & 0.0 & 0.0 & 144.3 & 106.0 \\
 & Dynhor & 38.1 & 82.0 & 30.2 & 2.2 & 20.4 & 57.5 & 4.8 & 88.2 & 9.4 \\
 & BIT & 27.9 & 48.8 & 29.3 & 5.5 & 26.9 & 42.0 & 7.7 & 54.9 & 23.3 \\
 & \textbf{Ours} & 87.4 & 98.0 & 96.4 & 67.9 & \textbf{92.5} & \textbf{96.2} & \textbf{95.8} & \textbf{6.4} & \textbf{0.5} \\
\bottomrule
\end{tabular}}
\end{minipage}\\[8pt]
\begin{minipage}{\linewidth}
\centering
\textit{LINEMOD}\par\smallskip
\suppfit{\linewidth}{\begin{tabular}{llrrrrrrrrr}
\toprule
Object CAD & Method & AR$\uparrow$ & AR$_{\mathrm{MSPD}}\uparrow$ & AR$_{\mathrm{MSSD}}\uparrow$ & AR$_{\mathrm{VSD}}\uparrow$ & \shortstack{ADD\\AUC$\uparrow$} & \shortstack{ADD-S\\AUC$\uparrow$} & ADD-$0.1d\uparrow$ & \shortstack{Rot.\\(deg.)$\downarrow$} & \shortstack{Trans.\\(cm)$\downarrow$} \\
\midrule
\multirow{7}{*}{SAM3D CAD} & FoundationPose & 43.1 & 64.0 & 61.7 & 3.6 & 38.2 & 84.1 & 5.6 & 94.1 & 3.3 \\
 & RGBTrack & 2.3 & 6.1 & 0.8 & 0.1 & 0.4 & 2.1 & 0.1 & 115.4 & 63.6 \\
 & GigaPose & 24.0 & 53.0 & 18.8 & 0.1 & 10.4 & 37.5 & 1.0 & 90.1 & 17.1 \\
 & FreePose & 14.4 & 36.1 & 7.0 & 0.0 & 1.7 & 18.3 & 0.0 & 125.9 & 26.4 \\
 & Dynhor & \textbf{66.5} & \textbf{94.5} & \textbf{73.6} & \textbf{31.5} & \textbf{59.9} & 85.5 & \textbf{47.2} & \textbf{47.1} & 3.0 \\
 & BIT & 5.3 & 14.7 & 1.1 & 0.1 & 0.6 & 1.7 & 0.1 & 120.4 & 2345.3 \\
 & \textbf{Ours} & 42.8 & 62.1 & 63.6 & 2.7 & 44.8 & \textbf{86.1} & 6.2 & 83.8 & \textbf{2.6} \\
\midrule
\multirow{7}{*}{Metric CAD} & FoundationPose & 2.4 & 3.6 & 2.2 & 1.4 & 1.8 & 3.4 & 1.7 & 111.2 & 83.7 \\
 & RGBTrack & 3.8 & 9.4 & 1.5 & 0.4 & 1.0 & 3.3 & 0.6 & 106.6 & 82.9 \\
 & GigaPose & 76.0 & 95.8 & 86.0 & 46.3 & 75.1 & 89.6 & 70.6 & 18.5 & 3.0 \\
 & FreePose & 15.0 & 44.0 & 0.9 & 0.0 & 0.2 & 4.0 & 0.0 & 125.2 & 33.7 \\
 & Dynhor & 66.5 & 94.5 & 73.6 & 31.5 & 59.9 & 85.5 & 47.2 & 47.1 & 3.0 \\
 & BIT & 8.1 & 21.5 & 1.7 & 1.2 & 1.6 & 1.9 & 1.6 & 111.2 & 1240.1 \\
 & \textbf{Ours} & \textbf{85.6} & \textbf{96.1} & \textbf{96.4} & \textbf{64.2} & \textbf{86.7} & \textbf{96.4} & \textbf{88.0} & \textbf{14.7} & \textbf{0.6} \\
\bottomrule
\end{tabular}}
\end{minipage}\\[8pt]
\end{longtable}
}

\FloatBarrier

\begin{table}[tbp]
\caption{HOT3D temporal-threshold sensitivity.  $\mathrm{M}@p/qd$ denotes
correctness-gated consecutive-pair retention at MSPD threshold $p$ pixels and
MSSD threshold $qd$.}
\centering
\suppTableFont
\setlength{\tabcolsep}{4pt}
\suppfit{\textwidth}{%
\begin{tabular}{llccccc}
\toprule
Object CAD & Method & Macro AR$\uparrow$ & Frame AR$\uparrow$ & M@20/.2$d\uparrow$ & M@30/.3$d\uparrow$ & M@50/.5$d\uparrow$ \\
\midrule
\multirow{7}{*}{SAM3D CAD}
& FoundationPose & .033 & .034 & .000 & .009 & .024 \\
& RGBTrack & .043 & .034 & .000 & .000 & .013 \\
& GigaPose & .251 & .314 & \textbf{.031} & .060 & .119 \\
& FreePose & .010 & .019 & .000 & .000 & .000 \\
& Dynhor & \textbf{.296} & .319 & .009 & .057 & .170 \\
& BIT & .048 & .016 & .000 & .000 & .000 \\
& \textbf{Ours} & .293 & \textbf{.340} & .022 & \textbf{.086} & \textbf{.269} \\
\midrule
\multirow{7}{*}{Metric CAD}
& FoundationPose & .031 & .025 & .000 & .000 & .004 \\
& RGBTrack & .051 & .043 & .011 & .011 & .015 \\
& GigaPose & .363 & \textbf{.485} & \textbf{.170} & .201 & .221 \\
& FreePose & .009 & .016 & .000 & .000 & .000 \\
& Dynhor & .365 & .388 & .051 & .135 & .278 \\
& BIT & .027 & .024 & .000 & .000 & .000 \\
& \textbf{Ours} & \textbf{.373} & .431 & .119 & \textbf{.245} & \textbf{.369} \\
\bottomrule
\end{tabular}
}

\label{tab:supp_motion}
\end{table}

\begin{figure}[!htbp]
\centering
\includegraphics[width=\textwidth]{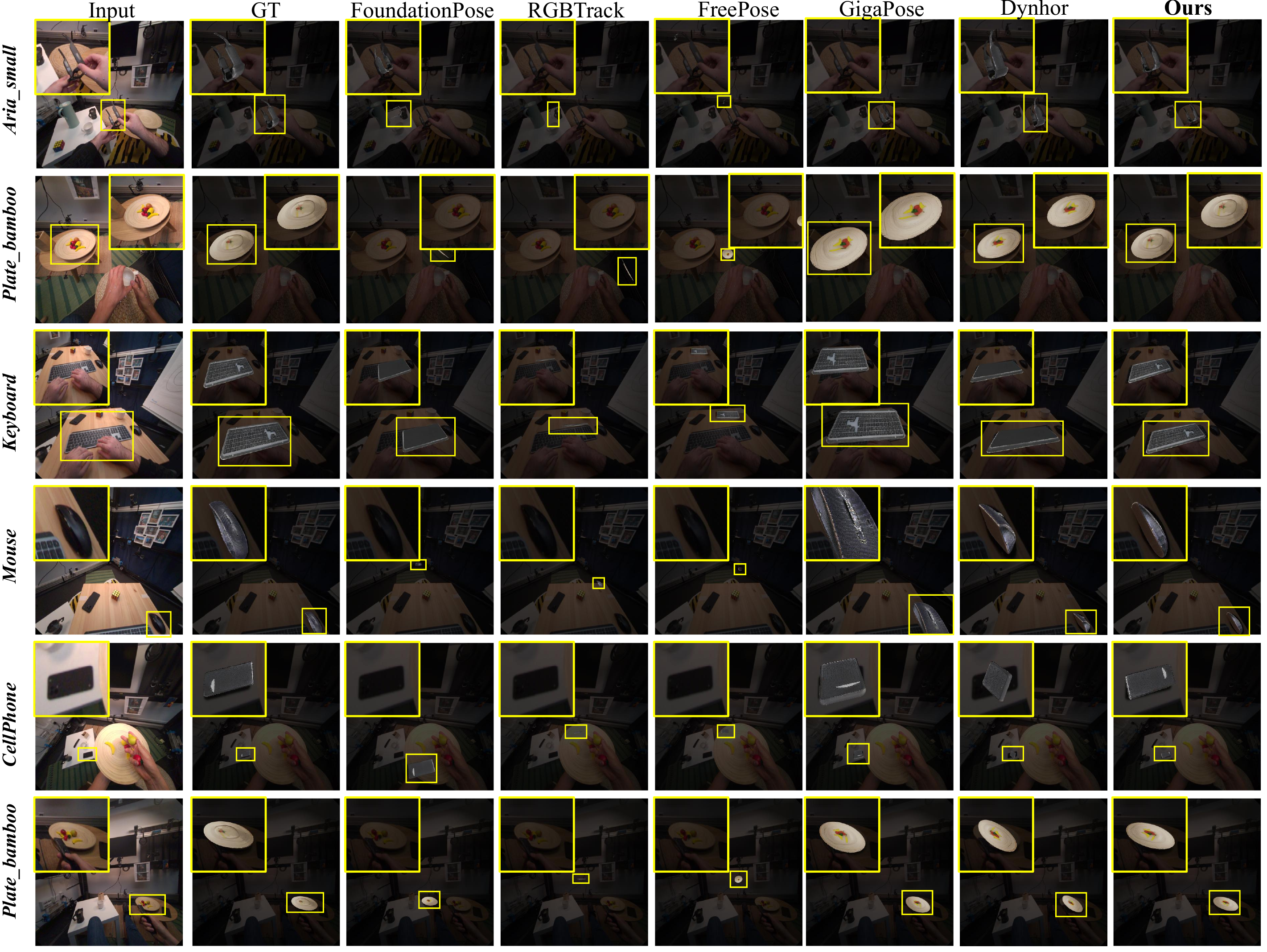}
\caption{\textbf{Qualitative comparison on HOT3D with SAM3D CAD.} Each row compares the shared RGB observation, ground truth, GRC-Pose, and the external methods for one target object. Yellow boxes identify target regions; rendered SAM3D CAD overlays visualize the recovered 6D poses.}
\label{fig:supp_hot3d_qualitative}
\end{figure}
\paragraph{HOT3D.}
The complete records cover all twelve assets and both CAD lanes.  The aggregate
rows use all 476 chronological keys and retain the common frame-weighted
evaluation protocol.  The threshold sweep separates high-precision endpoint
registration from retention across broader accepted pose neighborhoods.
GRC-Pose leads at both the intermediate and reported gates in the two CAD
conditions.  At the reported gate, its margin over Dynhor is .099 and .091,
respectively.  This consistent response across both CAD conditions supports
the temporal posterior interpretation of the main HOT3D gain.

The full threshold response makes this behavior explicit.  With SAM3D CAD,
GRC-Pose increases from .022 at M@20/.2$d$ to .086 at M@30/.3$d$ and .269 at
M@50/.5$d$; with Metric CAD, the corresponding values are .119, .245, and
.369.  The gain therefore persists across object charts and becomes strongest
when the metric admits the viewpoint, scale, and partial-occlusion variation
encountered along the chronological stream.  Figure~\ref{fig:supp_hot3d_qualitative}
shows the same setting on Aria, plate, keyboard, mouse, and cellphone targets,
with every method evaluated on the shared RGB observation and SAM3D CAD.

Table~\ref{tab:supp_hot3d_sam3d_cad_assets} prints the complete per-asset
breakdown for the primary SAM3D CAD setting.  The corresponding Metric CAD
rows remain in the frozen structured record together with the complete
object-level rows for the other benchmarks.  The main-text seven-of-nine
summary uses the nine assets with at least ten scored keys; the table retains
all twelve assets.

The per-asset rows localize the aggregate result.  Among the nine assets with
at least ten scored keys, GRC-Pose leads or ties AR on seven.  Its strongest
joint spatial--temporal responses occur on keyboard (34.2 AR and 55.9 Motion),
cellphone (34.9 and 31.3), mug (43.1 and 40.0), and eraser (42.5 and 39.1).
Coffee pot additionally reaches the best MSPD recall and Motion despite a
near-tie in AR, while plate reaches the best Motion score.  These rows connect
the aggregate temporal gain to targets that undergo hand occlusion, changing
viewpoint, and substantial image-scale variation.

The result is also supported by the longest object streams rather than the
three assets with the fewest scored keys.  Among cellphone ($N=86$), coffee pot
($N=86$), mug ($N=89$), and puzzle ($N=61$), GRC-Pose leads AR on cellphone,
mug, and puzzle, remains within .4 points of the best coffee-pot AR, and gives
the highest Motion score on cellphone, coffee pot, and mug.  These rows account
for most of the frame-weighted evaluation and show that the aggregate gain is
retained over extended chronological trajectories.

The macro and frame-weighted views emphasize different parts of the same
record.  With SAM3D CAD, GRC-Pose remains within .003 macro AR of Dynhor and
leads frame AR by .021.  With Metric CAD, it gives the highest macro AR and
reported Motion, while GigaPose gives the highest frame AR.  Reading these
columns together separates framewise registration from correctness retained
between adjacent scored keys: the principal GRC-Pose advantage is preserved at
the reported temporal gate in both CAD conditions.

\FloatBarrier

The qualitative rows expose the same behavior at image level.  Keyboard and
cellphone combine close hand interaction with large changes in visible object
support, while the two plate views cover distinct viewpoint and image-scale
conditions.  Mouse and Aria retain much smaller target regions.  Across these
cases, the shared target boxes and SAM3D CAD overlays make both object extent
and recovered orientation directly visible, complementing the per-asset AR
and Motion scores below.

The keyboard rows show the recovered long-axis orientation under two hand
configurations, consistent with the 34.2 AR and 55.9 Motion reported in
Table~\ref{tab:supp_hot3d_sam3d_cad_assets}.  On cellphone, the overlay remains
inside the shared target region as the visible face and image scale change;
this sequence reaches 34.9 AR and 31.3 Motion.  The two plate rows provide the
same comparison across near and distant views, where GRC-Pose reaches the best
Motion score.  Finally, the Aria and mouse rows test substantially smaller
image support.  Together, these examples connect the recovered 6D pose shown
by the overlays to the spatial and temporal measures in the per-asset table.
{\suppTableFont
\setlength{\tabcolsep}{5pt}
\renewcommand{\arraystretch}{1.08}
\begin{longtable}{@{}llrrrrrr@{}}
\caption{Complete HOT3D per-asset results with SAM3D CAD. The two panels jointly report all twelve assets and all seven methods. All values except the frame count $N$ are percentages; Motion is correctness-gated consecutive-pair retention at the reported $50$-pixel/$0.5d$ gate. Best per-asset values are bold.}\label{tab:supp_hot3d_sam3d_cad_assets}\\
\toprule
Asset & Method & $N$ & AR$\uparrow$ & \shortstack{AR$_{\mathrm{MSPD}}$\\$\uparrow$} & \shortstack{AR$_{\mathrm{MSSD}}$\\$\uparrow$} & \shortstack{AR$_{\mathrm{VSD}}$\\$\uparrow$} & Motion$\uparrow$ \\
\midrule\endfirsthead
\toprule
Asset & Method & $N$ & AR$\uparrow$ & \shortstack{AR$_{\mathrm{MSPD}}$\\$\uparrow$} & \shortstack{AR$_{\mathrm{MSSD}}$\\$\uparrow$} & \shortstack{AR$_{\mathrm{VSD}}$\\$\uparrow$} & Motion$\uparrow$ \\
\midrule\endhead
\bottomrule\endfoot
\bottomrule\endlastfoot
\multicolumn{8}{c}{\textit{Assets 1--6}}\\*
\midrule
aria & FoundationPose & 4 & 4.2 & 12.5 & 0.0 & \textbf{0.0} & \textbf{0.0} \\*
 & RGBTrack &  & 0.0 & 0.0 & 0.0 & \textbf{0.0} & \textbf{0.0} \\*
 & GigaPose &  & 4.2 & 12.5 & 0.0 & \textbf{0.0} & \textbf{0.0} \\*
 & FreePose &  & 0.0 & 0.0 & 0.0 & \textbf{0.0} & \textbf{0.0} \\*
 & Dynhor &  & \textbf{14.2} & \textbf{22.5} & \textbf{20.0} & \textbf{0.0} & \textbf{0.0} \\*
 & BIT &  & 4.2 & 12.5 & 0.0 & \textbf{0.0} & \textbf{0.0} \\*
 & \textbf{Ours} &  & 4.2 & 12.5 & 0.0 & \textbf{0.0} & \textbf{0.0} \\

\midrule
cellphone & FoundationPose & 86 & 2.8 & 8.3 & 0.0 & 0.0 & 0.0 \\*
 & RGBTrack &  & 4.1 & 12.0 & 0.2 & 0.0 & 0.0 \\*
 & GigaPose &  & 32.6 & 76.7 & \textbf{18.7} & \textbf{2.2} & 2.4 \\*
 & FreePose &  & 4.0 & 12.1 & 0.0 & 0.0 & 0.0 \\*
 & Dynhor &  & 33.7 & 88.4 & 12.4 & 0.3 & 16.9 \\*
 & BIT &  & 2.6 & 7.4 & 0.2 & 0.0 & 0.0 \\*
 & \textbf{Ours} &  & \textbf{34.9} & \textbf{90.9} & 13.4 & 0.3 & \textbf{31.3} \\

\midrule
coffee pot & FoundationPose & 86 & 1.0 & 2.7 & 0.5 & 0.0 & 0.0 \\*
 & RGBTrack &  & 1.7 & 4.8 & 0.2 & 0.0 & 0.0 \\*
 & GigaPose &  & 23.9 & 57.4 & 14.1 & 0.2 & 15.9 \\*
 & FreePose &  & 2.4 & 7.1 & 0.0 & 0.0 & 0.0 \\*
 & Dynhor &  & \textbf{27.8} & 67.1 & \textbf{15.7} & \textbf{0.6} & 13.4 \\*
 & BIT &  & 1.4 & 2.3 & 1.9 & 0.0 & 0.0 \\*
 & \textbf{Ours} &  & 27.4 & \textbf{67.3} & 14.7 & 0.3 & \textbf{23.2} \\

\midrule
remote & FoundationPose & 6 & 0.0 & 0.0 & 0.0 & 0.0 & 0.0 \\*
 & RGBTrack &  & 0.0 & 0.0 & 0.0 & 0.0 & 0.0 \\*
 & GigaPose &  & 37.8 & 60.0 & \textbf{28.3} & \textbf{25.0} & \textbf{20.0} \\*
 & FreePose &  & 0.0 & 0.0 & 0.0 & 0.0 & 0.0 \\*
 & Dynhor &  & 27.2 & 81.7 & 0.0 & 0.0 & 0.0 \\*
 & BIT &  & 0.0 & 0.0 & 0.0 & 0.0 & 0.0 \\*
 & \textbf{Ours} &  & \textbf{40.0} & \textbf{91.7} & 25.0 & 3.3 & 0.0 \\

\midrule
flask & FoundationPose & 1 & 0.0 & 0.0 & 0.0 & \textbf{0.0} & \textbf{0.0} \\*
 & RGBTrack &  & 23.3 & 70.0 & 0.0 & \textbf{0.0} & \textbf{0.0} \\*
 & GigaPose &  & 0.0 & 0.0 & 0.0 & \textbf{0.0} & \textbf{0.0} \\*
 & FreePose &  & 0.0 & 0.0 & 0.0 & \textbf{0.0} & \textbf{0.0} \\*
 & Dynhor &  & \textbf{50.0} & \textbf{80.0} & \textbf{70.0} & \textbf{0.0} & \textbf{0.0} \\*
 & BIT &  & 43.3 & 70.0 & 60.0 & \textbf{0.0} & \textbf{0.0} \\*
 & \textbf{Ours} &  & 16.7 & 50.0 & 0.0 & \textbf{0.0} & \textbf{0.0} \\

\midrule
keyboard & FoundationPose & 35 & 19.1 & 18.0 & 39.1 & \textbf{0.3} & 32.4 \\*
 & RGBTrack &  & 0.0 & 0.0 & 0.0 & 0.0 & 0.0 \\*
 & GigaPose &  & 22.2 & 20.3 & 46.3 & 0.0 & 14.7 \\*
 & FreePose &  & 0.0 & 0.0 & 0.0 & 0.0 & 0.0 \\*
 & Dynhor &  & 18.2 & 29.4 & 25.1 & 0.0 & 41.2 \\*
 & BIT &  & 0.0 & 0.0 & 0.0 & 0.0 & 0.0 \\*
 & \textbf{Ours} &  & \textbf{34.2} & \textbf{42.9} & \textbf{59.7} & 0.0 & \textbf{55.9} \\

\addlinespace[6pt]
\multicolumn{8}{c}{\textit{Assets 7--12}}\\*
\midrule
mouse & FoundationPose & 40 & 3.8 & 11.2 & 0.2 & 0.0 & 0.0 \\*
 & RGBTrack &  & 7.3 & 16.8 & 5.0 & 0.0 & \textbf{10.3} \\*
 & GigaPose &  & 25.1 & 75.0 & 0.2 & 0.0 & 0.0 \\*
 & FreePose &  & 0.0 & 0.0 & 0.0 & 0.0 & 0.0 \\*
 & Dynhor &  & 30.6 & 86.5 & \textbf{5.2} & 0.0 & 2.6 \\*
 & BIT &  & 2.7 & 8.2 & 0.0 & 0.0 & 0.0 \\*
 & \textbf{Ours} &  & \textbf{31.0} & \textbf{88.8} & 3.2 & \textbf{1.0} & 0.0 \\

\midrule
mug & FoundationPose & 89 & 1.3 & 3.7 & 0.2 & 0.0 & 0.0 \\*
 & RGBTrack &  & 4.5 & 12.0 & 1.1 & 0.3 & 1.2 \\*
 & GigaPose &  & 43.0 & 84.8 & \textbf{37.9} & 6.3 & 23.5 \\*
 & FreePose &  & 1.2 & 3.6 & 0.0 & 0.0 & 0.0 \\*
 & Dynhor &  & 41.8 & 88.0 & 33.8 & 3.5 & 29.4 \\*
 & BIT &  & 1.3 & 2.6 & 0.9 & 0.4 & 0.0 \\*
 & \textbf{Ours} &  & \textbf{43.1} & \textbf{89.9} & 32.6 & \textbf{6.9} & \textbf{40.0} \\

\midrule
plate & FoundationPose & 33 & 0.4 & 1.2 & 0.0 & 0.0 & 0.0 \\*
 & RGBTrack &  & 2.0 & 3.3 & 2.7 & 0.0 & 3.2 \\*
 & GigaPose &  & \textbf{44.3} & 69.4 & \textbf{63.0} & 0.6 & 38.7 \\*
 & FreePose &  & 0.0 & 0.0 & 0.0 & 0.0 & 0.0 \\*
 & Dynhor &  & 37.0 & \textbf{69.7} & 40.0 & 1.2 & 16.1 \\*
 & BIT &  & 0.0 & 0.0 & 0.0 & 0.0 & 0.0 \\*
 & \textbf{Ours} &  & 34.4 & 61.5 & 39.1 & \textbf{2.7} & \textbf{48.4} \\

\midrule
puzzle & FoundationPose & 61 & 4.0 & 12.1 & 0.0 & 0.0 & \textbf{0.0} \\*
 & RGBTrack &  & 4.5 & 13.4 & 0.0 & 0.0 & \textbf{0.0} \\*
 & GigaPose &  & 31.7 & 88.5 & \textbf{5.4} & \textbf{1.3} & \textbf{0.0} \\*
 & FreePose &  & 3.6 & 10.7 & 0.0 & 0.0 & \textbf{0.0} \\*
 & Dynhor &  & 31.0 & 89.5 & 2.6 & 0.8 & \textbf{0.0} \\*
 & BIT &  & 1.9 & 5.7 & 0.0 & 0.0 & \textbf{0.0} \\*
 & \textbf{Ours} &  & \textbf{33.1} & \textbf{96.9} & 2.5 & 0.0 & \textbf{0.0} \\

\midrule
eraser & FoundationPose & 24 & 1.8 & 5.4 & 0.0 & 0.0 & 0.0 \\*
 & RGBTrack &  & 1.4 & 4.2 & 0.0 & 0.0 & 0.0 \\*
 & GigaPose &  & 33.3 & 80.8 & 14.2 & 5.0 & 4.3 \\*
 & FreePose &  & 0.4 & 1.2 & 0.0 & 0.0 & 0.0 \\*
 & Dynhor &  & 35.3 & \textbf{90.4} & 15.0 & 0.4 & 30.4 \\*
 & BIT &  & 0.3 & 0.8 & 0.0 & 0.0 & 0.0 \\*
 & \textbf{Ours} &  & \textbf{42.5} & 85.0 & \textbf{34.6} & \textbf{7.9} & \textbf{39.1} \\

\midrule
marker & FoundationPose & 11 & 0.6 & 1.8 & \textbf{0.0} & \textbf{0.0} & \textbf{0.0} \\*
 & RGBTrack &  & 3.0 & 9.1 & \textbf{0.0} & \textbf{0.0} & \textbf{0.0} \\*
 & GigaPose &  & 2.7 & 8.2 & \textbf{0.0} & \textbf{0.0} & \textbf{0.0} \\*
 & FreePose &  & 0.0 & 0.0 & \textbf{0.0} & \textbf{0.0} & \textbf{0.0} \\*
 & Dynhor &  & 9.1 & 27.3 & \textbf{0.0} & \textbf{0.0} & \textbf{0.0} \\*
 & BIT &  & 0.0 & 0.0 & \textbf{0.0} & \textbf{0.0} & \textbf{0.0} \\*
 & \textbf{Ours} &  & \textbf{9.7} & \textbf{29.1} & \textbf{0.0} & \textbf{0.0} & \textbf{0.0} \\

\addlinespace[6pt]
\end{longtable}
}

\FloatBarrier

\begin{figure}[!htbp]
\centering
\includegraphics[width=\textwidth]{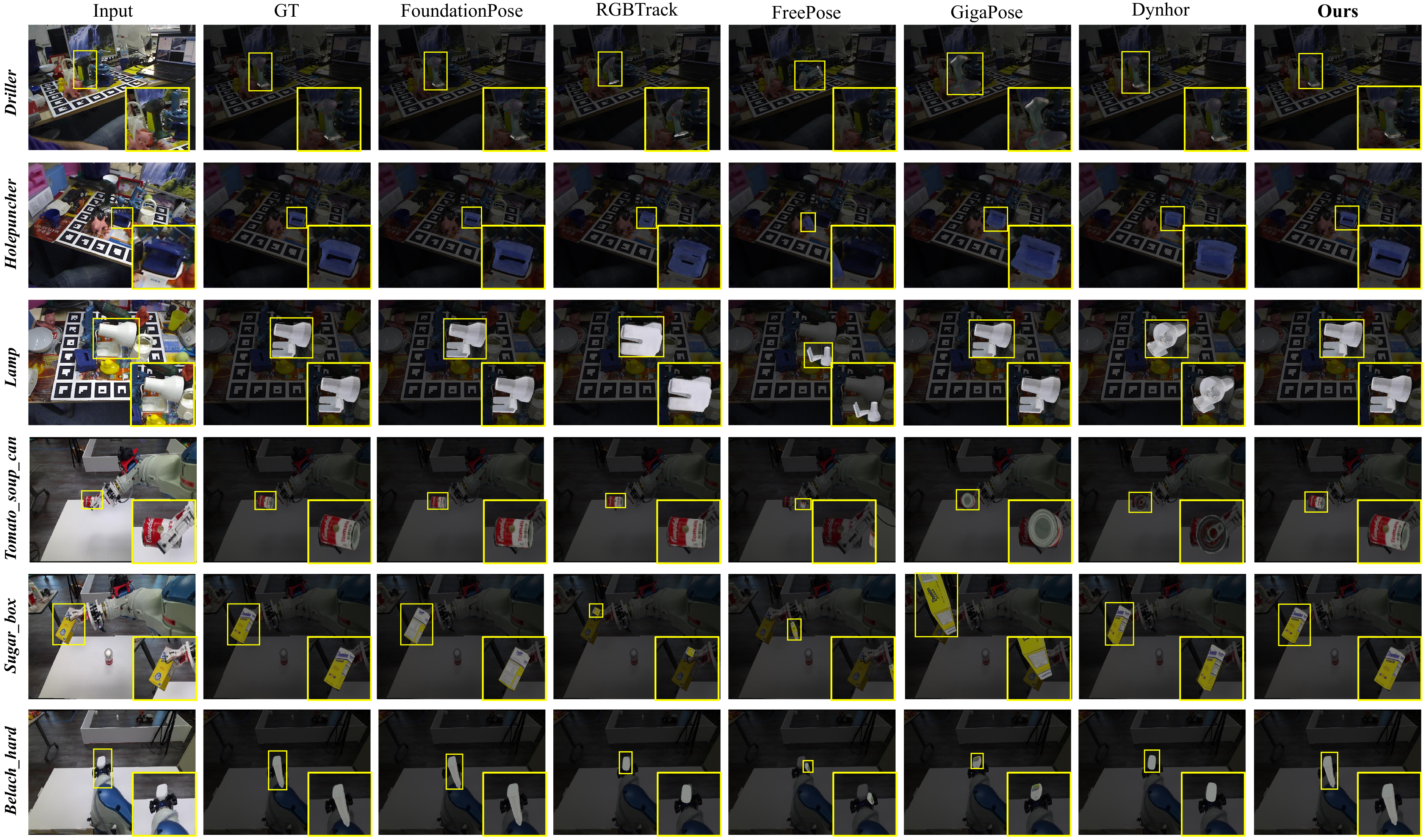}
\caption{\textbf{Qualitative comparison on object-centric benchmarks with SAM3D CAD.} Each row compares the shared RGB observation, ground truth, GRC-Pose, and the external methods for one target object. Rendered SAM3D CAD overlays visualize the recovered 6D poses.}
\label{fig:supp_ycb_geometry}
\end{figure}
\paragraph{YCBInEOAT.}
For FoundationPose under SAM3D CAD, the structured rows use independent register
on every fixed evaluation key, as in the main comparison.  They report the five
object identities under SAM3D CAD and Metric CAD, making the
orientation-sensitive ADD/ADD-S behavior and the BOP components directly
comparable on the same fixed keys.

The object-level breakdown is printed below because the two CAD lanes answer
different questions: SAM3D CAD evaluates the recording-local object chart,
whereas Metric CAD evaluates the dataset object chart. Reading AR together
with ADD, ADD-S, and the pose errors makes the effect of the object chart
explicit without replacing the aggregate comparison.

With SAM3D CAD, GRC-Pose leads all reported measures on bleach and mustard and
achieves the strongest sugar-box AR, ADD-S AUC, and translation.  At the
aggregate level it remains within .9 AR and .2 ADD-S points of
FoundationPose, while attaining the strongest VSD recall and translation.
With Metric CAD, GRC-Pose obtains the best aggregate ADD AUC, ADD-S AUC,
ADD-$0.1d$, rotation, and translation.  The object rows show that this metric
closure is not concentrated in one category: cracker reaches 91.4 ADD AUC and
95.2 ADD-S AUC, while sugar reaches 94.1 and 96.6, respectively.

Figure~\ref{fig:supp_ycb_geometry} complements these scores with driller,
holepuncher, lamp, tomato-soup-can, sugar-box, and bleach examples.  The
rendered overlays use the same SAM3D CAD within each row, so differences arise
from the recovered 6D pose rather than from a method-specific visualization.
Across the varied viewpoints and partial occlusions shown here,
GRC-Pose preserves the target extent and orientation while the shared yellow
regions keep the comparison focused on the same object instance.

{\suppTableFont
\setlength{\tabcolsep}{5pt}
\renewcommand{\arraystretch}{1.08}
\begin{longtable}{@{}llrrrrrr@{}}
\caption{Complete YCBInEOAT per-object results for SAM3D CAD and Metric CAD. AR is the BOP average recall; AUC and threshold values are percentages; translation is reported in centimeters; best per-object values are bold.}\label{tab:supp_ycbineoat_per_object}\\
\toprule
Object ($N$) & Method & AR$\uparrow$ & \shortstack{ADD\\AUC$\uparrow$} & \shortstack{ADD-S\\AUC$\uparrow$} & \shortstack{ADD-$0.1d$\\$\uparrow$} & \shortstack{Rot.\\(deg.)$\downarrow$} & \shortstack{Trans.\\(cm)$\downarrow$} \\
\midrule\endfirsthead
\toprule
Object ($N$) & Method & AR$\uparrow$ & \shortstack{ADD\\AUC$\uparrow$} & \shortstack{ADD-S\\AUC$\uparrow$} & \shortstack{ADD-$0.1d$\\$\uparrow$} & \shortstack{Rot.\\(deg.)$\downarrow$} & \shortstack{Trans.\\(cm)$\downarrow$} \\
\midrule\endhead
\bottomrule\endfoot
\bottomrule\endlastfoot
\multicolumn{8}{c}{\textit{SAM3D CAD}}\\*
\midrule
bleach (57) & FoundationPose & 59.2 & 56.1 & 83.9 & 49.1 & 55.5 & 3.8 \\*
 & RGBTrack & 26.8 & 3.8 & 26.9 & 0.0 & 131.6 & 48.1 \\*
 & GigaPose & 25.9 & 10.7 & 34.6 & 1.8 & 56.0 & 16.1 \\*
 & FreePose & 12.2 & 0.0 & 0.0 & 0.0 & 116.9 & 84.4 \\*
 & Dynhor & 28.5 & 3.5 & 40.8 & 0.0 & 135.5 & 15.2 \\*
 & BIT & 9.6 & 0.5 & 1.3 & 0.0 & 97.3 & 45.5 \\*
 & \textbf{Ours} & \textbf{65.8} & \textbf{72.5} & \textbf{90.0} & \textbf{59.6} & \textbf{9.7} & \textbf{2.5} \\

\midrule
cracker (86) & FoundationPose & \textbf{50.7} & \textbf{52.9} & \textbf{81.4} & \textbf{51.2} & 50.6 & 8.5 \\*
 & RGBTrack & 31.3 & 18.8 & 54.0 & 1.2 & \textbf{39.7} & 20.2 \\*
 & GigaPose & 3.4 & 0.0 & 9.4 & 0.0 & 70.0 & 165.3 \\*
 & FreePose & 11.4 & 0.0 & 25.1 & 0.0 & 115.4 & 95.9 \\*
 & Dynhor & 2.9 & 0.0 & 11.1 & 0.0 & 156.3 & 22.6 \\*
 & BIT & 3.1 & 2.1 & 5.5 & 0.0 & 119.7 & 30.5 \\*
 & \textbf{Ours} & 39.4 & 37.9 & 73.7 & 40.7 & 87.2 & \textbf{5.6} \\

\midrule
mustard (72) & FoundationPose & 38.7 & 36.9 & 66.4 & 36.1 & 62.2 & 6.7 \\*
 & RGBTrack & 12.8 & 3.8 & 24.0 & 0.0 & 91.8 & 35.1 \\*
 & GigaPose & 19.8 & 14.9 & 32.1 & 4.2 & 56.3 & 12.7 \\*
 & FreePose & 16.8 & 0.0 & 0.0 & 0.0 & 98.6 & 67.7 \\*
 & Dynhor & 17.0 & 6.9 & 33.2 & 0.0 & 156.1 & 13.6 \\*
 & BIT & 16.8 & 2.5 & 15.1 & 1.4 & 73.4 & 22.3 \\*
 & \textbf{Ours} & \textbf{41.6} & \textbf{43.6} & \textbf{67.3} & \textbf{48.6} & \textbf{44.7} & \textbf{6.5} \\

\midrule
sugar (97) & FoundationPose & 73.5 & \textbf{60.7} & 90.8 & \textbf{58.8} & 61.3 & 1.8 \\*
 & RGBTrack & 17.7 & 0.3 & 1.5 & 0.0 & 152.8 & 79.6 \\*
 & GigaPose & 45.8 & 45.2 & 68.7 & 20.6 & \textbf{19.4} & 5.9 \\*
 & FreePose & 12.8 & 0.0 & 0.0 & 0.0 & 97.9 & 46.6 \\*
 & Dynhor & 28.4 & 0.7 & 40.7 & 0.0 & 154.9 & 14.1 \\*
 & BIT & 7.3 & 4.9 & 13.5 & 0.0 & 78.0 & 34.8 \\*
 & \textbf{Ours} & \textbf{75.1} & 52.0 & \textbf{93.1} & 51.5 & 86.1 & \textbf{1.4} \\

\midrule
tomato (66) & FoundationPose & 2.4 & 0.0 & 0.1 & \textbf{0.0} & 77.1 & \textbf{18.5} \\*
 & RGBTrack & 0.8 & 0.0 & \textbf{0.9} & \textbf{0.0} & 139.4 & 19.7 \\*
 & GigaPose & 3.0 & \textbf{0.1} & 0.8 & \textbf{0.0} & \textbf{74.7} & 29.2 \\*
 & FreePose & \textbf{15.2} & 0.0 & 0.0 & \textbf{0.0} & 136.4 & 43.7 \\*
 & Dynhor & 0.3 & 0.0 & 0.0 & \textbf{0.0} & 157.7 & 28.1 \\*
 & BIT & 0.1 & 0.0 & 0.0 & \textbf{0.0} & 130.9 & 54.7 \\*
 & \textbf{Ours} & 0.9 & 0.0 & 0.0 & \textbf{0.0} & 114.5 & 19.5 \\

\addlinespace[6pt]
\multicolumn{8}{c}{\textit{Metric CAD}}\\*
\midrule
bleach (57) & FoundationPose & \textbf{86.7} & \textbf{91.6} & \textbf{95.5} & \textbf{98.2} & \textbf{4.3} & \textbf{0.7} \\*
 & RGBTrack & 57.2 & 43.8 & 66.3 & 42.1 & 11.1 & 8.0 \\*
 & GigaPose & 36.5 & 19.3 & 47.3 & 7.0 & 112.0 & 16.1 \\*
 & FreePose & 12.2 & 0.0 & 0.0 & 0.0 & 149.4 & 98.3 \\*
 & Dynhor & 37.9 & 17.4 & 52.8 & 1.8 & 53.4 & 11.4 \\*
 & BIT & 11.9 & 9.2 & 14.1 & 0.0 & 85.2 & 38.3 \\*
 & \textbf{Ours} & 84.3 & 91.0 & 95.1 & \textbf{98.2} & 5.6 & 0.8 \\

\midrule
cracker (86) & FoundationPose & \textbf{80.3} & 91.2 & 94.5 & \textbf{97.7} & \textbf{4.1} & 1.0 \\*
 & RGBTrack & 37.2 & 16.5 & 19.3 & 12.8 & 88.9 & 121.8 \\*
 & GigaPose & 16.9 & 1.1 & 25.7 & 1.2 & 118.8 & 277.4 \\*
 & FreePose & 15.3 & 0.0 & 0.0 & 0.0 & 158.3 & 228.6 \\*
 & Dynhor & 28.8 & 6.0 & 50.3 & 1.2 & 105.9 & 10.5 \\*
 & BIT & 37.8 & 43.8 & 64.5 & 16.3 & 25.0 & 28.1 \\*
 & \textbf{Ours} & 77.5 & \textbf{91.4} & \textbf{95.2} & \textbf{97.7} & 4.7 & \textbf{0.7} \\

\midrule
mustard (72) & FoundationPose & \textbf{95.4} & \textbf{96.1} & \textbf{97.5} & \textbf{100.0} & \textbf{2.8} & \textbf{0.3} \\*
 & RGBTrack & 78.9 & 81.8 & 90.6 & 72.2 & 3.7 & 1.8 \\*
 & GigaPose & 8.6 & 0.8 & 12.9 & 0.0 & 120.0 & 23.9 \\*
 & FreePose & 15.1 & 0.0 & 0.0 & 0.0 & 139.4 & 82.0 \\*
 & Dynhor & 47.1 & 50.3 & 74.7 & 18.1 & 44.6 & 5.2 \\*
 & BIT & 35.0 & 27.3 & 50.2 & 8.3 & 56.5 & 11.1 \\*
 & \textbf{Ours} & 93.4 & 95.0 & 97.2 & \textbf{100.0} & 4.9 & 0.4 \\

\midrule
sugar (97) & FoundationPose & 90.9 & 60.9 & 96.0 & 47.4 & 94.7 & 0.6 \\*
 & RGBTrack & 41.2 & 31.2 & 37.4 & 20.6 & 61.6 & 76.7 \\*
 & GigaPose & 47.3 & 21.9 & 60.2 & 2.1 & 120.4 & 10.4 \\*
 & FreePose & 14.3 & 0.0 & 0.0 & 0.0 & 131.9 & 47.4 \\*
 & Dynhor & 41.7 & 21.6 & 63.1 & 3.1 & 95.5 & 9.3 \\*
 & BIT & 8.1 & 9.5 & 14.4 & 3.1 & 61.6 & 31.2 \\*
 & \textbf{Ours} & \textbf{92.2} & \textbf{94.1} & \textbf{96.6} & \textbf{99.0} & \textbf{4.1} & \textbf{0.5} \\

\midrule
tomato (66) & FoundationPose & \textbf{89.1} & \textbf{91.9} & 96.4 & \textbf{97.0} & \textbf{8.5} & 0.6 \\*
 & RGBTrack & 68.1 & 71.4 & 85.4 & 42.4 & 8.9 & 2.8 \\*
 & GigaPose & 23.5 & 0.9 & 14.0 & 0.0 & 121.3 & 17.9 \\*
 & FreePose & 16.9 & 0.0 & 0.0 & 0.0 & 145.3 & 65.3 \\*
 & Dynhor & 34.8 & 6.9 & 43.7 & 0.0 & 132.4 & 10.7 \\*
 & BIT & 49.1 & 45.3 & 68.5 & 9.1 & 56.2 & 5.8 \\*
 & \textbf{Ours} & 89.0 & 90.2 & \textbf{96.6} & 81.8 & 14.6 & \textbf{0.5} \\

\addlinespace[6pt]
\end{longtable}
}

\paragraph{LINEMOD.}
The complete records report all fifteen object identities under the same two CAD
lanes and share the same fixed evaluation keys.
Table~\ref{tab:supp_linemod_objects} gives the per-object breakdown.
{\suppTableFont
\setlength{\tabcolsep}{3pt}
\renewcommand{\arraystretch}{1.08}
\begin{longtable}{@{}llrrrrrrr@{}}
\caption{Complete LINEMOD per-object results for SAM3D CAD and Metric CAD. All metrics except rotation and translation are percentages; $N$ is the number of scored frames. Best per-object values are bold.}\label{tab:supp_linemod_objects}\\
\toprule
Object & Method & $N$ & AR$\uparrow$ & \shortstack{ADD\\AUC$\uparrow$} & \shortstack{ADD-S\\AUC$\uparrow$} & ADD-$0.1d\uparrow$ & \shortstack{Rot.\\(deg.)$\downarrow$} & \shortstack{Trans.\\(cm)$\downarrow$} \\
\midrule\endfirsthead
\toprule
Object & Method & $N$ & AR$\uparrow$ & \shortstack{ADD\\AUC$\uparrow$} & \shortstack{ADD-S\\AUC$\uparrow$} & ADD-$0.1d\uparrow$ & \shortstack{Rot.\\(deg.)$\downarrow$} & \shortstack{Trans.\\(cm)$\downarrow$} \\
\midrule\endhead
\bottomrule\endfoot
\bottomrule\endlastfoot
\multicolumn{9}{c}{\textit{SAM3D CAD}}\\*
\midrule
ape & FoundationPose & 62 & 45.1 & 45.7 & 88.5 & 0.0 & 149.6 & 2.1 \\*
 & RGBTrack &  & 1.8 & 1.0 & 4.6 & 0.0 & 122.8 & 30.9 \\*
 & GigaPose &  & 22.3 & 1.9 & 11.1 & 0.0 & 151.2 & 27.4 \\*
 & FreePose &  & 13.2 & 0.7 & 5.5 & 0.0 & 132.9 & 24.0 \\*
 & Dynhor &  & \textbf{59.3} & \textbf{53.5} & 82.1 & \textbf{22.6} & \textbf{80.7} & 3.3 \\*
 & BIT &  & 7.0 & 0.7 & 1.4 & 0.0 & 154.5 & 830.5 \\*
 & \textbf{Ours} &  & 44.0 & 48.3 & \textbf{90.0} & 0.0 & 157.4 & \textbf{1.3} \\
\midrule
benchvise & FoundationPose & 61 & 28.4 & 0.0 & 71.2 & 0.0 & 137.7 & 7.4 \\*
 & RGBTrack &  & 11.1 & 0.0 & 0.0 & 0.0 & 137.4 & 175.7 \\*
 & GigaPose &  & 19.5 & 0.5 & 18.2 & 0.0 & 123.5 & 40.5 \\*
 & FreePose &  & 16.4 & 0.0 & 0.0 & 0.0 & 130.4 & 68.1 \\*
 & Dynhor &  & \textbf{73.3} & \textbf{83.7} & \textbf{92.5} & \textbf{86.9} & \textbf{7.1} & \textbf{1.5} \\*
 & BIT &  & 0.5 & 0.0 & 1.2 & 0.0 & 112.9 & 1689.8 \\*
 & \textbf{Ours} &  & 30.9 & 0.1 & 76.8 & 0.0 & 141.9 & 5.6 \\
\midrule
bowl & FoundationPose & 62 & 55.2 & 19.7 & 92.5 & 1.6 & 100.0 & 2.1 \\*
 & RGBTrack &  & 0.5 & 0.0 & 1.6 & 0.0 & 144.7 & 54.0 \\*
 & GigaPose &  & 39.0 & 15.9 & 76.6 & 0.0 & 98.8 & 4.8 \\*
 & FreePose &  & 16.7 & 1.4 & 31.6 & 0.0 & 124.1 & 13.9 \\*
 & Dynhor &  & \textbf{76.5} & \textbf{26.0} & 90.6 & \textbf{11.3} & 89.5 & 2.1 \\*
 & BIT &  & 0.8 & 0.0 & 1.5 & 0.0 & 148.0 & 15290.5 \\*
 & \textbf{Ours} &  & 55.2 & 22.8 & \textbf{92.7} & 3.2 & \textbf{87.9} & \textbf{2.1} \\
\midrule
camera & FoundationPose & 61 & 37.9 & 33.4 & 82.5 & 0.0 & 87.6 & 3.2 \\*
 & RGBTrack &  & 0.5 & 0.0 & 1.3 & 0.0 & 126.1 & 35.0 \\*
 & GigaPose &  & 21.4 & 7.4 & 26.8 & 0.0 & 54.5 & 27.2 \\*
 & FreePose &  & 15.5 & 1.6 & 26.5 & 0.0 & 118.9 & 16.0 \\*
 & Dynhor &  & \textbf{60.8} & 58.0 & 87.3 & \textbf{41.0} & 41.4 & 2.7 \\*
 & BIT &  & 0.9 & 1.0 & 1.6 & 0.0 & 122.9 & 3184.0 \\*
 & \textbf{Ours} &  & 44.3 & \textbf{69.5} & \textbf{88.2} & 1.6 & \textbf{26.8} & \textbf{2.4} \\
\midrule
can & FoundationPose & 60 & 32.6 & 19.3 & 71.8 & 0.0 & 109.0 & 5.3 \\*
 & RGBTrack &  & 0.0 & 0.0 & 0.0 & 0.0 & 136.4 & 76.6 \\*
 & GigaPose &  & 6.0 & 0.1 & 4.4 & 0.0 & 79.3 & 21.4 \\*
 & FreePose &  & 16.5 & 0.7 & 52.9 & 0.0 & 136.3 & 9.0 \\*
 & Dynhor &  & \textbf{68.8} & \textbf{70.4} & \textbf{90.4} & \textbf{75.0} & \textbf{35.3} & \textbf{1.8} \\*
 & BIT &  & 18.9 & 0.0 & 1.8 & 0.0 & 101.0 & 1125.7 \\*
 & \textbf{Ours} &  & 34.9 & 51.4 & 81.4 & 1.7 & 46.0 & 2.2 \\
\midrule
cat & FoundationPose & 59 & 46.8 & 68.4 & 90.3 & 0.0 & 61.4 & 1.7 \\*
 & RGBTrack &  & 1.5 & 0.7 & 2.3 & 0.0 & 148.8 & 49.3 \\*
 & GigaPose &  & 30.1 & 32.0 & 60.1 & 0.0 & 83.4 & 10.2 \\*
 & FreePose &  & 16.9 & 0.0 & 1.0 & 0.0 & 125.6 & 39.9 \\*
 & Dynhor &  & \textbf{81.7} & \textbf{84.8} & \textbf{93.4} & \textbf{78.0} & \textbf{11.5} & 1.3 \\*
 & BIT &  & 7.3 & 1.1 & 2.6 & 0.0 & 115.4 & 1003.8 \\*
 & \textbf{Ours} &  & 48.1 & 72.4 & 92.1 & 5.1 & 55.2 & \textbf{1.3} \\
\midrule
cup & FoundationPose & 62 & 45.9 & 37.1 & 90.9 & 0.0 & 123.1 & \textbf{1.8} \\*
 & RGBTrack &  & 0.9 & 0.0 & 3.3 & 0.0 & 105.0 & 37.9 \\*
 & GigaPose &  & 28.0 & 12.1 & 56.3 & 0.0 & 104.6 & 8.9 \\*
 & FreePose &  & 13.5 & 0.8 & 9.8 & 0.0 & 135.8 & 21.0 \\*
 & Dynhor &  & \textbf{66.7} & \textbf{69.5} & 89.2 & \textbf{45.2} & \textbf{33.6} & 2.3 \\*
 & BIT &  & 2.0 & 0.4 & 1.3 & 0.0 & 121.3 & 772.4 \\*
 & \textbf{Ours} &  & 47.4 & 47.3 & \textbf{91.2} & 0.0 & 97.5 & 2.2 \\
\midrule
driller & FoundationPose & 60 & 42.1 & 55.4 & 78.8 & 0.0 & \textbf{18.6} & 3.9 \\*
 & RGBTrack &  & 4.5 & 0.9 & 1.3 & 0.0 & 112.4 & 99.6 \\*
 & GigaPose &  & 19.6 & 18.1 & 40.4 & 8.3 & 65.4 & 17.3 \\*
 & FreePose &  & 4.1 & 0.0 & 14.7 & 0.0 & 121.8 & 20.5 \\*
 & Dynhor &  & \textbf{64.5} & \textbf{60.4} & \textbf{82.6} & \textbf{61.7} & 39.6 & \textbf{3.3} \\*
 & BIT &  & 1.1 & 0.9 & 1.3 & 0.0 & 130.7 & 2102.9 \\*
 & \textbf{Ours} &  & 41.3 & 55.6 & 79.5 & 0.0 & 18.7 & 3.8 \\
\midrule
duck & FoundationPose & 63 & 44.6 & \textbf{72.4} & 88.3 & 0.0 & \textbf{35.4} & 1.6 \\*
 & RGBTrack &  & 1.7 & 1.5 & 1.9 & 1.6 & 116.0 & 35.5 \\*
 & GigaPose &  & 22.4 & 14.6 & 38.0 & 0.0 & 65.1 & 17.3 \\*
 & FreePose &  & 15.9 & 9.7 & 39.2 & 0.0 & 117.4 & 11.4 \\*
 & Dynhor &  & \textbf{73.8} & 71.7 & \textbf{91.4} & \textbf{47.6} & 49.0 & 1.6 \\*
 & BIT &  & 6.1 & 1.2 & 1.4 & 0.0 & 99.1 & 780.6 \\*
 & \textbf{Ours} &  & 44.6 & 71.5 & 88.2 & 0.0 & 38.0 & \textbf{1.6} \\
\midrule
eggbox & FoundationPose & 63 & \textbf{60.3} & 49.5 & \textbf{92.6} & \textbf{52.4} & 83.5 & \textbf{1.3} \\*
 & RGBTrack &  & 3.2 & 0.1 & 4.0 & 0.0 & 95.6 & 45.9 \\*
 & GigaPose &  & 25.7 & 3.7 & 37.6 & 0.0 & 87.8 & 12.0 \\*
 & FreePose &  & 13.8 & 3.4 & 27.1 & 0.0 & 102.2 & 19.0 \\*
 & Dynhor &  & 58.6 & 38.5 & 80.7 & 20.6 & 80.9 & 4.4 \\*
 & BIT &  & 8.1 & 0.0 & 1.5 & 0.0 & 95.9 & 1800.1 \\*
 & \textbf{Ours} &  & 55.4 & \textbf{52.7} & 91.9 & 41.3 & \textbf{76.1} & 1.5 \\
\midrule
glue & FoundationPose & 61 & 38.6 & 18.9 & 86.1 & 4.9 & 150.7 & 4.0 \\*
 & RGBTrack &  & 1.8 & 0.7 & 1.3 & 0.0 & 97.8 & 116.9 \\*
 & GigaPose &  & 27.3 & 13.0 & 44.8 & 1.6 & 104.8 & 10.9 \\*
 & FreePose &  & 12.6 & 1.0 & 8.8 & 0.0 & 131.4 & 31.5 \\*
 & Dynhor &  & \textbf{54.2} & \textbf{47.2} & 68.5 & \textbf{23.0} & \textbf{54.3} & 7.3 \\*
 & BIT &  & 0.5 & 0.2 & 1.5 & 0.0 & 122.9 & 1212.6 \\*
 & \textbf{Ours} &  & 39.2 & 15.5 & \textbf{89.8} & 0.0 & 162.3 & \textbf{3.6} \\
\midrule
holepuncher & FoundationPose & 62 & 43.3 & 23.8 & 84.7 & 0.0 & 145.3 & 3.7 \\*
 & RGBTrack &  & 0.5 & 0.2 & 1.9 & 0.0 & 90.4 & 38.4 \\*
 & GigaPose &  & 20.1 & 3.5 & 25.8 & 0.0 & 120.0 & 14.0 \\*
 & FreePose &  & 9.7 & 2.7 & 25.2 & 0.0 & 124.4 & 16.6 \\*
 & Dynhor &  & \textbf{70.2} & \textbf{72.4} & \textbf{89.4} & \textbf{53.2} & \textbf{20.6} & \textbf{2.3} \\*
 & BIT &  & 14.1 & 0.2 & 1.4 & 0.0 & 97.1 & 689.5 \\*
 & \textbf{Ours} &  & 29.8 & 21.0 & 80.6 & 0.0 & 159.5 & 3.1 \\
\midrule
iron & FoundationPose & 58 & 45.5 & 57.0 & 83.1 & 17.2 & 44.2 & 3.1 \\*
 & RGBTrack &  & 1.0 & 0.1 & 2.7 & 0.0 & 91.1 & 31.3 \\*
 & GigaPose &  & 26.6 & 12.1 & 43.0 & 1.7 & 50.7 & 13.1 \\*
 & FreePose &  & 17.4 & 0.0 & 4.0 & 0.0 & 131.5 & 38.3 \\*
 & Dynhor &  & \textbf{51.4} & 33.8 & 73.2 & 29.3 & 89.8 & 5.3 \\*
 & BIT &  & 0.6 & 0.9 & 1.5 & 0.0 & 160.1 & 775.1 \\*
 & \textbf{Ours} &  & 45.5 & \textbf{63.2} & \textbf{85.7} & \textbf{32.8} & \textbf{39.7} & \textbf{2.0} \\
\midrule
lamp & FoundationPose & 29 & 30.8 & 35.5 & 72.3 & 0.0 & 35.4 & 4.8 \\*
 & RGBTrack &  & 0.8 & 0.0 & 0.9 & 0.0 & 68.1 & 83.5 \\*
 & GigaPose &  & 24.0 & 15.2 & 50.8 & 3.4 & 44.6 & 11.7 \\*
 & FreePose &  & 14.7 & 0.0 & 0.0 & 0.0 & 143.6 & 56.6 \\*
 & Dynhor &  & \textbf{66.7} & \textbf{76.3} & \textbf{87.9} & \textbf{82.8} & \textbf{11.7} & \textbf{2.1} \\*
 & BIT &  & 1.4 & 1.8 & 2.7 & 0.0 & 97.2 & 658.0 \\*
 & \textbf{Ours} &  & 26.0 & 35.1 & 70.1 & 0.0 & 33.6 & 5.2 \\
\midrule
phone & FoundationPose & 10 & 27.0 & 26.6 & 59.2 & 0.0 & 94.7 & 7.2 \\*
 & RGBTrack &  & 6.7 & 0.0 & 11.9 & 0.0 & 111.5 & 18.6 \\*
 & GigaPose &  & 10.7 & 2.2 & 25.9 & 0.0 & 90.5 & 16.5 \\*
 & FreePose &  & 10.7 & 0.0 & 5.3 & 0.0 & 107.3 & 39.0 \\*
 & Dynhor &  & \textbf{60.3} & \textbf{66.8} & \textbf{80.2} & \textbf{70.0} & \textbf{14.9} & \textbf{3.7} \\*
 & BIT &  & 18.3 & 7.6 & 9.1 & 10.0 & 98.4 & 1266.2 \\*
 & \textbf{Ours} &  & 29.0 & 26.9 & 74.2 & 0.0 & 90.1 & 4.4 \\
\midrule
\multicolumn{9}{c}{\textit{Metric CAD}}\\*
\midrule
ape & FoundationPose & 62 & 1.5 & 1.5 & 2.8 & 1.6 & 116.5 & 49.4 \\*
 & RGBTrack &  & 4.5 & 1.9 & 3.7 & 1.6 & 106.8 & 62.9 \\*
 & GigaPose &  & 87.8 & 83.9 & 88.3 & 82.3 & 12.7 & 7.0 \\*
 & FreePose &  & 13.7 & 0.0 & 0.5 & 0.0 & 132.1 & 24.1 \\*
 & Dynhor &  & 59.3 & 53.5 & 82.1 & 22.6 & 80.7 & 3.3 \\*
 & BIT &  & 10.5 & 1.5 & 1.5 & 1.6 & 99.7 & 656.3 \\*
 & \textbf{Ours} &  & \textbf{89.6} & \textbf{92.8} & \textbf{96.9} & \textbf{95.2} & \textbf{9.8} & \textbf{0.5} \\
\midrule
benchvise & FoundationPose & 61 & 2.7 & 1.6 & 5.5 & 1.6 & 120.6 & 199.6 \\*
 & RGBTrack &  & 5.0 & 1.3 & 3.7 & 1.6 & 126.9 & 101.4 \\*
 & GigaPose &  & 71.9 & 82.5 & 92.0 & 95.1 & \textbf{3.5} & 1.7 \\*
 & FreePose &  & 12.6 & 0.0 & 2.5 & 0.0 & 104.3 & 34.4 \\*
 & Dynhor &  & 73.3 & 83.7 & 92.5 & 86.9 & 7.1 & 1.5 \\*
 & BIT &  & 0.8 & 1.1 & 1.4 & 0.0 & 112.1 & 673.4 \\*
 & \textbf{Ours} &  & \textbf{82.6} & \textbf{93.2} & \textbf{96.3} & \textbf{96.7} & 5.0 & \textbf{0.6} \\
\midrule
bowl & FoundationPose & 62 & 2.2 & 1.2 & 2.4 & 0.0 & \textbf{84.0} & 43.7 \\*
 & RGBTrack &  & 6.1 & 0.5 & 2.5 & 0.0 & 94.8 & 141.6 \\*
 & GigaPose &  & 70.5 & 24.3 & 93.4 & 9.7 & 88.9 & 1.6 \\*
 & FreePose &  & 15.8 & 0.2 & 12.2 & 0.0 & 136.1 & 19.7 \\*
 & Dynhor &  & 76.5 & \textbf{26.0} & 90.6 & \textbf{11.3} & 89.5 & 2.1 \\*
 & BIT &  & 1.4 & 1.2 & 1.6 & 0.0 & 114.0 & 259.7 \\*
 & \textbf{Ours} &  & \textbf{84.1} & 21.7 & \textbf{96.5} & 9.7 & 95.9 & \textbf{0.8} \\
\midrule
camera & FoundationPose & 61 & 1.6 & 1.6 & 1.6 & 1.6 & 97.5 & 59.8 \\*
 & RGBTrack &  & 1.5 & 1.5 & 3.0 & 1.6 & 89.9 & 41.5 \\*
 & GigaPose &  & \textbf{89.3} & 91.9 & 95.6 & 98.4 & \textbf{2.6} & 0.7 \\*
 & FreePose &  & 17.7 & 0.4 & 3.6 & 0.0 & 95.4 & 31.5 \\*
 & Dynhor &  & 60.8 & 58.0 & 87.3 & 41.0 & 41.4 & 2.7 \\*
 & BIT &  & 14.6 & 1.3 & 2.0 & 1.6 & 122.7 & 1293.0 \\*
 & \textbf{Ours} &  & 87.0 & \textbf{93.9} & \textbf{96.5} & \textbf{100.0} & 3.8 & \textbf{0.5} \\
\midrule
can & FoundationPose & 60 & 2.7 & 1.6 & 3.1 & 1.7 & 113.2 & 38.2 \\*
 & RGBTrack &  & 3.0 & 1.8 & 4.6 & 1.7 & 117.0 & 118.5 \\*
 & GigaPose &  & 83.6 & 90.2 & 95.4 & 96.7 & 4.6 & 0.8 \\*
 & FreePose &  & 14.2 & 1.1 & 21.4 & 0.0 & 129.3 & 20.5 \\*
 & Dynhor &  & 68.8 & 70.4 & 90.4 & 75.0 & 35.3 & 1.8 \\*
 & BIT &  & 22.2 & 1.5 & 2.4 & 1.7 & 129.2 & 1310.6 \\*
 & \textbf{Ours} &  & \textbf{86.9} & \textbf{94.2} & \textbf{96.7} & \textbf{98.3} & \textbf{3.1} & \textbf{0.5} \\
\midrule
cat & FoundationPose & 59 & 2.0 & 1.7 & 4.2 & 1.7 & 112.2 & 93.6 \\*
 & RGBTrack &  & 1.7 & 0.5 & 1.3 & 0.0 & 127.7 & 75.4 \\*
 & GigaPose &  & 89.8 & 91.3 & 95.7 & 96.6 & \textbf{2.4} & 0.9 \\*
 & FreePose &  & 17.7 & 0.0 & 0.4 & 0.0 & 127.7 & 44.4 \\*
 & Dynhor &  & 81.7 & 84.8 & 93.4 & 78.0 & 11.5 & 1.3 \\*
 & BIT &  & 8.3 & 1.5 & 1.6 & 1.7 & 104.0 & 1030.2 \\*
 & \textbf{Ours} &  & \textbf{90.9} & \textbf{95.8} & \textbf{97.5} & \textbf{100.0} & 3.3 & \textbf{0.4} \\
\midrule
cup & FoundationPose & 62 & 2.0 & 1.9 & 3.0 & 1.6 & 116.6 & 70.4 \\*
 & RGBTrack &  & 2.2 & 0.4 & 2.2 & 0.0 & 113.8 & 29.5 \\*
 & GigaPose &  & 51.4 & 45.8 & 76.5 & 9.7 & 47.9 & 4.7 \\*
 & FreePose &  & 14.8 & 0.0 & 0.5 & 0.0 & 139.9 & 34.4 \\*
 & Dynhor &  & 66.7 & 69.5 & 89.2 & 45.2 & 33.6 & 2.3 \\*
 & BIT &  & 7.3 & 1.4 & 1.5 & 1.6 & 118.4 & 1478.2 \\*
 & \textbf{Ours} &  & \textbf{83.0} & \textbf{80.9} & \textbf{96.9} & \textbf{67.7} & \textbf{32.4} & \textbf{0.9} \\
\midrule
driller & FoundationPose & 60 & 4.4 & 1.6 & 3.4 & 1.7 & 117.0 & 23.5 \\*
 & RGBTrack &  & 0.7 & 0.0 & 0.2 & 0.0 & 115.3 & 221.0 \\*
 & GigaPose &  & 72.4 & 74.1 & 84.6 & 81.7 & 17.6 & 3.5 \\*
 & FreePose &  & 9.4 & 0.0 & 4.1 & 0.0 & 124.1 & 33.1 \\*
 & Dynhor &  & 64.5 & 60.4 & 82.6 & 61.7 & 39.6 & 3.3 \\*
 & BIT &  & 18.3 & 1.6 & 1.6 & 1.7 & 99.1 & 2120.1 \\*
 & \textbf{Ours} &  & \textbf{85.6} & \textbf{93.9} & \textbf{96.5} & \textbf{100.0} & \textbf{1.9} & \textbf{0.6} \\
\midrule
duck & FoundationPose & 63 & 1.5 & 1.5 & 1.6 & 1.6 & 93.2 & 70.4 \\*
 & RGBTrack &  & 1.7 & 1.5 & 1.9 & 1.6 & 88.4 & 40.5 \\*
 & GigaPose &  & \textbf{89.0} & 87.6 & 92.5 & 84.1 & \textbf{8.7} & 1.7 \\*
 & FreePose &  & 16.7 & 0.2 & 1.4 & 0.0 & 108.6 & 32.0 \\*
 & Dynhor &  & 73.8 & 71.7 & 91.4 & 47.6 & 49.0 & 1.6 \\*
 & BIT &  & 1.5 & 1.5 & 1.5 & 1.6 & 112.2 & 890.6 \\*
 & \textbf{Ours} &  & 85.0 & \textbf{90.4} & \textbf{96.2} & \textbf{88.9} & 12.7 & \textbf{0.6} \\
\midrule
eggbox & FoundationPose & 63 & 1.4 & 1.5 & 1.8 & 1.6 & 118.9 & 49.6 \\*
 & RGBTrack &  & 2.3 & 1.2 & 4.6 & 0.0 & 108.9 & 27.7 \\*
 & GigaPose &  & 80.6 & 81.0 & 91.1 & 71.4 & 8.4 & 3.4 \\*
 & FreePose &  & 14.9 & 0.0 & 1.6 & 0.0 & 146.3 & 34.2 \\*
 & Dynhor &  & 58.6 & 38.5 & 80.7 & 20.6 & 80.9 & 4.4 \\*
 & BIT &  & 1.4 & 1.5 & 1.5 & 1.6 & 102.7 & 1730.4 \\*
 & \textbf{Ours} &  & \textbf{91.6} & \textbf{94.9} & \textbf{97.0} & \textbf{100.0} & \textbf{3.2} & \textbf{0.4} \\
\midrule
glue & FoundationPose & 61 & 1.7 & 1.6 & 2.4 & 1.6 & 111.2 & 217.8 \\*
 & RGBTrack &  & 16.6 & 1.0 & 1.5 & 0.0 & 99.1 & 135.6 \\*
 & GigaPose &  & 73.7 & 76.1 & 82.8 & 72.1 & 14.7 & 5.0 \\*
 & FreePose &  & 12.7 & 0.6 & 4.9 & 0.0 & 144.9 & 37.4 \\*
 & Dynhor &  & 54.2 & 47.2 & 68.5 & 23.0 & 54.3 & 7.3 \\*
 & BIT &  & 4.0 & 1.5 & 1.6 & 1.6 & 122.9 & 1442.9 \\*
 & \textbf{Ours} &  & \textbf{81.7} & \textbf{90.4} & \textbf{95.6} & \textbf{90.2} & \textbf{13.2} & \textbf{0.7} \\
\midrule
holepuncher & FoundationPose & 62 & 2.7 & 1.5 & 4.0 & 1.6 & 127.4 & 52.6 \\*
 & RGBTrack &  & 1.8 & 0.0 & 8.5 & 0.0 & 110.7 & 25.4 \\*
 & GigaPose &  & 57.8 & 60.3 & 84.5 & 19.4 & 31.0 & 3.3 \\*
 & FreePose &  & 19.8 & 0.0 & 0.2 & 0.0 & 118.6 & 36.6 \\*
 & Dynhor &  & 70.2 & 72.4 & 89.4 & 53.2 & 20.6 & 2.3 \\*
 & BIT &  & 16.8 & 1.6 & 1.6 & 1.6 & 93.8 & 1361.8 \\*
 & \textbf{Ours} &  & \textbf{90.4} & \textbf{94.4} & \textbf{97.1} & \textbf{100.0} & \textbf{3.9} & \textbf{0.5} \\
\midrule
iron & FoundationPose & 58 & 4.0 & 2.8 & 7.0 & 1.7 & 114.4 & 154.1 \\*
 & RGBTrack &  & 1.5 & 0.7 & 2.9 & 0.0 & 108.9 & 74.7 \\*
 & GigaPose &  & 74.4 & 88.7 & 93.9 & \textbf{98.3} & 2.9 & 1.1 \\*
 & FreePose &  & 20.5 & 0.0 & 0.1 & 0.0 & 133.1 & 50.9 \\*
 & Dynhor &  & 51.4 & 33.8 & 73.2 & 29.3 & 89.8 & 5.3 \\*
 & BIT &  & 1.4 & 1.5 & 1.6 & 1.7 & 106.6 & 2398.8 \\*
 & \textbf{Ours} &  & \textbf{76.4} & \textbf{91.8} & \textbf{95.1} & \textbf{98.3} & \textbf{2.6} & \textbf{0.8} \\
\midrule
lamp & FoundationPose & 29 & 4.6 & 3.1 & 5.0 & 3.4 & 116.3 & 42.5 \\*
 & RGBTrack &  & 1.5 & 1.4 & 2.7 & 0.0 & 65.3 & 80.5 \\*
 & GigaPose &  & 58.0 & 75.6 & 88.4 & 79.3 & \textbf{4.0} & 2.3 \\*
 & FreePose &  & 10.3 & 0.0 & 2.4 & 0.0 & 108.2 & 46.9 \\*
 & Dynhor &  & 66.7 & 76.3 & 87.9 & 82.8 & 11.7 & 2.1 \\*
 & BIT &  & 2.9 & 3.1 & 5.0 & 3.4 & 132.0 & 346.0 \\*
 & \textbf{Ours} &  & \textbf{72.6} & \textbf{87.6} & \textbf{93.8} & \textbf{93.1} & 7.4 & \textbf{0.9} \\
\midrule
phone & FoundationPose & 10 & 12.0 & 9.6 & 12.6 & 10.0 & 121.0 & 25.2 \\*
 & RGBTrack &  & 9.7 & 1.5 & 15.7 & 0.0 & 101.0 & 19.0 \\*
 & GigaPose &  & \textbf{78.3} & 80.8 & 85.2 & \textbf{90.0} & \textbf{13.2} & 22.7 \\*
 & FreePose &  & 10.0 & 0.0 & 3.3 & 0.0 & 91.9 & 34.3 \\*
 & Dynhor &  & 60.3 & 66.8 & 80.2 & 70.0 & 14.9 & 3.7 \\*
 & BIT &  & 9.3 & 9.0 & 9.4 & 10.0 & 103.0 & 1122.6 \\*
 & \textbf{Ours} &  & 76.7 & \textbf{81.0} & \textbf{93.6} & 80.0 & 22.4 & \textbf{1.5} \\
\midrule
\end{longtable}
}

\FloatBarrier

\subsection{Matched Controls for the Main-Text Ablations}
\label{sec:supp_matched_ablations}

The following controls use the same 23 HOT3D recording--asset streams, three
all-frame adaptation seeds, SAM3D CAD, RGB-derived
VGGT-$\Omega$ scene evidence, candidate bank, and solver settings as
the main-text component ablation.  They isolate the stated
mechanisms before final pose evaluation.

\paragraph{Pose-conditioned correspondence.}
For each hard object-to-proxy-3D assignment, Inlier@$0.1d$ counts transformed
canonical surface points within one tenth of the object diameter of the assigned
proxy point. The duplicate-assignment rate uses the same hard assignments.
The 3D residual separately summarizes the correspondence-weighted
3D error normalized by object diameter, before solver selection and temporal
filtering. We average each metric over the three
seeds within a recording--asset stream and bootstrap the 23 streams with 10,000
paired draws.

\paragraph{Factorized solver support and selection.}
The solver diagnostic reads frozen candidate measurements and posterior
selections.  Best
residual is the minimum correspondence residual among retained candidates in a
frame, whereas selected residual is measured for the candidate selected by the
posterior decoder.
The 3D--3D control retains its geometric parent when independent alignment
is unsupported. Its 65.3\% alignment-selection rate therefore describes
the selected branch, rather than the availability of a 3D--3D estimate on
every frame. All variants retain the same scored frames.

\begin{table}[tbp]
\caption{Matched mechanism controls on the shared HOT3D recording--asset streams under SAM3D CAD and
adaptation seeds.  (a) Values and intervals are percentages; intervals report
the full model minus the appearance-and-geometry baseline.  (b) Residuals are
percentages of object diameter.}
\centering
\suppTableFont
\setlength{\tabcolsep}{4pt}
\renewcommand{\arraystretch}{1.12}
\begin{minipage}[t]{\linewidth}
\centering
\textbf{(a) Pose-conditioned correspondence}\\[-2pt]
\suppfit{\linewidth}{%
\begin{tabular}{lcccc}
\toprule
Metric & \shortstack{w/o pose\\cond.} & \shortstack{+ pose\\cond.} & + contrastive & \shortstack{Full--base\\95\% CI} \\
\midrule
Inlier@$0.1d$ $\uparrow$ & 13.94 & 19.35 & \textbf{20.34} & [+4.83, +7.95] \\
Duplicate assignment $\downarrow$ & 37.04 & 34.32 & \textbf{30.98} & [-7.52, -4.62] \\
3D residual$/d$ $\downarrow$ & 58.42 & 54.78 & \textbf{49.99} & [-11.46, -5.58] \\
Initial solver residual$/d$ $\downarrow$ & 39.01 & 37.62 & \textbf{36.92} & [-3.47, -0.56] \\
\bottomrule
\end{tabular}
}
\end{minipage}\par\medskip
\begin{minipage}[t]{\linewidth}
\centering
\textbf{(b) Factorized solver support and selection}\\[-2pt]
\suppfit{\linewidth}{%
\begin{tabular}{lcccc}
\toprule
Variant & \shortstack{Best residual$/d$\\$\downarrow$} & \shortstack{Selected residual$/d$\\$\downarrow$} & \shortstack{Candidates/frame\\$\uparrow$} & \shortstack{3D--3D selected\\$\uparrow$} \\
\midrule
2D--3D only & 21.64 & 24.25 & 64.0 & 0.0 \\
3D--3D only & \underline{19.75} & \textbf{22.52} & 64.0 & \textbf{65.3} \\
Direct pose average & 20.48 & 22.98 & 64.0 & 0.0 \\
\textbf{Delayed solver union} & \textbf{19.74} & \underline{22.63} & \textbf{86.1} & 52.4 \\
\bottomrule
\end{tabular}
}
\end{minipage}

\label{tab:supp_pose_conditioning_matched}
\label{tab:supp_solver_matched}
\end{table}

Pose conditioning improves the reciprocal assignments before solver selection,
and the contrastive objective further concentrates them.  The
intervals exclude zero for inlier rate, duplicate assignment, 3D
residual, and the solver-input residual, linking the correspondence change to
the geometric measurements used by FGH-Solver.

Independent 3D--3D alignment supplies a retained candidate for 34.47\% of
parent candidates and differs from its paired 2D--3D candidate by a median
$25.75^\circ$ and 2.47~cm.  Delayed solver union retains both candidates until
posterior decoding; it selects 3D--3D support on 52.4\% of frames and 2D--3D
support on the remaining 47.6\%, rather than averaging their poses before
selection.

\subsection{Additional Metric CAD Controls}
\label{sec:supp_metric_cad_controls}

The preceding tables provide the complete common-output Metric CAD results.
We retain the additional LINEMOD controls below because they distinguish
independent per-key registration from native continuous tracking in their
stated inference families.

\begin{table}[tbp]
\caption{LINEMOD pose quality with Metric CAD on the same 833 evaluation keys.
Inference family denotes sequence-level posterior inference, per-key
registration, or independent framewise estimation. ADD and ADD-S are AUC
percentages; translation is in metres.}
\centering
\suppTableFont
\setlength{\tabcolsep}{4pt}
\suppfit{\textwidth}{%
\begin{tabularx}{\linewidth}{@{}>{\raggedright\arraybackslash}p{0.26\linewidth} >{\raggedright\arraybackslash}X ccccc@{}}
\toprule
Method & Inference family & AR $\uparrow$ & \shortstack{ADD\\AUC $\uparrow$} & \shortstack{ADD-S\\AUC $\uparrow$} & \shortstack{Rot.\\(deg.) $\downarrow$} & \shortstack{Trans.\\(m) $\downarrow$} \\
\midrule
\textbf{Ours} & offline posterior inference & 0.856 & 86.7 & 96.4 & 14.69 & 0.0061 \\
FoundationPose-Register & independent per-key registration & \textbf{0.897} & \textbf{89.0} & \textbf{97.1} & \textbf{11.32} & \textbf{0.0051} \\
FoundationPose-Register+Posterior & register + shared posterior decoder & 0.894 & 87.9 & \textbf{97.1} & 13.66 & 0.0054 \\
RGBTrack-Register & independent per-key registration & 0.591 & 54.6 & 77.3 & 27.80 & 0.0482 \\
GigaPose & independent framewise estimation & 0.760 & 75.1 & 89.6 & 18.51 & 0.0297 \\
\bottomrule
\end{tabularx}}

\label{tab:linemod_tracking}
\end{table}

\begin{table}[tbp]
\caption{Native no-reset continuous-video stress test over all 16,570
LINEMOD frames, evaluated on the same fixed keys.}
\centering
\suppTableFont
\setlength{\tabcolsep}{4pt}
\suppfit{\columnwidth}{%
\begin{tabular}{lccccc}
\toprule
Native tracker & AR $\uparrow$ & \shortstack{ADD\\AUC $\uparrow$} & \shortstack{ADD-S\\AUC $\uparrow$} & \shortstack{Rot.\\(deg.) $\downarrow$} & \shortstack{Trans.\\(m) $\downarrow$} \\
\midrule
FoundationPose & 0.024 & 1.8 & 3.4 & 111.22 & 0.8369 \\
RGBTrack & 0.038 & 1.0 & 3.3 & 106.61 & 0.8288 \\
FreePose & 0.150 & 0.2 & 4.0 & 125.17 & 0.3370 \\
Dynhor & \textbf{0.665} & \textbf{59.9} & \textbf{85.5} & \textbf{47.12} & \textbf{0.0299} \\
\bottomrule
\end{tabular}}

\label{tab:linemod_native_stress}
\end{table}

BIT results for the two CAD conditions are reported in
Table~\ref{tab:supp_all_protocol_aggregate};
Table~\ref{tab:linemod_native_stress} lists the separate continuous-video
stress-test controls.

All component rows below use the same 378-key YCBInEOAT Metric CAD control,
candidate budget, evaluator, and sequence-specific geometric targets.

\begin{table}[tbp]
\caption{Component controls on the fixed YCBInEOAT Metric CAD protocol.
Panel (a) gives the sequential system assembly; panels (b) and (c) separate
solver support and posterior selection from frame-level emission and
sequence-specific self-adaptation.  ADD and ADD-S are AUC percentages, translation is in meters,
and all rows use identical frame keys.}
\centering
\suppTableFont
\setlength{\tabcolsep}{4pt}
\renewcommand{\arraystretch}{1.12}
\textbf{(a) System assembly}\par\vspace{2pt}
\begin{tabular*}{\textwidth}{@{\extracolsep{\fill}}lccccc@{}}
\toprule
Configuration & AR $\uparrow$ & ADD AUC $\uparrow$ & ADD-S AUC $\uparrow$ & Rot. (deg.) $\downarrow$ & Trans. (m) $\downarrow$\\
\midrule
GeoCorr-Matcher + FGH-Solver & .6475 & 37.1 & 79.3 & 109.12 & .0842\\
$+$ posterior-gated surface memory & .8174 & 83.4 & 92.8 & 21.55 & .0110\\
$+$ temporal Bayesian filtering & .8436 & 90.6 & 94.9 & 7.49 & .0076\\
$+$ dense silhouette/RGB measurement & .8693 & 92.3 & 96.1 & 6.56 & .0055\\
$+$ trust-region rotation refinement & \textbf{.8738} & \textbf{92.5} & \textbf{96.2} & \textbf{6.45} & \textbf{.0054}\\
\bottomrule
\end{tabular*}

\vspace{5pt}
\begin{minipage}[t]{\linewidth}
\centering
\textbf{(b) Solver support and selection}\par\vspace{2pt}
\begin{tabular*}{\linewidth}{@{\extracolsep{\fill}}lcc@{}}
\toprule
Variant & AR $\uparrow$ & Rot. (deg.) $\downarrow$ \\
\midrule
\multicolumn{3}{@{}l}{\textit{Solver support}} \\
2D--3D only & .4525 & 105.48 \\
Joint support only & .5376 & 73.62 \\
Propagated candidate only & .8293 & 9.39 \\
3D--3D only & .8725 & 5.71 \\
\midrule
\multicolumn{3}{@{}l}{\textit{Posterior selection}} \\
Direct pose avg. & .8041 & 32.00 \\
Early posterior selection & .8735 & \textbf{5.70} \\
Delayed posterior selection & .8734 & 6.14 \\
Delayed solver union & \textbf{.8742} & 6.14 \\
\bottomrule
\end{tabular*}
\end{minipage}\par\medskip
\begin{minipage}[t]{\linewidth}
\centering
\textbf{(c) Emission and self-adaptation}\par\vspace{2pt}
\begin{tabular*}{\linewidth}{@{\extracolsep{\fill}}lccc@{}}
\toprule
Variant & AR $\uparrow$ & ADD AUC $\uparrow$ & Rot. (deg.) $\downarrow$ \\
\midrule
\multicolumn{4}{@{}l}{\textit{Frame-level emission}} \\
Feature emission & .8578 & 91.90 & 6.78 \\
Joint emission & \textbf{.8768} & \textbf{93.00} & \textbf{5.65} \\
\midrule
\multicolumn{4}{@{}l}{\textit{Sequence-specific self-adaptation}} \\
Full $\mathcal L_{\mathrm{seq}}$ & .8311 & 83.99 & 19.69 \\
w/o $\mathcal L_{\mathrm{post}}$ & .8102 & 72.86 & 46.00 \\
w/o $\mathcal L_{\mathrm{geom}}$ & .8118 & 82.80 & 21.09 \\
\bottomrule
\end{tabular*}
\end{minipage}

\label{tab:ablation_full}
\label{tab:supp_solver}
\label{tab:supp_measurement}
\end{table}

The sequential assembly summarizes the complete Metric CAD system as its
modules are enabled.  The matched HOT3D controls isolate pose-conditioned
correspondence and FGH-Solver support; panel (a) records their integration with
posterior-gated surface memory, temporal Bayesian filtering, dense
silhouette/RGB measurement, and trust-region rotation refinement.

3D--3D support is already strong on this calibrated benchmark, while the
2D--3D, joint support, and propagated candidates retain complementary candidates.  The
critical control is posterior selection over their shared support: direct pose averaging
loses 7.01 AR points and raises rotation error by $25.86^\circ$, while early,
delayed, and union results remain tightly grouped.  The delayed solver union thus
preserves a valid mode for measurement rather than requiring one solver family
to dominate every scene.

Appearance and geometry provide complementary candidate evidence: their joint
emission adds 1.90 AR points and 1.10 ADD-AUC points over feature emission
alone.  The self-adaptation controls act at a different stage.  Removing
$\mathcal L_{\mathrm{post}}$ reduces AR by 2.09 points and raises rotation error
from $19.69^\circ$ to $46.00^\circ$; removing $\mathcal L_{\mathrm{geom}}$
reduces AR by 1.93 points.  Together they connect sequence-specific
self-adaptation to the geometric support consumed by the posterior decoder.

\begin{table}[tbp]
\caption{Posterior-gated surface memory and sequence-specific self-adaptation
on the fixed YCBInEOAT Metric CAD protocol.  Observability is the learned
reliability weight used in temporal filtering; memory values are scene means
over three matched seeds.}
\centering
\suppTableFont
\setlength{\tabcolsep}{4pt}
\renewcommand{\arraystretch}{1.12}
\begin{minipage}[t]{\linewidth}
\centering
\textbf{(a) Posterior-gated surface memory}\par\vspace{2pt}
\begin{tabular*}{\linewidth}{@{\extracolsep{\fill}}lccc@{}}
\toprule
Memory state & Write prob. & Observability & Pose update \\
\midrule
No memory & .0000 & .2508 & $4.178^\circ$/ .01121 m \\
Gated memory & \textbf{.3351} & \textbf{.3365} & \textbf{$4.072^\circ$/ .01072 m} \\
\bottomrule
\end{tabular*}
\end{minipage}\par\medskip
\begin{minipage}[t]{\linewidth}
\centering
\textbf{(b) Sequence-specific self-adaptation}\par\vspace{2pt}
\begin{tabular*}{\linewidth}{@{\extracolsep{\fill}}lcccc@{}}
\toprule
Protocol & AR $\uparrow$ & \shortstack{ADD\\AUC} $\uparrow$ & \shortstack{ADD-S\\AUC} $\uparrow$ & \shortstack{Rot.\\(deg.)} $\downarrow$ \\
\midrule
All-frame adaptation & .8317 & 84.0 & 93.2 & 19.94 \\
Prefix-causal adaptation & .8184 & 78.8 & 93.2 & 31.24 \\
Temporal cross-fit & .8325 & 83.0 & 93.1 & 22.81 \\
Cross-fit + Viterbi & \textbf{.8382} & \textbf{84.0} & \textbf{93.9} & \textbf{20.14} \\
\bottomrule
\end{tabular*}
\end{minipage}

\label{tab:supp_memory}
\label{tab:supp_adaptation}
\end{table}

Posterior-gated surface memory raises observability by 34.2\% relative while
reducing both rotation and translation update magnitudes.  Reliable
observations therefore become more available to temporal filtering without
broadening the local pose correction.

Cross-fit + Viterbi gives the strongest cross-fit result (.8382 AR), while
all-frame and temporal cross-fit results remain within .0008 AR.  Prefix-causal
adaptation is stricter because only the observed prefix can shape adaptation.
The HOT3D causal audit in Table~\ref{tab:supp_causal} shows that prefix-causal
and all-frame adaptation differ by .0003 under trust-region rotation
refinement.

\subsection{Additional Controls}
\label{sec:supp_additional_controls}

The remaining controls examine candidate budget, temporal posterior selection,
causal execution, sensitivity to SAM3D CAD and reconstructed scene
evidence, seed variation, optimization stability, and stage-resolved runtime.
They use the fixed protocols stated above and complement the matched controls
without changing the main comparison setting.

\paragraph{Candidate budget and temporal posterior.}
\label{sec:supp_candidate_budget_v24}
\label{sec:supp_posterior_diversity}

We vary only the number of pose candidates retained from the same fixed geometric
bank.  All rows use the same 23 HOT3D recording--assets, 476 scored keys,
SAM3D CAD, optimizer, random seed, trust-region rotation refinement, and BOP evaluator.
The raw bank contains 223.5 pose candidates per frame on average; every evaluated
frame has at least 127 raw candidates, so $K\in\{16,32,64\}$ preserves frame
coverage exactly.

We distinguish the retained candidate count from the effective support
$H_{\mathrm{eff}}=\exp(\mathcal H(q))$.  The former measures which geometric
candidates reach temporal filtering; the latter measures posterior
concentration after frame-level emission and temporal Bayesian filtering.

\begin{table}[tbp]
\caption{Candidate budget and temporal posterior on HOT3D with SAM3D
CAD. Panel (a) reports median sequence-specific self-adaptation time over 23
recording--assets. Panel (b) reports effective posterior support over 195
frames and three adaptation seeds with $K=64$.}
\centering
\suppTableFont
\setlength{\tabcolsep}{4pt}
\renewcommand{\arraystretch}{1.12}
\begin{minipage}[t]{\linewidth}
\centering
\textbf{(a) Candidate budget}\par\vspace{2pt}
\begin{tabular*}{\linewidth}{@{\extracolsep{\fill}}lcccc@{}}
\toprule
Support & Raw bank & AR $\uparrow$ & AR$_{\rm MSSD}$ $\uparrow$ & Adapt. (s) $\downarrow$ \\
\midrule
Geometric parent & -- & .3399 & .2086 & -- \\
$K=16$ & 7.2\% & .3398 & .2086 & 29.31 \\
$K=32$ & 14.3\% & .3400 & .2086 & \textbf{29.25} \\
$K=64$ & 28.6\% & \textbf{.3401} & \textbf{.2090} & 29.80 \\
\bottomrule
\end{tabular*}
\end{minipage}\par\medskip
\begin{minipage}[t]{\linewidth}
\centering
\textbf{(b) Posterior support}\par\vspace{2pt}
\begin{tabular*}{\linewidth}{@{\extracolsep{\fill}}lccc@{}}
\toprule
Stage & $H_{\mathrm{eff}}$ & $P(H_{\mathrm{eff}}>1.5)$ & $P(H_{\mathrm{eff}}>2)$ \\
\midrule
Frame-level emission & 2.43 & .342 & .243 \\
Temporal posterior & 1.27 & .138 & .084 \\
\bottomrule
\end{tabular*}
\end{minipage}

\label{tab:supp_candidate_budget_v24}
\label{tab:supp_posterior_support}
\end{table}

The complete accuracy span is .0003 AR and the adaptation-time span is 0.55 s,
showing that the fixed network and sequence-specific self-adaptation dominate cost
once the support bank is available.  More importantly, the response is
localized: 21/23 recording--assets are identical for all three budgets.
$K=64$ improves the ambiguity-rich cellphone and mug sequences by .0009 and
.0032 AR relative to the geometric parent, respectively, whereas $K=16$
reduces the mug sequence by .0016.  These are also among the scenes with high
framewise effective support in the posterior support control.  The retained
budget therefore acts as capacity for unresolved solver modes, while the
posterior concentrates to an effective support near one on unambiguous frames.

Temporal transport changes the framewise candidate on 20.5\% of all frames and
25.6\% of frames with valid motion.  Those changed frames have substantially
larger framewise effective support than unchanged frames (6.19 versus 1.46),
showing that temporal evidence acts where the single-frame emission is most
ambiguous.  After transport, their effective support remains 1.50 versus 1.21.
The final candidate is identical across all three seeds on 95.4\% of unique
frames, connecting ambiguity preservation with stable posterior selection.

The sequence-level response agrees with the candidate-budget control.  The
P0014\_\allowbreak 9a25ec6a/\allowbreak cellphone and P0014\_\allowbreak 9b7a0725/\allowbreak mug-white sequences have framewise effective support
2.90 and 4.00 and temporal-choice rates .269 and .143, respectively.  They are
also the two recording--assets improved by the $K=64$ support.  Temporal
transport therefore uses the additional bank capacity in the sequences where
single-frame evidence remains multi-modal, while concentrating near one mode
elsewhere.

\paragraph{Causal execution and trust-region rotation refinement.}
\label{sec:supp_causal_trust}

For strict causal evaluation, every target chunk is excluded from adaptation,
temporal filtering uses only the observed prefix, and the initial chunk uses
the fixed geometric parent.  The experiment contains 195 frames from nine
recording--assets; 132 frames have a learned historical prefix.

Panel (b) isolates the components allowed to change in the learned pose
update. Its geometric parent has AR .8738; the rotation-only output reaches
.8743. This is a separate control from the system assembly in
Table~\ref{tab:ablation_full}, whose last row reports .8738.

\begin{table}[tbp]
\caption{Causal execution and trust-region rotation refinement. Panel (a)
uses the same candidate bank and scored HOT3D keys. Panel (b) uses YCBInEOAT
with Metric CAD; all rows reuse the same fixed posterior update and differ only
in the refined pose components.}
\centering
\suppTableFont
\setlength{\tabcolsep}{4pt}
\renewcommand{\arraystretch}{1.12}
\begin{minipage}[t]{\linewidth}
\centering
\textbf{(a) Causal execution}\par\vspace{2pt}
\begin{tabular*}{\linewidth}{@{\extracolsep{\fill}}lccc@{}}
\toprule
Mode & AR $\uparrow$ & AR$_{\mathrm{MSPD}}$ $\uparrow$ & AR$_{\mathrm{MSSD}}$ $\uparrow$ \\
\midrule
Geometric parent & .5551 & .8262 & .2841 \\
All-frame adaptation & \textbf{.5554} & .8262 & \textbf{.2846} \\
Prefix-causal adaptation & .5551 & .8262 & .2841 \\
\bottomrule
\end{tabular*}
\end{minipage}\par\medskip
\begin{minipage}[t]{\linewidth}
\centering
\textbf{(b) Trust-region rotation refinement}\par\vspace{2pt}
\begin{tabular*}{\linewidth}{@{\extracolsep{\fill}}lccc@{}}
\toprule
Refinement & AR $\uparrow$ & ADD AUC $\uparrow$ & Rot. (deg.) $\downarrow$ \\
\midrule
Geometric parent & .8738 & 92.5 & $6.45^{\circ}$ \\
Rotation only & \textbf{.8743} & \textbf{92.5} & $\mathbf{6.43}^{\circ}$ \\
Rotation + translation & .8711 & 92.4 & $\mathbf{6.43}^{\circ}$ \\
Rotation + translation + scale & .8711 & 92.4 & $\mathbf{6.43}^{\circ}$ \\
Unrestricted refinement & .8688 & 92.3 & $6.44^{\circ}$ \\
\bottomrule
\end{tabular*}
\end{minipage}

\label{tab:supp_causal}
\label{tab:supp_pose_subspace}
\end{table}

The prefix-causal result preserves the all-frame score within .0003, and its
paired difference from the geometric parent is $.0000$ with 95\% CI
$[-.0005,.0005]$.  Thus future-frame exclusion has negligible effect, while
trust-region rotation refinement bounds learned updates during short-prefix
operation.

Seven of the nine recording--assets are identical under all-frame and causal
execution.  On P0014\_9a25ec6a/cellphone, both modes improve the
geometric parent by .0014 AR.  On P0014\_9b7a0725/mug-white,
all-frame adaptation adds .0024 while prefix-causal adaptation returns to the parent;
the keyboard response is shared by both modes.  The small aggregate difference
is therefore localized to the availability of future evidence in one sequence,
while trust-region rotation refinement preserves the parent whenever prefix support is
insufficient.

Bounded rotation gives the highest AR.  Admitting translation lowers AR by
.0032, and the unrestricted update lowers it by .0055.  Scale is numerically
inactive in this Metric CAD control because the selected candidate and parent
share the calibrated metric scale.  The result supports the rotation-only
refinement used in the main system.

\paragraph{Sensitivity to SAM3D CAD and reconstructed evidence.}
\label{sec:supp_upstream_robustness}

We perturb the frozen object-side correspondence evidence, scene-side motion
evidence, or both while keeping the candidate generator, parent poses, scored
keys, and evaluator fixed.  The perturbation levels were registered before
evaluation and are repeated with three seeds over nine recording--assets.

\paragraph{Object-generation seed control.}
\label{sec:supp_seed_control}

We pre-register three RGB-only seed policies and rerun the complete SAM3D CAD
path: SAM3D inference, onboarding calibration, candidate construction,
recording-local adaptation, and evaluation.  The policies select the largest
valid mask over the full recording, within a fixed 5--25\% early window, or
within a fixed 70--95\% late window.  They are fixed before tracking outcomes
are observed and applied under the same SAM3D CAD protocol.

\begin{table}[tbp]
\caption{Sensitivity to SAM3D CAD and reconstructed scene evidence.  Panel (a)
uses the registered HOT3D subset; ``Temporal change'' is the fraction of frames
on which posterior transport changes the framewise candidate choice.  Object-side
perturbations combine descriptor noise, candidate dropout, pose jitter, and
reliability attenuation; scene-side perturbations combine motion dropout and
rotation, translation, and feature noise.  Panel (b) reruns all downstream
stages for every non-default seed policy on the same 74 evaluation keys.}
\centering
\suppTableFont
\setlength{\tabcolsep}{4pt}
\renewcommand{\arraystretch}{1.12}
\textbf{(a) Controlled upstream-evidence sensitivity}\\[-2pt]
\begin{tabular}{lccccc}
\toprule
Evidence condition & AR $\uparrow$ & Retained $H$ & Effective $H$ & Temporal change & Update rate \\
\midrule
Clean & $.5552{\pm}.0001$ & 64.0 & 1.20 & .205 & .049 \\
SAM3D mild & $.5550{\pm}.0001$ & 61.0 & 1.22 & .178 & .041 \\
SAM3D medium & $.5553{\pm}.0001$ & 55.0 & 1.18 & .202 & .031 \\
SAM3D severe & $.5551{\pm}.0000$ & 45.9 & 1.16 & .221 & .015 \\
VGGT mild & $.5550{\pm}.0001$ & 64.0 & 1.16 & .166 & .049 \\
VGGT medium & $.5552{\pm}.0003$ & 64.0 & 1.14 & .097 & .049 \\
VGGT severe & $.5552{\pm}.0001$ & 64.0 & 1.14 & .058 & .049 \\
Combined medium & $.5550{\pm}.0001$ & 55.1 & 1.19 & .113 & .040 \\
\bottomrule
\end{tabular}

\vspace{4pt}
\begin{minipage}[t]{\linewidth}
\centering
\textbf{(b) End-to-end SAM3D seed-view control}\\[-2pt]
\suppfit{\linewidth}{%
\begin{tabular}{lcccc}
\toprule
Seed policy & Macro AR$\uparrow$ & Keyboard & Mug & Puzzle \\
\midrule
Largest visible support & \textbf{.5751} & \textbf{.5129} & \textbf{.7429} & .4694 \\
Early-window maximum & .4884 & .4343 & .5310 & \textbf{.5000} \\
Late-window maximum & .5465 & .5029 & .6976 & .4389 \\
\bottomrule
\end{tabular}
}
\end{minipage}

\label{tab:supp_upstream_robustness}
\label{tab:supp_seed_sensitivity}
\end{table}

The final AR varies by at most .0002 because the trust gate returns to the
geometric parent when perturbed evidence is unsupported.  The internal response
is nevertheless systematic: increasing SAM3D corruption reduces the update
rate from .041 to .015, while increasing VGGT corruption reduces the frequency
with which temporal transport changes the framewise choice from .166 to .058.
Across recording--assets, the worst controlled AR change is $-.0009$.  This
experiment isolates the bounded posterior outlet; the end-to-end seed-view
study in Table~\ref{tab:supp_seed_sensitivity} separately regenerates the
SAM3D CAD and all downstream support.

The global largest-support policy is the strongest deterministic rule.  Its
selected masks contain substantially more object evidence: the early/late mask
areas are 63/54\% of the selected support for keyboard, 51/34\% for the mug,
and 81/80\% for the puzzle.  The per-object response also rules out a preferred
temporal segment: the early puzzle view gains .0306 AR, while the same policy
reduces keyboard and mug AR by .0786 and .2119.  Visible support therefore
provides a reproducible onboarding criterion without choosing a category-
specific view after observing tracking accuracy.  The largest-support learned
outlet differs from its frozen structured parent by only +.0008 macro AR, so
the variation above is attributable to regenerated SAM3D CAD rather than
an unbounded posterior correction.

\paragraph{Cross-protocol adaptation stability.}
\label{sec:supp_cross_protocol_convergence}

We replay the final recording-local optimization from every frozen candidate
bank in the six evaluation protocols and parse the complete loss trajectory.
This isolates optimization stability from upstream generation, reconstruction,
and geometric support construction.  Table~\ref{tab:supp_cross_protocol_convergence}
reports the first logged epoch that reaches 95\% of the improvement attained by
the full run; losses are logged every 20 epochs, so the statistic is a
conservative discrete estimate.

\begin{table}[tbp]
\caption{Sequence-specific self-adaptation convergence across six evaluation
protocols. Median and p95 report the first logged epoch reaching 95\% of the
full-run improvement.}
\centering
\suppTableFont
\setlength{\tabcolsep}{4pt}
\suppfit{\columnwidth}{%
\begin{tabular}{lrrrr}
\toprule
Protocol & Fits & $K$ & Median @95\% & p95 @95\% \\
\midrule
HOT3D / SAM3D CAD & 23 & 64 & 260 & 480 \\
HOT3D / Metric CAD & 23 & 128 & \textbf{240} & \textbf{360} \\
YCBInEOAT / SAM3D CAD & 9 & 128 & 320 & 440 \\
YCBInEOAT / Metric CAD & 9 & 128 & 400 & 500 \\
LINEMOD / SAM3D CAD & 15 & 64 & 340 & 460 \\
LINEMOD / Metric CAD & 15 & 128 & 320 & 500 \\
\bottomrule
\end{tabular}
}

\label{tab:supp_cross_protocol_convergence}
\end{table}

All 94 recording--object fits reach epoch 600 with finite loss. The median
epoch@95\% is 310.  The median
observed loss reduction ranges from 86.1\% to 97.7\% across protocols.  Thus,
the same geometry-derived objective remains numerically stable under SAM3D CAD
and Metric CAD on all three datasets; convergence rates remain bounded across
the fixed protocols.

\paragraph{Candidate budget and system cost.}
\label{sec:supp_system_cost}

Table~\ref{tab:supp_runtime_system} separates one-time support construction and
recording-local adaptation from recurrent posterior inference.

\begin{table}[tbp]
\caption{System cost and refinement controls. Panel (a), on HOT3D with SAM3D CAD, reports support
construction on four A100 GPUs, sequence-specific self-adaptation, and
one-A100 posterior inference; upstream SAM3D and VGGT passes are excluded.
Panel (b) reports the YCBInEOAT Metric CAD refinement control after onboarding.}
\centering
\suppTableFont
\setlength{\tabcolsep}{4pt}
\renewcommand{\arraystretch}{1.12}
\begin{minipage}[t]{0.485\linewidth}
\vspace{0pt}\centering
\setlength{\tabcolsep}{2pt}
\textbf{(a) Stage-resolved runtime}\\[-2pt]
\suppfit{\columnwidth}{%
\begin{tabular}{lrr}
\toprule
Stage & Cost $\downarrow$ & Throughput $\uparrow$ \\
\midrule
Support construction & 2.061 s/frame & 0.49 FPS \\
Recording-local adaptation & 27.23 s & -- \\
Posterior inference & 1.202 ms/frame & 831.9 FPS \\
\shortstack[l]{Adaptation amortized\\over a sequence} & 1.269 s/frame & 0.79 FPS \\
\bottomrule
\end{tabular}
}
\end{minipage}\hfill
\begin{minipage}[t]{0.485\linewidth}
\vspace{0pt}\centering
\setlength{\tabcolsep}{2pt}

\textbf{(b) Bounded-refinement speed--accuracy}\\[-2pt]
\begin{tabular}{rccc}
\toprule
Iterations & AR $\uparrow$ & ADD-S $\uparrow$ & Time (s/frame) $\downarrow$\\
\midrule
0   & .8693 & 96.1 & \textbf{.1034}\\
10  & .8730 & \textbf{96.2} & .1381\\
20  & .8732 & \textbf{96.2} & .1484\\
40  & .8730 & \textbf{96.2} & .1826\\
100 & \textbf{.8738} & \textbf{96.2} & .2418\\
\bottomrule
\end{tabular}

\end{minipage}
\label{tab:supp_runtime_system}
\label{tab:refine_speed}
\end{table}

The raw bank contains 223.5 pose candidates per frame on average (240.0 at the
95th percentile), GeoCorr-Matcher retains $K=64$, and the posterior reaches an
effective support of 1.12 after measurement and transport.  This separates
broad geometric support, fixed-budget inference, and posterior concentration.
The 1.27M-parameter posterior reserves at most 148 MB.

The complete-stream cost of 0.417 s/object--frame amortizes the four-GPU
staged wall time over 9,546 chronological frames; the 2.061-s support row is
the isolated 981-s stage divided by 476 scored object--frames.  They report
complete-stream throughput and scored-bank construction, respectively.

For the YCBInEOAT Metric CAD control, the selected posterior without
refinement reaches .8693 AR. Ten iterations
retain 96.2 ADD-S AUC at .1381 s/frame, while one hundred iterations reach the
highest AR (.8738) at .2418 s/frame, providing two output settings.

The 95\% improvement time reported in the main text refers to reduction of
the adaptation objective, as defined in
Table~\ref{tab:supp_cross_protocol_convergence}.

\FloatBarrier

\end{document}